\documentclass{article}
\let\paperaddcontentsline\addcontentsline
\usepackage{iclr2027_conference,times}
\let\addcontentsline\paperaddcontentsline
\usepackage[T1]{fontenc}
\usepackage{amsmath,amssymb,amsthm,mathtools,mathrsfs}
\usepackage{booktabs,tabularx,longtable,array,graphicx,enumitem,microtype,url,xcolor}
\usepackage{hyperref}
\usepackage[capitalise]{cleveref}
\theoremstyle{plain}
\newtheorem{theorem}{Theorem}[section]
\newtheorem{proposition}[theorem]{Proposition}
\newtheorem{lemma}[theorem]{Lemma}
\newtheorem{corollary}[theorem]{Corollary}
\theoremstyle{definition}
\newtheorem{assumption}[theorem]{Assumption}
\newtheorem{definition}[theorem]{Definition}
\newtheoremstyle{boldremark}{6pt}{6pt}{\normalfont}{}{\bfseries}{.}{0.5em}
 {\thmname{#1}\thmnumber{ #2}\thmnote{ (#3)}}
\theoremstyle{boldremark}
\newtheorem{remark}[theorem]{Remark}
\usepackage{caption}
\usepackage{multirow}

\title{An Optimization Framework for Denoising Generation}
\author{Xinyu Zhou, Jiawei Zhang\thanks{Equal advising.}, and
  Stephen J. Wright\footnotemark[1]\\
  \normalfont Department of Computer Sciences, University of Wisconsin--Madison\\
  \normalfont Madison, WI, USA\\
  \texttt{\{xzhou555, jzhang2924, sjwright2\}@wisc.edu}}
\hypersetup{hidelinks,pdftitle={An Optimization Framework for Denoising Generation},
  pdfauthor={Xinyu Zhou, Jiawei Zhang, and Stephen J. Wright},pdfsubject={},pdfkeywords={}}
\iclrfinalcopy
\begin{document}
\maketitle

\begin{abstract}
Noise-unconditioned denoising learns a single score, so generation can be viewed as gradient ascent on a fixed log-density landscape. This optimization interpretation leaves two questions: do trajectories approach meaningful data, and do different initializations retain diverse outcomes? We establish both phenomena for empirical and population multi-scale Gaussian-mixture objectives. Under explicit posterior and landscape conditions, ascent converges almost surely within $O(m\sigma_{\min}^2/\tau)$ of an $m$-dimensional reference manifold with reach at least $\tau$. We derive a source--sink identity for the same fixed field and use it to prove finite-resolution regional coverage. In an affine construction, we further identify a positive-bandwidth regime where even empirical-score optimization can avoid training-sample memorization while preserving diversity at the training resolution. The same fixed objective naturally supports penalty-guided controllable generation. Experiments support our predicted mechanisms, show improved robustness under large-step guidance, and demonstrate downstream gains consistent with finite-data generative generalization.
\end{abstract}

\section{Introduction and Background}
\label{sec:introduction}

Diffusion models have become a central technology of generative AI. They produce high-quality data \citep{ho2020denoising,rombach2022high}, support scientific applications such as protein design \citep{watson2023novo}, and provide reusable priors for inverse problems \citep{chung2022diffusion}. Increasingly, the goal is not just to generate a meaningful object: It must also agree with measurements, obey physical laws, or improve  downstream tasks. 

The standard score-based description learns denoising directions indexed by noise level and combines them through a prescribed reverse-time process \citep{song2020score}. Its distributional analysis uses evolution equations; some formulations of control additionally use stochastic optimal control \citep{han2024stochastic}. These tools are powerful, while having high theoretical thresholds for the common generation and application community. However, the demands mentioned above can be formulated elegantly under an optimization framework, without losing comprehensibility. This observation motivates a simpler question:

\begin{quote}
\centering
\emph{Can optimization on a fixed objective be defined as a useful denoising generator?}
\end{quote}

To answer the question, we first revisit the development of diffusion to find its reasons for success.

\textbf{From score matching to multi-scale generation.}
Score matching \citep{hyvarinen2005estimation} learns the gradient of a log density without its normalizing constant. Denoising score matching (DSM) \citep{vincent2011connection} turns this estimation problem into regression on corrupted data. Diffusion probabilistic models \citep{sohl2015deep} connect forward corruption to reverse generation; noise-conditional score networks (NCSNs) \citep{song2019generative} and denoising diffusion probabilistic models (DDPMs) \citep{ho2020denoising} made multi-scale denoising an effective generative methodology. Reverse-time SDEs and probability-flow ODEs subsequently supplied a common language \citep{song2020score}.

Two ingredients in this progression are especially relevant here: a tractable denoising objective, and information from many noise-perturbed distributions. Under the manifold hypothesis, data occupy a low-dimensional part of ambient space \citep{fefferman2016testing}. The ideal clean distribution can even lack an ambient density, whereas generation usually starts from an ambient Gaussian. For $G\sim\mathcal N(0,I_d)$, its norm is typically about $\sqrt d$ in high dimensions; Gaussian initialization can therefore be far from a bounded data region. Learning directions only near clean data does not constrain the entire route back. NCSN addresses this low-density difficulty by perturbing data at multiple scales \citep{song2019generative}. Broad perturbations reach the initialization region; fine perturbations resolve local data structure.

\textbf{Noise unconditioning: keep the scales, remove the noise input.}
The multi-scale information need not be presented to a network as an explicit noise label. Noise unconditioning retains corrupted examples from all scales but gives the network only the current state. Optimizing this noise-unconditional DSM is equivalent to optimizing the score of a unified Gaussian mixture $p_{MN}$ with $M$ training centers and $N$ noise levels \citep{zhou2026smoothing}. The contrast is
\begin{equation}
\begin{aligned}
\text{noise conditioning:}\quad &\{\nabla\log\widehat p_{\sigma_i}\}_{i=1}^N
                   &&\longrightarrow\ \text{reverse-time denoising},\\
\text{noise unconditioning:}\quad &\nabla\log p_{MN}
                   &&\longrightarrow\ \text{ascent on a single landscape}.
\end{aligned}
\label{eq:main_conditioning_contrast}
\end{equation}


In the second case, the current state reweights all center--scale contributions. The objective remains fixed, while the current sample determines the contribution of each noise scale. This gives an optimization interpretation of the \emph{exact target field}. Recent work has studied denoising without explicit noise conditioning: \citet{sun2025noise} analyze conditioning-removal error, while \citet{kadkhodaie2026blind} show that blind denoisers can implicitly infer the relevant noise level. Related autonomous perspectives also include Equilibrium Matching \citep{wang2025equilibrium} and Beckmann Transport Models \citep{cheuk2026beckmann}. Building on the unified-density formulation, we study a complementary
question: when does ascent on this fixed landscape provide manifold fidelity, regional coverage, and non-memorizing endpoints?

\textbf{What else is missing to bridge the optimization and generation?} The key missing step is not merely to observe that noise-unconditioned sampling can be written as optimization on a fixed landscape, but to explain why optimizing that landscape constitutes a meaningful generative process. More importantly, \textbf{convergence is necessary but not sufficient for generation.} A trajectory may converge perfectly and still be useless as a generator if its limit is away from the data manifold or if nearly all initializations collapse to known modes. A complete optimization-based account therefore has to establish both \emph{fidelity} and \emph{diversity}: trajectories should approach the data manifold, and different initializations should continue to populate its different regions.

We prove complementary geometric and probability guarantees for empirical and population noise-unconditioned landscapes. Positive reach $\tau$ and localized posterior geometry bound the endpoint's distance to the manifold by $O(m\sigma_{\min}^2/\tau)$, where $m$ is the dimension of the manifold.
A source-sink identity links points on the same ascent paths to a minimum-noise smoothing of the reference population; geometric retention transfers this probability to the stopped output.

The same framework also reveals a regime in which empirical-score optimization can generalize beyond the training set. In affine geometry (large reach), normal accuracy no longer requires vanishing noise, leaving a positive bandwidth where score smoothing \citep{zhou2026smoothing} suppresses sample-level memorization while preserving manifold variation. We construct finite training sets whose infinite-time endpoints remain separated from training samples while retaining coverage at the training resolution. But whether such a window extends to general manifolds remains open. Consistent with finite-data generative generalization, augmentation from a noise-unconditioned generator trained on only 100 minority-class examples outperforms repetition and noise-conditioned diffusion on an imbalanced CIFAR-10 task.

\textbf{Why the optimization view matters.}
A fixed prior objective turns \textit{controllable} generation into a familiar constrained- or penalized-optimization problem. Measurements, inequalities, or downstream preferences can be represented by penalty functions and added directly to the generative objective, without introducing a separate projected sampling process. This approach is particularly useful when condition-output pairs are scarce but a scientific requirement can be formulated explicitly, as a constraint or penalty. Guidance and diffusion inverse-problem methods already support inference-time control \citep{dhariwal2021diffusion,ho2022classifier,chung2022diffusion}; plug-and-play methods likewise use learned denoisers in reconstruction \citep{venkatakrishnan2013plug,zhu2023denoising}. Our formulation makes the prior objective explicit and uses the same gradient-based interface, with or without control.

Our optimization-based approach also suggests an advantage under aggressive updates or strong guidance. A large update in a noise-conditioned sampler can push the iterate outside the high-density region associated with its prescribed noise level (marginal distribution), producing a state--noise mismatch and an inaccurate subsequent score estimate \citep{zhong2026mitigating}. Noise unconditioning instead learns a single field from data perturbed across all noise scales, giving the training distribution much broader coverage of state space. At inference, the current state itself reweights the contributions of different scales, allowing the
field to adapt to the locally relevant noise scale rather than relying on an externally scheduled label. This motivates our large-step guided-sampling experiments.

\textbf{Relation to prior work.}
For the unified DSM density of \citet{zhou2026smoothing},
we use the center--scale posterior to bound normal endpoint error and
tangential drift during refinement. These bounds relate to convergence in
Gaussian mean shift \citep{comaniciu2002mean,carreira2007gaussian}
and to geometric bias in density-ridge estimation
\citep{genovese2014nonparametric}.
Our affine construction complements work on associative memory and closed-form
diffusion scores \citep{pham2025memorization,scarvelis2025closed}
by establishing training-set separation and regional coverage for the same endpoints.

Poisson flow and Beckmann transport construct generative fields from prescribed
flux constraints \citep{xu2022poisson,cheuk2026beckmann}.
For the field specified by DSM, the $\sigma^2$ weights give a source--sink
balance between its two positive-noise cutoff marginals, up to a finite-grid
residual. We use this balance to bound regional mass at stopping, complementing
trajectory regularity results for time-dependent diffusion ODEs
\citep{chen2024trajectory}.
Our penalty formulation shares the use of denoising priors for optimization
with RED and gradient-step plug-and-play \citep{romano2017little,hurault2021gradient}.

Overall, our main contributions are as follows.
\begin{itemize}[leftmargin=*,topsep=0pt,itemsep=1pt]

\item \textbf{A generative theory for noise-unconditioned optimization.}
Under explicit geometric and posterior conditions, we prove that fixed-objective
ascent converges within $O(m\sigma_{\min}^2/\tau)$ of the data manifold for
both empirical and population objectives.

\item \textbf{Diversity beyond convergence.}
We derive a source--sink identity for the same field and use it to prove
finite-resolution regional coverage for stopped outputs, showing how initialized
trajectories retain probability across data regions.

\item \textbf{Non-memorizing empirical-score generation.}
For a constructive affine family, we identify a positive-bandwidth regime in
which exact empirical-score optimization avoids training-sample memorization
while preserving diversity at the training resolution.

\item \textbf{Control and practical utility.}
The fixed landscape provides a direct penalty-guided interface for controllable
generation. Experiments support the theoretical mechanisms, demonstrate
large-step robustness, and show downstream gains consistent with finite-data
generative generalization.
\end{itemize}

\section{A Fixed Landscape and Manifold Convergence}
\label{sec:gradient_ascent}
\vspace{-0.1cm}

We use empirical notation to expose finite-data effects (corresponding to practice); the same framework applies to population objectives (Remark~\ref{remark:main_population_extension}).

\subsection{Learning the gradient, optimizing the sample}
\label{subsec:main_fixed_landscape}
\vspace{-0.1cm}

Fix training centers $\mathcal S_M=\{\mu_1,\ldots,\mu_M\}\subset\mathbb R^d$ and scales $0<\sigma_{\min}=\sigma_1\le\cdots\le\sigma_N=\sigma_{\max}$. Let $\widehat Q_M=M^{-1}\sum_j\delta_{\mu_j}$ be the empirical training distribution. Uniform sampling over centers and scales gives the noise-unconditioned loss
\begin{equation}
\mathcal L_u(\theta)=\frac1{2MN}\sum_{j,i}\sigma_i^2
 \mathbb E_{X\sim\mathcal N(\mu_j,\sigma_i^2I)}
 \left\|s_\theta(X)-\frac{\mu_j-X}{\sigma_i^2}\right\|^2.
\label{eq:main_dsm_loss}
\end{equation}
The network receives only $X$. Define the unified density
\begin{equation}
p_{MN}(x)=\frac1{Z_{MN}}\sum_{j,i}\sigma_i^2\mathcal N(x;\mu_j,\sigma_i^2I),
\qquad Z_{MN}=M\sum_i\sigma_i^2.
\label{eq:main_unified_density}
\end{equation}
Completing the square identifies its gradient as the exact learning target \citep{zhou2026smoothing}:
\begin{equation}
\mathcal L_u(\theta)=\frac{Z_{MN}}{2MN}
 \mathbb E_{X\sim p_{MN}}\|s_\theta(X)-\nabla\log p_{MN}(X)\|^2+C,
\label{eq:main_excess_loss}
\end{equation}
where $C$ is independent of $\theta$. Thus \emph{training optimizes the parameters to learn a gradient; generation optimizes the sample along that gradient}:
\begin{equation}
\dot x=\nabla\log p_{MN}(x),\qquad
x_{n+1}=x_n+\eta_n s_\theta(x_n).
\label{eq:main_ascent_dynamics}
\end{equation}
Our theory analyzes the exact empirical field $\nabla\log p_{MN}$ and its
population analogue; $s_\theta$ is the learned approximation used in practice.
Appendix~\ref{app:learning_gradient} discusses the relation among them.

\paragraph{How one objective uses multiple scales.}
Let $w_{ij}(x)$ be the mixture responsibility proportional to $\sigma_i^2\mathcal N(x;\mu_j,\sigma_i^2I)$, and define
\begin{equation}
\beta=\sum_{j,i}\frac{w_{ij}}{\sigma_i^2},\qquad
\widetilde w_{ij}=\frac{w_{ij}}{\sigma_i^2\beta},\qquad
\bar\mu=\sum_{j,i}\widetilde w_{ij}\mu_j.
\label{eq:main_posterior_definitions}
\end{equation}
All three depend on $x$. Differentiation yields
\begin{equation}
\nabla\log p_{MN}(x)=\beta(x)(\bar\mu(x)-x),\qquad
\sigma_{\max}^{-2}\le\beta(x)\le\sigma_{\min}^{-2}.
\label{eq:main_barycentric_score}
\end{equation}
The field moves toward a state-dependent average of training centers, with $\beta$ setting its speed. Broad kernels contribute away from the data and finer kernels resolve nearby structure, within one unchanged objective. Every critical point is a convex combination of training centers. This already bounds where optimization can end, but a convex combination can lie off a curved manifold. The next argument controls precisely that geometric error.

\subsection{Geometry controls normal convergence}
\label{subsec:main_manifold_convergence}
Let $\mathcal M$ be a compact embedded $C^2$ reference manifold without boundary, containing the centers, with dimension $1\le m<d$ and reach at least $\tau>0$. Reach measures the size of a neighborhood with unique nearest-point projection. Set $c=d-m$, $r(x)=\operatorname{dist}(x,\mathcal M)$ and $z_x=\Pi_{\mathcal M}(x)$.

For initialization supported on a nonempty compact set $K$, define
$\Omega=\{x:p_{MN}(x)\ge\min_Kp_{MN}\}$. Ascent increases $\log p_{MN}$, so the trajectory stays in $\Omega$; Gaussian decay makes this set compact. Analyticity then gives convergence to a critical point \citep{absil2005convergence}. To show that the limit represents the data geometry, choose $0<r_0<\tau$, $r_0\le\sigma_{\max}\sqrt c$, and require, for some dimensionless constant $C_{\mathrm{loc}} >0$
\begin{equation}
\sum_{j,i}\widetilde w_{ij}(x)\|\mu_j-z_x\|^2
\le C_{\mathrm{loc}}m\max\{\sigma_{\min}^2,r(x)^2/c\},
\quad x\in\Omega,\quad 0<r(x)<r_0.
\label{eq:main_posterior_localization}
\end{equation}
This condition controls how widely the active training centers spread around the nearest manifold point. The scale $r/\sqrt c$ is motivated by a flat-patch calculation; the inequality controls the full posterior, not just its most likely scale (Appendix~\ref{sec:dominant_precision_scale}).

Put
\begin{equation}
\varepsilon_0=\frac{C_{\mathrm{loc}}m\sigma_{\min}^2}{2\tau},\qquad
r_{\min}=\sigma_{\min}\sqrt c,\qquad
\kappa_0=\frac{C_{\mathrm{loc}}mr_0}{2c\tau},
\label{eq:main_geometric_constants}
\end{equation}
and assume $\kappa_0<1$ and $\varepsilon_0<\min\{r_{\min},r_0\}$. Reach bounds the normal component of a chord by its squared length divided by $2\tau$. Averaging this bound with \eqref{eq:main_posterior_localization} gives
\begin{equation}
\dot r\le
\begin{cases}
-\beta(x)(r-\varepsilon_0),&0<r\le r_{\min},\\
-\beta(x)(1-\kappa_0)r,&r_{\min}<r<r_0.
\end{cases}
\label{eq:main_radial_contraction}
\end{equation}
The direct pull toward the manifold dominates the curvature bias of the average, excluding critical points from $\varepsilon_0<r<r_0$. Large reach allows stronger smoothing before that bias becomes significant.

We also need to exclude attracting equilibria farther away. Write $a_{ij}=(\mu_j-x)/\sigma_i^2$ and $C_x=\operatorname{Cov}_{w(x)}(a_{ij})$. Since $\nabla^2\log p_{MN}=C_x-\beta I$, the conditions
$\operatorname{tr}C_x\ge\alpha_d d\beta$ and
$\operatorname{tr}C_x/\lambda_{\max}(C_x)\le R_d<\alpha_dd$
at off-tube critical points imply
\begin{equation}
\lambda_{\max}(\nabla^2\log p_{MN}(x))
\ge\left(\frac{\alpha_d d}{R_d}-1\right)\beta(x)>0,
\qquad 0<\alpha_d\le1.
\label{eq:main_off_tube_instability}
\end{equation}
Posterior variation in sufficiently few directions therefore makes such points unstable for ascent. Local attraction and off-tube instability together locate the generated endpoints.

\begin{theorem}[Manifold convergence]
\label{thm:main_convergence}
Under the preceding conditions, suppose each off-tube critical point in $\Omega$ has a nonsingular Hessian $\nabla^2\log p_{MN}$, and $\nu$ is absolutely continuous with respect to ambient Lebesgue measure, with $\nu(K)=1$. Every exact trajectory from $K$ converges, and
\begin{equation}
\operatorname{dist}(x_\infty,\mathcal M)\le\varepsilon_0
=\frac{C_{\mathrm{loc}}m}{2\tau}\sigma_{\min}^2
\quad\nu\text{-almost surely}.
\label{eq:main_convergence_result}
\end{equation}
Every local maximizer in $\Omega$ and every global maximizer satisfy the same bound, without the nonsingularity or initialization-density conditions.
\end{theorem}
Appendix~\ref{app:convergence_details} proves the result. It supplies the fidelity part of the optimization framework: optimization converges, and the limit is close to the reference data manifold.

\paragraph{Flat geometry permits smoothing without normal error.}
If the centers lie in an affine subspace $\mathcal A=a_0+V$, so does $\bar\mu(x)$, giving the exact identity
\begin{equation}
r_{\mathcal A}(x(t))\le r_{\mathcal A}(x(0))e^{-t/\sigma_{\max}^2}\longrightarrow0.
\label{eq:main_affine_normal_convergence}
\end{equation}
This holds for every fixed positive schedule (Proposition~\ref{prop:affine_subspace_exact_normal_convergence}). It separates the smoothing used to form the objective from its final normal accuracy. Section~\ref{subsec:main_nonmemorization} uses this separation to preserve smoothing against memorization while retaining variation along the manifold.

\section{Diversity, Non-Memorization, and Control}
\label{sec:diversity_control}
Manifold convergence makes the optimizer geometrically meaningful. To make it a generator, we must also control how initialization probability is distributed among its outcomes. We first establish that coverage/diversity, then combine it with positive smoothing to obtain non-memorizing endpoints, and finally use the same fixed objective as a prior for penalty-guided controllable generation.

\subsection{Retaining probability across data regions}
\label{subsec:main_diversity}
\label{subsec:main_source_sink}
Assume $\mathcal M$ is connected and specify a reference data distribution $Q_{\mathcal M}$ on it. The manifold describes geometry; $Q_{\mathcal M}$ supplies the regional probabilities that generation should retain. We first state the result for the empirical $\nabla\log p_{MN}$. For a Borel region $U\subseteq\mathcal M$, define
\begin{equation}
U^{-\delta}=\{z\in U:d_{\mathcal M}(z,\mathcal M\setminus U)>\delta\},
\label{eq:main_safe_interior}
\end{equation}
where $d_{\mathcal M}$ is intrinsic distance. This safe interior ensures that any point displaced by at most $\delta$ remains within the region.

\paragraph{Local refinement preserves incoming diversity.}
Choose $\varepsilon_0<r_{\mathrm{stop}}<r_{\min}<r_{\mathrm{tr}}<r_0$. Let $\tau_{\mathrm{tr}}$ and $\tau_{\mathrm{stop}}$ be the first entrance times into these distance tubes, and write $X_{\mathrm{stop}}=x(\tau_{\mathrm{stop}})$, $Z_{\mathrm{stop}}=\Pi_{\mathcal M}(X_{\mathrm{stop}})$ and $Z_{\mathrm{tr}}=\Pi_{\mathcal M}(x(\tau_{\mathrm{tr}}))$. Comparing projected tangential speed with inward speed gives
\begin{equation}
\int_{\tau_{\mathrm{tr}}}^{\tau_{\mathrm{stop}}}
 \left\|\frac{d}{dt}\Pi_{\mathcal M}(x(t))\right\|dt\le\delta_{\mathrm{stop}}
\quad\text{almost surely}.
\label{eq:main_path_retention}
\end{equation}
\cref{prop:early_stopping_retention} gives an explicit bound of order
$\sqrt m\,\sigma_{\min}$ for admissible noise-proportional radii and
controlled geometric constants. Hence the refinement stage can only induce
limited redistribution of probability after trajectories enter the manifold
neighborhood. The remaining question is therefore how much probability reaches
each region.

\textbf{The same multi-scale objective supplies the probability argument.}
Let $p_\sigma^Q=Q_{\mathcal M}*\mathcal N(0,\sigma^2I)$. On a finite geometric grid $\sigma_i=\sigma_{\min}e^{(i-1)h}$, set $A_N=h\sum_i\sigma_i^2$. Gaussian scale calculus gives
\begin{equation}
\nabla\!\cdot\!\left(A_Np_{MN}\nabla\log p_{MN}\right)
 =p_{\sigma_{\max}}^Q-p_{\sigma_{\min}}^Q+\mathcal R_N,
\qquad \epsilon_{\mathrm{Pois}}=\tfrac12\|\mathcal R_N\|_1.
\label{eq:main_flux_residual}
\end{equation}
The broad smoothing supplies probability and the fine smoothing receives it. The residual accounts for the finite grid and the difference between training and reference distributions. The $\sigma^2$ weighting is important: with log-scale integration, $\partial_\sigma p_\sigma^Q=\sigma\Delta p_\sigma^Q$ makes the balance telescope to the two cutoffs.

This stationary balance concerns paths, not their distributions at one common time. Its occupation-density argument places an auxiliary point $Y$ on each original trajectory such that
\begin{equation}
\operatorname{TV}(P_Y,p_{\sigma_{\min}}^Q(x)\,dx)
\le\epsilon_{\mathrm{Pois}}+\epsilon_{\mathrm{init}},\qquad
\epsilon_{\mathrm{init}}=\operatorname{TV}(\nu,p_{\sigma_{\max}}^Q(x)\,dx),
\label{eq:main_marked_law}
\end{equation}
where $\operatorname{TV}$ is the Total Variation distance and $P_Y$ denotes the distribution of $Y$. The sampler still returns $X_{\mathrm{stop}}$; the mark connects that path to reference data probabilities in the proof (Appendix~\ref{app:source_sink}).

For confidence parameter $\zeta\in(0,1)$, let $a_d,a_m$ be upper Gaussian norm thresholds and $b_c$ a lower normal threshold, each with tail probability $\zeta/3$. Define
\begin{equation}
\delta_{\mathrm{noise}}=\frac{\sigma_{\min}a_m}{1-\sigma_{\min}a_d/\tau},
\qquad \delta_{\mathrm{cov}}=\delta_{\mathrm{noise}}+\delta_{\mathrm{stop}}.
\label{eq:main_coverage_resolution}
\end{equation}
If $\sigma_{\min}a_d\le r_{\mathrm{tr}}$ and
$r_{\mathrm{stop}}<\sigma_{\min}b_c-\delta_{\mathrm{noise}}^2/(2\tau)$, a reference perturbation falls in the refinement segment with probability at least $1-\zeta$. Its projection is within $\delta_{\mathrm{noise}}$ of its reference point. Combining this displacement with \eqref{eq:main_path_retention} transfers reference mass to the returned sample.

\begin{theorem}[Regional coverage]
\label{thm:main_coverage}
Under Theorem~\ref{thm:main_convergence} and the preceding radius, grid and Gaussian conditions, $r(X_{\mathrm{stop}})\le r_{\mathrm{stop}}$ almost surely, and for every Borel $U\subseteq\mathcal M$,
\begin{equation}
\mathbb P_\nu[Z_{\mathrm{tr}}\in U,\ Z_{\mathrm{stop}}\in U]
\ge(1-\zeta)Q_{\mathcal M}(U^{-\delta_{\mathrm{cov}}})
 -\epsilon_{\mathrm{Pois}}-\epsilon_{\mathrm{init}}.
\label{eq:main_routing_free_coverage}
\end{equation}
\end{theorem}
Together, these theorems establish manifold fidelity and finite-resolution
regional coverage for stopped outputs through the same fixed-objective ascent
process. Appendix~\ref{app:diversity_details} gives the proof and error bounds.
The next result adds a distinct property: diverse, non-memorizing endpoints
even with an exact empirical score.

\subsection{Non-memorizing endpoints at the training resolution}
\label{subsec:main_nonmemorization}
\vspace{-0.1cm}

Beyond fidelity and diversity, a positive smoothing floor can also prevent training-set memorization. At one scale,
$w_{ij}/w_{i\ell}=\exp((\|x-\mu_\ell\|^2-\|x-\mu_j\|^2)/(2\sigma_i^2))$:
small noise amplifies a training center's distance advantage \citep{zhou2026smoothing}. Affine contraction allows us to smooth this contrast while keeping zero normal endpoint error. We now show that this can also preserve generalization.

In $\mathcal A=\mathbb R^m\times\{0\}$, with $c=d-m\ge2$, choose spacing $\ell>0$ and the interlaced training grid
\begin{equation}
\mathcal S_M=\{(\ell(\mathbf k+\boldsymbol\epsilon/6),0):
 \mathbf k\in\{-L,\ldots,L\}^m,\ \boldsymbol\epsilon\in\{-1,1\}^m\},
\quad M=2^m(2L+1)^m.
\label{eq:resolved_grid_training_centers}
\end{equation}
Let $Q_{\mathcal M}$ be uniform on the connected box $D_L=[-(L+1/2)\ell,(L+1/2)\ell]^m\times\{0\}$. Its training covering radius is $h_{\mathrm{cov},M}=\sqrt m\,\ell/3$. Each grid cell carries training and reference mass $2^m/M$, and contains the output core
$C_{\mathbf k}=\{(u,0):|u_j-\ell k_j|\le\ell/12\ \text{for all }j\}$.

\begin{corollary}[Non-memorization with sample-resolution diversity]
\label{cor:main_nonmemorization}
\label{cor:sample_resolution_nonmemorization}
Fix $m\ge1$, $c\ge2$ and $\epsilon\in(0,1/4)$. There are finite constants $s_0\ge1/2,L_0$ such that, for every $L\ge L_0$, one finite relative geometric schedule works throughout $\sigma_{\min}\in[s_0\ell,2s_0\ell]$. Every exact trajectory converges to $x_\infty\in\mathcal A\cap\operatorname{conv}(\mathcal S_M)$. Under initialization $\mathcal N(0,\sigma_{\max}^2I_d)$, for every Borel $U\subseteq\mathcal A$,
\begin{equation}
\begin{split}
\mathbb P[x_\infty\in U,\ \operatorname{dist}(x_\infty,\mathcal S_M)\ge h_{\mathrm{cov},M}/4]\ge Q_{\mathcal M}(U^{-7h_{\mathrm{cov},M}/4})-\epsilon.
\end{split}
\label{eq:resolved_endpoint_population_coverage}
\end{equation}
For every collection $I\subseteq\{-L,\ldots,L\}^m$,
\begin{equation}
\left|\mathbb P\!\left[x_\infty\in\bigcup_{\mathbf k\in I}C_{\mathbf k}\right]
       -\frac{|I|}{(2L+1)^m}\right|\le\epsilon.
\label{eq:resolved_endpoint_cell_mass}
\end{equation}
\end{corollary}
The same outputs carry regional coverage and a training-distance margin. At least $(1-2\epsilon)M/2^m$ disjoint cores receive probability at least $2^{m-1}/M$, so a connected patch retains many outcomes rather than one representative. Both the margin and the coverage resolution scale with the training covering radius. The cell-probability comparison also controls \emph{any union} of cells, giving distributional rather than merely geometric coverage.

The proof verifies finite-mixture gradient signs in small capture cells, then follows the source--sink marks to their endpoints (Appendix~\ref{app:resolved_endpoint_proof}). It establishes a nonempty bandwidth window and a sequence whose endpoint distributions converge in Wasserstein-1 distance $W_1$ to a continuous reference on a fixed box. Its significance is that \emph{even exact empirical-score optimization can avoid memorization while retaining diversity at the data's sampling resolution}.

The window balances sample-level smoothing against the variation needed to distinguish nearby regions. Curvature adds the normal-accuracy restriction $C_{\mathrm{loc}}m\sigma_{\min}^2/(2\tau)\le\varepsilon$; larger reach relaxes it when localization remains controlled. The contrast with diffusion is specifically its exact empirical zero-noise endpoint, which recovers $\widehat Q_M$ (Appendix~\ref{subsec:diffusion_endpoint_comparison}).

\begin{remark}[Population distributions]
\label{remark:main_population_extension}
Theorems~\ref{thm:main_convergence} and~\ref{thm:main_coverage}
also hold for
$p^{\mathrm{pop}}(x)=
\sum_i\sigma_i^2p_{\sigma_i}^Q(x)/\sum_i\sigma_i^2$
under the corresponding population assumptions.
Center sums become $Q_{\mathcal M}$-expectations, and
$\epsilon_{\mathrm{Pois}}$ contains only noise-grid error;
$\epsilon_{\mathrm{init}}$ remains.
The same coupling gives
$P_{X_{\mathrm{stop}}}\to Q_{\mathcal M}$ in $W_1$
along admissible model families on a fixed compact manifold as
spatial-resolution and probability errors vanish
(Appendix~\ref{app:population_extension}).
Corollary~\ref{cor:main_nonmemorization} complements this population
approximation with diverse, non-memorizing empirical-score endpoints;
its fixed-box construction also approaches a continuous reference
distribution in $W_1$.
\end{remark}

\subsection{Controllable generation by penalty-guided optimization}
\label{subsec:optimization_control}
\vspace{-0.1cm}

The fixed objective also provides a simple interface for controllable generation.
Let $y$ denote a task, observation, or desired property, and let
$E_y(x)\ge0$ be a differentiable task penalty whose smaller values indicate
better task satisfaction. Rather than changing the generative prior, we maximize
\begin{equation}
F_y(x)=\log p_{MN}(x)-\lambda E_y(x),\quad \mbox{for some $\lambda\ge0$},
\label{eq:main_control_objective}
\end{equation}
by ordinary gradient ascent:
\begin{equation}
x_{n+1}
=
x_n+\eta_n\left[
\nabla\log p_{MN}(x_n)-\lambda\nabla E_y(x_n)
\right], \quad n=0,1,2,\dotsc,
\label{eq:main_penalty_update}
\end{equation}
for a sequence of steplengths $\{ \eta_n \}$. 
The parameter $\lambda$ trades off the generative landscape against the task objective.
(In practice, $\nabla\log p_{MN}$ can be replaced by the learned state-only score
$s_\theta$.) With the exact score, this is gradient ascent on the single objective
$F_y$; under the analytic conditions detailed in
Appendix~\ref{app:control_details}, the corresponding flow converges
to a stationary point satisfying
\[
\nabla\log p_{MN}(x_{\infty,y})
=
\lambda\nabla E_y(x_{\infty,y}).
\]

\textbf{Compatibility with generative geometry.}
For tasks that primarily modify semantic or latent degrees of freedom along the
data manifold, the guidance near the manifold is naturally expected to be
predominantly tangential; its normal component measures how strongly the task
pushes against the learned data geometry. Inside the manifold tube, let
$u_x=(x-\Pi_{\mathcal M}(x))/r(x)$. Assume that the same posterior-localization bound holds on the guided tube
region and that there exists $\delta_y\ge0$ such that
\begin{equation}
\lambda[-\langle u_x,\nabla E_y(x)\rangle]_+
\le \beta(x)\delta_y,
\label{eq:main_guidance_normal_budget}
\end{equation}
where $\beta(x)$ is defined in \eqref{eq:main_barycentric_score}.
Here $\delta_y$ quantifies the outward normal guidance relative to the prior's
local contraction scale. Combining this bound with
\eqref{eq:main_radial_contraction} preserves inward attraction outside the
inflated radius
\begin{equation}
\varepsilon_y
=
\max\left\{
\varepsilon_0+\delta_y,\,
\frac{\delta_y}{1-\kappa_0}
\right\}.
\label{eq:main_guided_manifold_radius}
\end{equation}
Provided $\varepsilon_y<r_0$, the guided flow retains this inward attraction after entering the tube. Tangential or inward guidance ($\delta_y=0$) leaves the local normal bound unchanged, while a small outward component enlarges it continuously. Analogously, multiple task-compatible regions remain locally available when guidance does not overwhelm their separating/inward margins. Appendix~\ref{app:control_details} gives the convergence analysis and sufficient conditions for preserving local manifold attraction and multiple task-compatible regions under penalty guidance.

\clearpage
\section{Experiments}
\label{sec:experiments}
\vspace{-0.13cm}

\begin{figure}[!htb]
\centering
\includegraphics[width=0.9\linewidth]{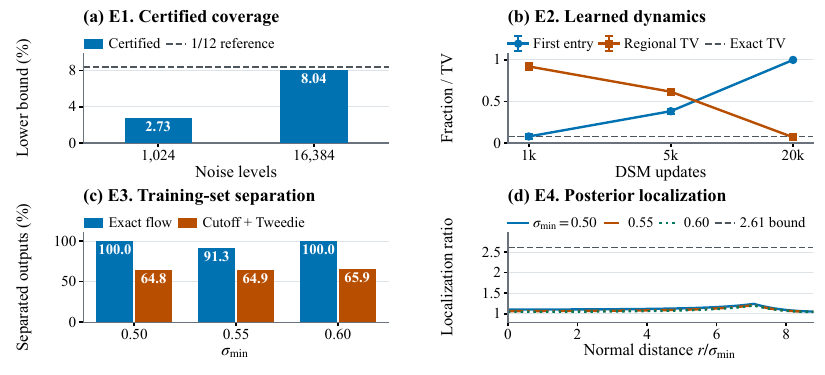}
\captionsetup{skip=4pt}
\caption{\textbf{Numerical verification for E1--E4.}
(a) Certified first-entry mass per region for two population-sphere grids.
(b) Tube entry and twelve-region TV for learned sphere scores, counting failures in a separate bin (mean $\pm$ SD across three seeds). The dashed line gives exact-score TV.
(c) Fraction of affine outputs at least $1/12$ from the training set.
(d) Maximum sampled localization ratios and their common uniform bound on the affine grids.}
\label{fig:revision_e1_e4}
\end{figure}

We test the theoretical mechanisms with exact and learned scores (Figure~\ref{fig:revision_e1_e4}),
then evaluate noise unconditioning for image generation, guided sampling, and data augmentation.

\subsection{Numerical verification of the theoretical mechanisms}
\label{subsec:main_numerical}
\vspace{-0.07cm}

\textbf{E1. Regional coverage.}
We partition the uniform sphere $S^2(1)\subset\mathbb R^{12}$ into twelve
equal-mass bands and check the conditions of
Corollary~\ref{cor:population_regional_coverage}.
The 1,024- and 16,384-level grids certify first-entry mass of at
least $2.73\%$ and $8.04\%$ per band, respectively, against a reference mass
of $8.33\%$. For a 24-level grid on $[0.02,2]$, the Poisson error is at
least $0.168$, exceeding the reference mass of one band
(Appendix~\ref{subsec:revision_e1}).

\textbf{E2. Off-tube instability and learned dynamics.}
On three 512-center sphere mixtures, the exact scores satisfy
Assumption~\ref{assumption:off_tube_geometry}, with a trace-to-precision
ratio of at least $10.27$ and effective rank at most three at off-tube
critical points. From 1,000 to 20,000 ordinary DSM updates, mean
relative score RMSE falls from $0.68$ to $0.26$, entry into the
distance-$0.03$ tube rises from $7.9\%$ to $99.9\%$, and regional TV falls
from $0.92$ to $0.07$, compared with $0.08$ for the exact score
(Appendix~\ref{subsec:revision_e2}).

\textbf{E3. Endpoint separation and regional mass.}
On an 82-center affine grid in $\mathbb R^{51}$, we compare exact-score
ascent with sampling from the empirical diffusion cutoff marginal followed
by a Tweedie readout. For all tested bandwidths
$s=\sigma_{\min}/\ell\in\{0.50,0.55,0.60\}$, ascent meets the training-set
margin $1/12$ more often. Its empirical TV over three regions is $0.0114$ at
each floor, with a common certified upper bound of $0.0555$
(Appendix~\ref{subsec:revision_e3}).

\textbf{E4. Localization and curvature bias.}
The E3 grids satisfy the local mass and schedule conditions of
Appendix~\ref{subsec:local_mass_localization}.
The bound $C_{\mathrm{loc}}<2.61$ holds uniformly throughout all
three affine tubes. On population spheres, the normal bias scales as
$\sigma_{\min}^2/\tau$, with fitted exponents $2.0006$--$2.0091$
(Appendix~\ref{sec:e4}).

\textbf{Learned first-entry coverage.}
On a uniform circle, the exact field satisfies localization and contraction,
and its off-tube equilibrium is repelling. We fit three MLPs with DSM
followed by supervision on the exact score and its derivatives.
We verify repulsion, inward motion, bounded tangential drift, and
ascent on the relevant domains. With the full Gaussian prior, each learned
field guarantees first-entry mass of at least $7.18\%$ in every region of a
twelve-region partition in continuous time, and $7.14\%$ with
real-arithmetic Euler integration
(Appendix~\ref{sec:learned_circle}).
\label{experiments:theory_end}

\subsection{Noise unconditioning across diffusion and flow models}
\label{sec:noise_unconditioning}
\vspace{-0.15cm}

\begin{table*}[!htb]
\centering

\begin{minipage}[t]{0.46\textwidth}
\centering
\captionsetup{skip=0pt}
\captionof{table}{Comparison on CIFAR-10.}
\label{tab:time_cond_cifar}

\scriptsize
\setlength{\tabcolsep}{2.0pt}
\renewcommand{\arraystretch}{1.00}

\begin{tabularx}{\linewidth}{
@{}
>{\raggedright\arraybackslash}p{0.26\linewidth}
>{\centering\arraybackslash}p{0.105\linewidth}
>{\centering\arraybackslash}p{0.115\linewidth}
>{\raggedright\arraybackslash}X
@{}}
\toprule
\multicolumn{1}{c}{\multirow{2}{*}{\normalsize\bfseries Model}}
& \multicolumn{2}{c}{\bfseries FID $\downarrow$}
& \multicolumn{1}{c}{\multirow{2}{*}{\normalsize\bfseries Setting}} \\
\cmidrule(lr){2-3}
& \multicolumn{1}{c}{\bfseries w/ $t$}
& \multicolumn{1}{c}{\bfseries w/o $t$}
& \\
\midrule

VP-SDE\textsuperscript{R}
& 3.42 & 5.77
& remove $t$; SDE-1000 \\

EDM [1]\textsuperscript{P}
& 1.99 & 3.36
& remove $t$; Heun-35 \\

uEDM [1]\textsuperscript{P}
& 2.04 & 2.23
& uEDM precond.; Heun-35 \\

FM / 1-RF [1]\textsuperscript{P}
& 3.01 & \textbf{2.61}
& remove $t$; Euler-100 \\

VE-SDE [2]\textsuperscript{R}
& 6.43 & 7.47
& remove emb.; PC-SDE-1K \\

Linear FM [3]\textsuperscript{P}
& 5.40 & \textbf{5.31}
& set $t=0$; Euler-250 \\

\bottomrule
\end{tabularx}
\end{minipage}
\hfill
\begin{minipage}[t]{0.52\textwidth}
\centering
\captionsetup{skip=0pt}
\captionof{table}{Comparison on ImageNet $256{\times}256$.}
\label{tab:time_cond_imagenet}

\scriptsize
\setlength{\tabcolsep}{2.0pt}
\renewcommand{\arraystretch}{1.00}

\begin{tabularx}{\linewidth}{
@{}
>{\raggedright\arraybackslash}p{0.30\linewidth}
>{\centering\arraybackslash}p{0.12\linewidth}
>{\centering\arraybackslash}p{0.13\linewidth}
>{\raggedright\arraybackslash}X
@{}}
\toprule
\multicolumn{1}{c}{\multirow{2}{*}{\normalsize\bfseries Model}}
& \multicolumn{2}{c}{\bfseries FID $\downarrow$}
& \multicolumn{1}{c}{\multirow{2}{*}{\normalsize\bfseries Setting}} \\
\cmidrule(lr){2-3}
& \multicolumn{1}{c}{\bfseries w/ $t$}
& \multicolumn{1}{c}{\bfseries w/o $t$}
& \\
\midrule

SiT-B/2 [1]\textsuperscript{P}
& 7.83 & 7.96
& remove $t$; Euler-250 + CFG \\

DiT-B/2 [5]\textsuperscript{R}
& 19.83 & 38.54
& remove $t$; VP-DDIM \\

DiT-B/2 [5]\textsuperscript{R}
& 19.83 & \textbf{10.01}
& modified VP schedule; no $t$ \\

EqM-B/2 [4]\textsuperscript{R}
& 38.21 & 42.74
& remove $t$; const. target \\

SiT-XL/2$\to$EqM [4]\textsuperscript{R}
& 2.42 & \textbf{2.17}
& EqM target + NAG-GD \\

FM-XL/2$\to$EqF [6]\textsuperscript{P}
& 18.91 & \textbf{15.15}
& $\eta$-warp + NAG; no $t$ \\

\bottomrule
\end{tabularx}
\end{minipage}

\begin{minipage}{0.99\textwidth}
\scriptsize
\textit{Sources.}
[1]~\citealp{sun2025noise};
[2]~\citealp{zhou2026smoothing};
[3]~\citealp{helbling2026time};
[4]~\citealp{wang2025equilibrium};
[5]~\citealp{li2026exploring};
[6]~\citealp{lillemark2026equilibrium}.
\textsuperscript{R} Reproduced from the indicated setup.
\textsuperscript{P} Published result.
\end{minipage}

\end{table*}

Tables~\ref{tab:time_cond_cifar} and~\ref{tab:time_cond_imagenet}
report Fr\'echet Inception Distance (FID) with and without explicit time or
noise inputs. Rows marked R report our reproductions, including VP-SDE
\citep{ho2020denoising}. Direct removal improves FID for FM / 1-RF and
Linear FM on CIFAR-10, while DiT-B/2 on ImageNet is sensitive to direct
removal. Variants that also adapt the schedule, target, preconditioning,
or sampler attain competitive FID in several settings.

\subsection{Large-step controllable generation}
\label{subsec:large_step_experiments}
\vspace{-0.1cm}

\begin{figure}[!htb]
  \centering
  \begin{minipage}[t]{0.69\linewidth}
    \vspace{0pt}
    \includegraphics[width=0.98\linewidth]{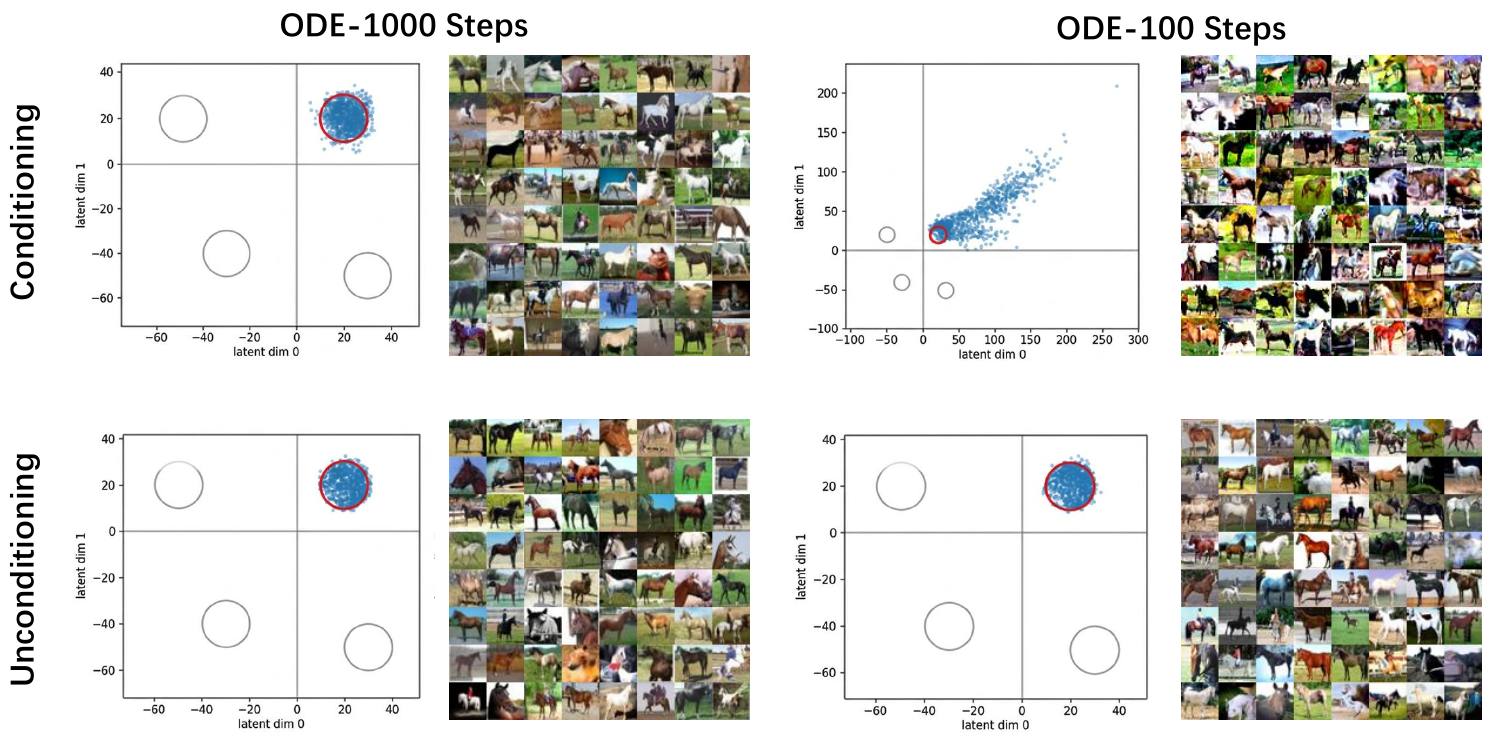}
  \end{minipage}
  \hfill
  \begin{minipage}[t]{0.28\linewidth}
    \vspace{0pt}
    \captionsetup{justification=raggedright,singlelinecheck=false}
    \caption{\textbf{Guided ODE sampling at two step budgets.} The top and bottom rows show noise-conditioned and noise-unconditioned samples for four-disk and CIFAR-10 horse guidance. The left and right panels use 1,000 and 100 steps, respectively.}
    \label{fig:large_step_control}
  \end{minipage}
\end{figure}

We test coarse guided updates with the ODE sampler of
\citet{zhou2026smoothing}. Guidance selects the first-quadrant disk in the
four-disk task and the \emph{horse} class on CIFAR-10. Both tasks use linearly
decaying guidance, with score-norm matching on CIFAR-10. At 1,000 steps, both
samplers produce target-consistent samples. At 100 steps, the conditioned
sampler shows visible degradation on both guided tasks while the unconditioned
sampler remains stable (Figure~\ref{fig:large_step_control}). The conditioned
sampler is also stable at 100 steps without guidance
(Appendix~\ref{app:control_implementation}).

We quantify the effect of step budget on the four-disk task using reverse
VE-SDE updates with $(\sigma_{\min},\sigma_{\max})=(0.01,50)$. For the
noise-conditioned (C) and noise-unconditioned (U) models, we select the
initial penalty weight $\lambda_0$ separately at each budget by maximizing
target success on 128 validation samples using the same 14 candidates.

\begin{table}[!htb]
\centering
\small
\setlength{\tabcolsep}{5pt} 
\caption{\textbf{Four-disk guidance across step budgets.} Entries give C / U. Success is the fraction in the target disk. Geometric error is the mean planar distance to the four-disk union. Target entropy is computed over target hits in 32 equal-area cells and divided by $\log 32$.}
\label{tab:revision_guidance}
\begin{tabular}{@{}rrrrr@{}}
\toprule
Main steps & Selected $\lambda_0$ & Success (\%) $\uparrow$ & Geometric error $\downarrow$ & Target entropy $\uparrow$ \\
\midrule
50 & {$0.01/0.025$} & {$62.2/100.0$} & {$1.15/0.00$} & {$0.95/0.87$} \\
100 & {$0.025/0.025$} & {$73.7/96.1$} & {$0.57/0.02$} & {$0.96/0.96$} \\
250 & {$0.025/0.025$} & {$90.4/97.9$} & {$0.16/0.02$} & {$0.95/0.97$} \\
500 & {$0.025/0.025$} & {$88.0/98.2$} & {$0.15/0.01$} & {$0.94/0.97$} \\
1,000 & {$0.025/0.025$} & {$87.5/97.1$} & {$0.16/0.02$} & {$0.95/0.98$} \\
\bottomrule
\end{tabular}
\end{table}

U attains higher target success at all five tuned budgets and at
all 45 matched settings (nine weights in $[0.025,1]$ per budget). At
$\lambda_0=0.1$, 50 main steps yield $99.7\%$ success for U and $4.4\%$ for C,
whose success reaches $87.5\%$ at 1,000 steps. U has lower geometric error
at every tuned budget. Its target entropy is lower at 50 steps and
comparable or higher at larger budgets. For C, stronger guidance at coarse steps
increases both the state--noise mismatch proxy and the normal-score error
(Appendix~\ref{subsec:revision_guidance}).

\subsection{Downstream classification under class imbalance}
\label{subsec:downstream_experiments}
\vspace{-0.05cm}

We train a binary ResNet-18 on an imbalanced CIFAR-10 split with 100 cats
and 1,000 dogs. We compare the imbalanced split, repetition of the 100 cats,
and augmentation with 900 synthetic cats from a diffusion or unconditioned
generator trained on the same 100 cats. A split with 1,000 real cats provides
a balanced reference. We evaluate on the CIFAR-10 cat--dog test set
(Appendix~\ref{app:downstream_augmentation}).

\begin{table}[!htb]
\centering
\captionsetup{skip=3pt}
\caption{\textbf{Downstream classification under minority-class scarcity.}
Test accuracy (\%, mean$\pm$std).}
\label{tab:downstream_imbalance}

\resizebox{0.65\linewidth}{!}{%
\begin{tabular}{@{}lccc@{}}
\toprule
Training condition & Cat & Dog & Bal. Acc. \\
\midrule
Biased real & 11.80$\pm$1.25 & 96.73$\pm$0.21 & 54.27$\pm$0.56 \\
Repeated cats (1000) & 13.63$\pm$0.76 & 97.13$\pm$0.46 & 55.38$\pm$0.41 \\
Diffusion aug. & 26.17$\pm$1.65 & 93.33$\pm$0.47 & 59.75$\pm$0.95 \\
Unconditioning aug. & \textbf{38.80$\pm$3.45} & 91.87$\pm$0.95 &
\textbf{65.33$\pm$2.03} \\
Balanced real & 79.17$\pm$0.76 & 79.63$\pm$1.40 & 79.40$\pm$1.01 \\
\bottomrule
\end{tabular}%
}
\end{table}

\vspace{-0.1cm}
We can see that repetition raises balanced accuracy by 1.11 percentage points.
Noise-unconditioned augmentation gains 5.58 points over diffusion
augmentation and 9.95 over repetition. Cat accuracy improves by 12.63 points
over diffusion augmentation. The ten-class experiment shows the same
pattern (Appendix~\ref{app:downstream_augmentation}).

Across three independent 100-cat subsets, with three independently
trained generators per method and subset, noise-unconditioned augmentation
gains $5.07$--$5.64$ points in balanced accuracy over diffusion augmentation
(mean $5.28$).

We draw 1,000 images per method from the generators used for the binary
comparison in Table~\ref{tab:downstream_imbalance}. Following
\citet[Appendix B.3.2]{zhou2026smoothing}, we flag outputs whose pixel-space
distance to the nearest training image is less than $2/3$ of the distance to
the second nearest. We also measure PRDC coverage
\citep{naeem2020reliable} against 1,000 real cats in pretrained ResNet-18
features, using fifth-neighbor radii.
The proximity proxy flags $47.1\%$ of diffusion outputs and
$23.6\%$ of unconditioned outputs. Reference coverage rises from $72.2\%$
to $89.1\%$ (Appendix~\ref{subsec:revision_augmentation}).

\section{Conclusion}
\label{sec:conclusion}
\vspace{-0.1cm}
Noise unconditioning offers a simple reinterpretation of multi-scale denoising:
rather than following a prescribed sequence of noise levels, generation can be
viewed as optimization on a single fixed landscape. From this perspective, we
establish two basic generative properties---fidelity and diversity---through
manifold convergence and finite-resolution regional coverage. We further show
that, in a constructive affine setting, positive smoothing can prevent exact
empirical-score optimization from collapsing onto the training set while
preserving diversity at the training resolution. The same fixed objective also
provides a natural interface for penalty-guided controllable generation.

Our experiments connect these mechanisms to practice: exact-score studies
support the predicted geometric and diversity behavior, while learned models
show robustness under large guided updates and downstream benefits. Overall,
noise-unconditioned denoising provides an elegant framework for understanding the generation process from an optimization perspective.
\label{maintext:end}

\clearpage

\subsection*{AI Use Statement}
In this work, we used generative AI tools for polishing and drafting a few sections of the appendix. We also used it for theory proof brainstorming, checking, and correcting. 
We have \textbf{not} used generative AI tools for retrieval and discovery, research ideation or execution, and generating synthetic datasets does not apply to this work.
Additionally, we used generative AI tools for partially coding E1-E4 and the downstream classification task.
We have reviewed all AI-assisted work. For example, LLM-generated code, special theory case design, and mathematical calculations were verified and tested for correctness. We take responsibility for the final content of this work,
including text, claims, or artifacts produced with the aid of generative AI.

\subsection*{Reproducibility Statement}
Complete assumptions and proofs for the theoretical results are provided in
Appendices~A--E and G--I. Appendix~F reports the numerical settings,
procedures, diagnostics, and limitations for E1--E4 and the learned circle checks. Appendices~H.4--H.5
provide the implementation details for controllable generation and downstream
classification, including sampling procedures, data splits, optimization
settings, and random seeds.

\bibliographystyle{iclr2027_conference}
\bibliography{iclr2027_conference}

\clearpage
\begingroup
  \small
  \setlength{\parskip}{4pt}
  \tableofcontents
\endgroup
\clearpage

\section*{Notation and Rough Guide}
\label{sec:notation_guide}

Appendix~\ref{app:convergence_details} proves convergence and geometric
localization. Appendices~\ref{app:diversity_details}--\ref{app:source_sink}
derive regional coverage for the same empirical trajectories.
Appendix~\ref{app:resolved_endpoint_proof} proves the non-memorizing endpoint
construction; Appendix~\ref{app:control_details} develops penalty-guided
generation and its experimental instantiations; and
Appendix~\ref{app:population_extension} gives the population counterparts.
All norms are Euclidean unless an intrinsic distance is specified. We write
$P_X$ for the distribution of a random variable $X$.

\begin{center}
\footnotesize
\renewcommand{\arraystretch}{0.8}
\begin{tabularx}{\linewidth}{@{}p{.29\linewidth}X@{}}
\toprule
Notation & Meaning\\
\midrule

\multicolumn{2}{@{}l}{\textbf{Data, objective, and score}}\\
$d$, $M$, $N$
& Ambient dimension, number of training centers, and number of noise scales.\\
$\mathcal S_M=\{\mu_j\}_{j=1}^M$, $\widehat Q_M$
& Training centers and empirical distribution
$\widehat Q_M=M^{-1}\sum_j\delta_{\mu_j}$.\\
$\sigma_i$, $\sigma_{\min}$, $\sigma_{\max}$
& Noise standard deviations and their lower and upper cutoffs.\\
$\widehat p_\sigma$
& Gaussian smoothing of the empirical distribution,
$\widehat Q_M*\mathcal N(0,\sigma^2I)$.\\
$p_{MN}$, $Z_{MN}$
& Unified empirical multi-scale density and its normalization constant.\\
$s_\theta$
& Learned state-only approximation to $\nabla\log p_{MN}$.\\
$w_{ij}$, $\widetilde w_{ij}$
& Mixture responsibilities and precision-normalized responsibilities.\\
$a_{ij}=(\mu_j-x)/\sigma_i^2$
& Component score vector.\\
$\beta$, $\bar\mu$
& Average precision and posterior barycenter, with
$\nabla\log p_{MN}=\beta(\bar\mu-x)$.\\
$h$, $\eta_n$
& Log-noise grid spacing and optimization step size.\\

\addlinespace
\multicolumn{2}{@{}l}{\textbf{Geometry and convergence}}\\
$\mathcal M$, $Q_{\mathcal M}$
& Reference data manifold and its reference probability distribution.\\
$m$, $c=d-m$, $\tau$
& Manifold dimension, codimension, and lower bound on reach.\\
$r(x)$, $z_x=\Pi_{\mathcal M}(x)$
& Distance to the manifold and nearest-point projection.\\
$u_x$
& Unit normal $(x-z_x)/r(x)$ inside the reach tube.\\
$K$, $\nu$, $\Omega$
& Initialization support, initialization distribution, and invariant
prior superlevel set.\\
$r_0$, $r_{\min}=\sigma_{\min}\sqrt c$
& Analysis-tube radius and lower-cutoff transition radius.\\
$C_{\mathrm{loc}}$
& Constant controlling the precision-normalized posterior second moment.\\
$\varepsilon_0$
& Normal localization bound
$C_{\mathrm{loc}}m\sigma_{\min}^2/(2\tau)$.\\
$\kappa_0$
& Curvature/contraction ratio
$C_{\mathrm{loc}}mr_0/(2c\tau)$.\\
$C_x$
& Covariance of component scores, with
$\nabla^2\log p_{MN}=C_x-\beta I$.\\
$r_{\mathrm{eff}}(C)$
& Effective rank $\operatorname{tr}(C)/\lambda_{\max}(C)$.\\
$\alpha_d$, $R_d$
& Trace and effective-rank constants in the off-tube instability condition.\\

\addlinespace
\multicolumn{2}{@{}l}{\textbf{Regional coverage}}\\
$r_{\mathrm{tr}}$, $r_{\mathrm{stop}}$
& Transition and stopping thresholds for normal distance.\\
$\tau_{\mathrm{tr}}$, $\tau_{\mathrm{stop}}$
& Corresponding first entrance times.\\
$X_{\mathrm{stop}}$, $Z_{\mathrm{tr}}$, $Z_{\mathrm{stop}}$
& Stopped ambient output and manifold projections at transition/stop.\\
$d_{\mathcal M}$, $U^{-\delta}$
& Intrinsic distance and $\delta$-safe interior of a manifold region.\\
$\delta_{\mathrm{stop}}$
& Bound on projected tangential motion during refinement.\\
$p_\sigma^Q$
& Gaussian smoothing
$Q_{\mathcal M}*\mathcal N(0,\sigma^2I)$.\\
$A_N$
& Positive flux normalization $h\sum_i\sigma_i^2$.\\
$\mathcal R_N$, $\epsilon_{\mathrm{Pois}}$
& Source--sink residual and its half-$L^1$ norm.\\
$\epsilon_{\mathrm{init}}$
& Initialization mismatch
$\operatorname{TV}(\nu,p_{\sigma_{\max}}^Q dx)$.\\
$Y$, $T_{\mathrm{mark}}$
& Auxiliary marked point and marking time on the original trajectory.\\
$\zeta$
& Failure probability used in the Gaussian placement argument.\\
$a_d$, $a_m$, $b_c$
& Gaussian norm thresholds used in the coverage proof.\\
$\delta_{\mathrm{noise}}$, $\delta_{\mathrm{cov}}$
& Projection error from fine-scale noise and total coverage resolution.\\
$Q_k(a)$
& $a$-quantile of $\chi_k^2$, distinct from $Q_{\mathcal M}$.\\
$W_1$
& Wasserstein-1 distance.\\

\addlinespace
\multicolumn{2}{@{}l}{\textbf{Finite-sample non-memorization}}\\
$\mathcal A=a_0+V$
& Affine data subspace and its tangent linear subspace.\\
$P_V$, $P_{V^\perp}$
& Orthogonal projections onto $V$ and its complement.\\
$\ell$, $L$
& Training-grid spacing and half-width parameter of the affine construction.\\
$D_L$, $C_{\mathbf k}$
& Reference box and sample-resolution output cores.\\
$h_{\mathrm{cov},M}$
& Geometric covering radius of the finite training set.\\
$s=\sigma_{\min}/\ell$
& Dimensionless lower bandwidth used in the affine construction and E3.\\
$h_M$
& Additional posterior-moment residual for finite-sample refinements.\\
$\varepsilon_{\mathcal M,h}$
& Residual-aware normal localization radius.\\

\addlinespace
\multicolumn{2}{@{}l}{\textbf{Population and control}}\\
$p^{\mathrm{pop}}$
& Unified population multi-scale density.\\
$\pi$
& Distribution or finite measure over noise scales in the population objective.\\
$E_y$, $\lambda$
& Differentiable task penalty and its guidance strength.\\
$F_y$
& Penalized objective $F_y=\log p_{MN}-\lambda E_y$.\\
$\Omega_y(x_0)$, $\mathcal R_y$
& Guided superlevel set and its relevant manifold-tube region.\\
$\delta_y$
& Outward normal guidance budget relative to the prior contraction scale.\\
$\varepsilon_y$
& Resulting guided normal-attraction radius.\\

\bottomrule
\end{tabularx}
\end{center}

\clearpage

\appendix
\numberwithin{equation}{section}

\section{Detailed Setup, Manifold Convergence, and Representation Geometry}
\label{app:convergence_details}
\label{sec:manifold_convergence}
This appendix gives the detailed version of Section~\ref{sec:gradient_ascent}.
The formulas here use the exact empirical score $\nabla\log p_{MN}$;
Appendix~\ref{app:population_extension} proves their population counterparts.
The manifold hypotheses locate its limiting points; they do not define or
identify a learned manifold. The full convergence statement is
Theorem~\ref{thm:manifold_convergence}, and optimizer localization is
Proposition~\ref{prop:optimizers_in_manifold_tube}.

\subsection{The empirical objective and posterior identities}
We use the training centers and positive finite scale schedule of Section~\ref{subsec:main_fixed_landscape}. All gradients are with respect to the sample $x$, and all vector norms are Euclidean. For reference,
\begin{align}
 \widehat Q_M&=\frac1M\sum_j\delta_{\mu_j},&
 \widehat p_\sigma(x)&=\frac1M\sum_j\mathcal N(x;\mu_j,\sigma^2I),
 \label{eq:empirical_measure_and_smoothing}\\
 p_{MN}(x)&=\frac{\sum_{j,i}\sigma_i^2\mathcal N(x;\mu_j,\sigma_i^2I)}{Z_{MN}},&
 Z_{MN}&=M\sum_i\sigma_i^2.
 \label{eq:unified_distribution_manifold}
\end{align}
The analysis conditions on this finite training set; no independent-sampling assumption is needed. The reference manifold $\mathcal M$ describes its geometry, and the distribution $Q_{\mathcal M}$ introduced for coverage specifies the desired regional probabilities.

For $1\le m<d$ and $c=d-m$, define
\begin{align}
 w_{ij}(x)
 &:=\frac{\sigma_i^2\mathcal N(x;\mu_j,\sigma_i^2I)}
 {\sum_{\ell,k}\sigma_k^2\mathcal N(x;\mu_\ell,\sigma_k^2I)},
 \label{eq:posterior_weights_manifold}\\
 a_{ij}(x)
 &:=\frac{\mu_j-x}{\sigma_i^2},
 \qquad
 \nabla\log p_{MN}(x)
 =\sum_{j,i}w_{ij}(x)a_{ij}(x),
 \label{eq:unified_score_manifold}\\
 \beta(x)
 &:=\sum_{j,i}\frac{w_{ij}(x)}{\sigma_i^2},
 \qquad
 \widetilde w_{ij}(x)
 :=\frac{w_{ij}(x)}{\sigma_i^2\beta(x)},
 \qquad
 \bar\mu(x)
 :=\sum_{j,i}\widetilde w_{ij}(x)\mu_j.
 \label{eq:precision_normalized_weights}
\end{align}

The precision-normalized weights equal the normalized Gaussian densities, whereas $w_{ij}$ includes the additional scale weight. Direct substitution gives
\begin{equation}
 \nabla\log p_{MN}(x)=\beta(x)(\bar\mu(x)-x),\qquad
 \sigma_{\max}^{-2}\le\beta(x)\le\sigma_{\min}^{-2}.
 \label{eq:score_barycenter_identity}
\end{equation}
In particular every critical point satisfies
\begin{equation}
 x=\bar\mu(x)\in\operatorname{conv}(\mathcal S_M).
 \label{eq:critical_points_convex_hull}
\end{equation}
We study
\begin{equation}
 \dot x=\nabla\log p_{MN}(x).
 \label{eq:unified_gradient_ascent}
\end{equation}
The time change $u(t)=\int_0^t\beta(x(v))\,dv$ is increasing and unbounded, and gives $dx/du=\bar\mu(x)-x$. Thus this normalized continuous flow has the same paths and hitting locations; its Euler discretization can have different numerical errors.

\subsection{Learning the gradient of the unified log density}
\label{app:learning_gradient}

The identity below retains the weights in the noise-unconditioned objective
of \citet{zhou2026smoothing}:
\begin{align}
 \mathcal L_u(\theta)
 &=\frac1{2MN}\sum_{j,i}\sigma_i^2
   \mathbb E_{X\sim\mathcal N(\mu_j,\sigma_i^2I)}
      \|s_\theta(X)-a_{ij}(X)\|^2,
 \label{eq:unconditioned_training_loss}\\
 &=\frac{Z_{MN}}{2MN}\mathbb E_{X\sim p_{MN}}
       \|s_\theta(X)-\nabla\log p_{MN}(X)\|^2+C.
 \label{eq:unconditioned_training_excess}
\end{align}

\begin{proof}[Proof of \eqref{eq:main_excess_loss}]
At fixed $x$, $w_{ij}(x)$ are the normalized loss coefficients and their
component-score average is $\nabla\log p_{MN}(x)$. The variance
decomposition gives
\[
 \sum_{j,i}w_{ij}\|s_\theta-a_{ij}\|^2
 =
 \|s_\theta-\nabla\log p_{MN}\|^2
 +
 \sum_{j,i}w_{ij}
 \|a_{ij}-\nabla\log p_{MN}\|^2.
\]
Multiplication by $Z_{MN}p_{MN}(x)/(2MN)$ and integration proves the
identity. The second term is independent of $\theta$ and integrable, being
bounded by the component-score second moment. The unrestricted regression
optimum is therefore $\nabla\log p_{MN}$, uniquely up to a set of
$p_{MN}$-measure zero.
\end{proof}

\paragraph{Empirical, population, and learned scores.}
The three fields play distinct roles. For a fixed training set,
\eqref{eq:unconditioned_training_excess} shows that the exact regression
target of the empirical DSM objective is the empirical unified score
$\nabla\log p_{MN}$. Replacing the empirical center average by the
population distribution gives the population score
$\nabla\log p^{\mathrm{pop}}$, defined in
Appendix~\ref{app:population_objective}. A learned network differs from the
population field through the decomposition
\[
 s_\theta(x)-\nabla\log p^{\mathrm{pop}}(x)
 =
 \bigl[s_\theta(x)-\nabla\log p_{MN}(x)\bigr]
 +
 \bigl[\nabla\log p_{MN}(x)-\nabla\log p^{\mathrm{pop}}(x)\bigr].
\]
The first term is the learned-to-empirical target error, while the second
is the empirical-to-population target discrepancy.
For the unrestricted empirical regression problem, the former is measured
exactly by the excess DSM objective:
\[
 \mathcal L_u(\theta)-\inf_s\mathcal L_u(s)
 =
 \frac{Z_{MN}}{2MN}
 \mathbb E_{X\sim p_{MN}}
 \|s_\theta(X)-\nabla\log p_{MN}(X)\|^2.
\]

Positive smoothing also gives useful intuition for the second term. At
fixed positive cutoffs, both $p_{MN}$ and $p^{\mathrm{pop}}$ are formed
from averages of the same smooth Gaussian kernels, differing in whether
the center distribution is empirical or population. Thus, when the
training centers are sampled from the population, concentration of the
smoothed densities and their spatial derivatives naturally suggests
closeness of the corresponding score fields on regions where the smoothed
density is not too small. This connection becomes more delicate toward
vanishing bandwidth, where individual training centers can dominate the
empirical field.

For the learned dynamics, a simple deterministic relation further explains
why score approximation is relevant. Along a learned flow
$\dot x=s_\theta(x)$,
\[
 \frac{d}{dt}\log p_{MN}(x(t))
 \ge
 \frac12\|\nabla\log p_{MN}(x(t))\|^2
 -
 \frac12\|s_\theta(x(t))-\nabla\log p_{MN}(x(t))\|^2.
\]
Hence sufficiently accurate local score approximation preserves the ascent
mechanism. Turning the average DSM error above into a trajectory-wise
guarantee, however, requires additional control of the learned field on the
states visited during generation.

This separation between an exact target score and its learned neural
approximation is also standard in score-based diffusion theory. For example,
existing convergence analyses assume an $L^2$-accurate score estimator and
derive sampling guarantees conditional on that error
\citep{chen2023sampling,lee2023convergence}. Our geometric results therefore analyze the exact
empirical and population target fields, while an end-to-end theory connecting
finite-data neural training, score approximation, and the resulting
autonomous trajectories is a complementary direction for future work.

\subsection{Global existence and convergence}

Let \(K\subset\mathbb R^d\) be nonempty and compact, and let the
initialization distribution \(\nu\) be supported on \(K\). Define
\begin{equation}
 p_K:=\min_{x\in K}p_{MN}(x)>0,\qquad
 \Omega:=\{x:p_{MN}(x)\ge p_K\}.
 \label{eq:compact_initialization_superlevel}
\end{equation}
The set \(\Omega\) is compact because the mixture is positive and
continuous, and \(p_{MN}(x)\to0\) as \(\|x\|\to\infty\).

\begin{lemma}[Global existence, finite length, and convergence]
\label{lemma:boundedness_and_convergence}
Every trajectory initialized in \(K\) exists for all \(t\ge0\), remains
in \(\Omega\), has finite length, and converges to a critical point
\(x_\infty\in\Omega\) of \(\log p_{MN}\).

\end{lemma}
\begin{proof}
\textbf{A compact invariant region.}
The positive finite Gaussian mixture and its logarithm are real analytic. Local existence and uniqueness follow from local Lipschitz continuity of the gradient. Along a solution,
\begin{equation}
 \frac d{dt}\log p_{MN}(x(t))=\|\nabla\log p_{MN}(x(t))\|^2\ge0.
 \label{eq:objective_monotonicity}
\end{equation}
Hence it stays in $\Omega$. The vector field is bounded there, so a finite maximal existence time would give a limit in $\Omega$ from which the solution could be continued. The solution is therefore global.

\textbf{Critical accumulation points.}
The objective increases to a finite value $\ell_\infty$, and
\begin{equation}
 \int_0^\infty\|\nabla\log p_{MN}(x(t))\|^2dt
 =\ell_\infty-\log p_{MN}(x(0))<\infty.
 \label{eq:finite_gradient_energy}
\end{equation}
The integrand has bounded derivative on $\Omega$, since both the gradient and Hessian are bounded. Uniform continuity and integrability imply that it tends to zero. Thus every accumulation point is critical with value $\ell_\infty$.

\textbf{Finite path length.}
The analytic gradient inequality \citep{absil2005convergence}, made uniform over the compact accumulation set by a finite cover, gives constants $C_L>0$ and $0<\vartheta<1$ such that, for all sufficiently large $t$,
\begin{equation}
 e(t)^\vartheta\le C_L\|\nabla\log p_{MN}(x(t))\|,
 \qquad e(t)=\ell_\infty-\log p_{MN}(x(t)).
 \label{eq:detailed_uniform_lojasiewicz}
\end{equation}
The trajectory eventually lies in that cover; otherwise it would have an accumulation point outside it. If $e$ reaches zero, the trajectory is stationary thereafter. Otherwise,
\[
 -\frac d{dt}e(t)^{1-\vartheta}
 \ge\frac{1-\vartheta}{C_L}\|\dot x(t)\|,
\]
and hence
\begin{equation}
 \int_T^\infty\|\dot x(t)\|dt
 \le\frac{C_L}{1-\vartheta}e(T)^{1-\vartheta}<\infty.
 \label{eq:detailed_finite_tail_length}
\end{equation}
The initial segment also has finite length. The trajectory is Cauchy and converges to one of its critical accumulation points.
\end{proof}

\begin{remark}[Scope of the flow result]
\label{remark:global_forward_completeness}
Applying the lemma with \(K=\{x_0\}\) proves forward completeness and
convergence for every \(x_0\in\mathbb R^d\). Denote the flow by
\(\phi_t(x_0)=x(t;x_0)\). No manifold assumption is needed for this
fact. Geometric assumptions below locate the limiting point; they are
not needed to establish the existence of a limit.
\end{remark}

\paragraph{Gaussian initialization and confidence thresholds.}
For $G_k\sim\mathcal N(0,I_k)$, let
\begin{equation}
 Q_k(a)=\inf\{q:\mathbb P(\|G_k\|^2\le q)\ge a\},\qquad 0<a<1,
 \label{eq:chi_squared_quantile_definition}
\end{equation}
be the $a$-quantile of $\chi_k^2$. These scalar quantiles are distinct from the reference distribution $Q_{\mathcal M}$. For $0<\delta_{\mathrm{init}}<1$, the shell
\begin{equation}
 \Omega_{\mathrm{init}}=\{x:
 \sigma_{\max}\sqrt{Q_d(\delta_{\mathrm{init}}/2)}\le\|x\|
 \le\sigma_{\max}\sqrt{Q_d(1-\delta_{\mathrm{init}}/2)}\}
 \label{eq:initialization_shell}
\end{equation}
has probability $1-\delta_{\mathrm{init}}$ under the centered Gaussian. The compact-support theorem applies to
\begin{equation}
 \nu=\mathcal N(0,\sigma_{\max}^2I_d)\mid\Omega_{\mathrm{init}},\qquad
 K=\Omega_{\mathrm{init}}.
 \label{eq:truncated_gaussian_initialization}
\end{equation}
A probability-one conclusion for this truncated initialization gives probability at least $1-\delta_{\mathrm{init}}$ for the untruncated Gaussian.

\subsection{Local geometry and distance-matched scales}
\label{sec:dominant_precision_scale}

\begin{definition}[Reach]
\label{def:reach}
For a closed set \(A\subset\mathbb R^d\), its reach is the supremum
of radii \(r\ge0\) such that every point at distance less than \(r\)
from \(A\) has a unique nearest point in \(A\).
\end{definition}

\begin{assumption}[Local manifold geometry and available scales]
\label{assumption:local_manifold_geometry}
The manifold \(\mathcal M\) is compact, embedded, \(C^2\), and without
boundary, with \(\operatorname{reach}(\mathcal M)\ge\tau>0\). Choose
\begin{equation}
 0<r_0<\tau,\qquad r_0\le\sigma_{\max}\sqrt c.
 \label{eq:local_tube_radius}
\end{equation}
Write \(r(x)=\operatorname{dist}(x,\mathcal M)\). The unique nearest
point \(z_x=\Pi_{\mathcal M}(x)\) is defined on the entire reach
tube \(r(x)<\tau\); the smaller \(r_0\)-tube is where we impose
posterior estimates. Write \(T_z\mathcal M\) for the tangent space,
\(N_z\mathcal M=(T_z\mathcal M)^\perp\) for the normal space, and
\(P_{T_z\mathcal M},P_{N_z\mathcal M}\) for their orthogonal
projectors. For \(0<r(x)<\tau\), let
\(u_x=(x-z_x)/r(x)\): this is the unit normal pointing from the
nearest manifold point toward \(x\), not a globally chosen orientation.
\end{assumption}

Reach controls both bending and the approach of distinct branches.
The \(r_0\)-tube is the region \(r(x)<r_0\). The smaller region
\(\varepsilon_{\mathcal M}<r(x)<r_0\), after its inner radius is
defined below, will be called the intermediate contraction region.
Both are restricted to the invariant region \(\Omega\) whenever a
posterior estimate is used. Within the reach tube the nearest-point
projection is \(C^1\), and \(\nabla r(x)=u_x\) away from \(\mathcal M\)
\citep[Theorem 2]{leobacher2021existence}. These facts allow us to
separate motion toward the manifold from motion along it.

\paragraph{The distance-matched scale.}
In a locally flat population patch, write $\mu=z_x+Ut$ with $U^\top U=I_m$ and $U^\top(x-z_x)=0$. Integrating the tangent Gaussian gives
\begin{equation}
 \int_{\mathcal M}\mathcal N(x;\mu,\sigma^2I)q_{\mathcal M}(\mu)
        d\operatorname{vol}_{\mathcal M}(\mu)
 \approx q_{\mathcal M}(z_x)(2\pi\sigma^2)^{-c/2}e^{-r^2/(2\sigma^2)}.
 \label{eq:local_population_scale_contribution}
\end{equation}
The raw mixture and precision-normalized posterior therefore have profiles
\begin{equation}
 H_r(\sigma)=\sigma^{2-c}e^{-r^2/(2\sigma^2)},\qquad
 G_r(\sigma)=\sigma^{-c}e^{-r^2/(2\sigma^2)}.
 \label{eq:raw_local_scale_profile}
\end{equation}
Their logarithmic derivatives are respectively $[r^2-(c-2)\sigma^2]/\sigma^3$ and $[r^2-c\sigma^2]/\sigma^3$. Thus the precision profile peaks at
\begin{equation}
 \sigma_{\mathrm{match}}(r)=r/\sqrt c,\qquad r_{\min}=\sigma_{\min}\sqrt c,
 \label{eq:precision_matched_scale}
\end{equation}
with lower-cutoff saturation when $r\le r_{\min}$. A mode of this scale profile motivates, but does not prove, the finite-posterior moment bound below. Appendix~\ref{subsec:flat_model} computes both the mode and the moment for flat reference models.

\begin{assumption}[Scale-dependent posterior localization]
\label{assumption:posterior_localization}
There is a dimensionless constant \(C_{\mathrm{loc}}>0\) such that
\begin{equation}
 \sum_{j,i}\widetilde w_{ij}(x)\|\mu_j-z_x\|^2
 \le C_{\mathrm{loc}}m\max\{\sigma_{\min}^2,r(x)^2/c\},
 \qquad x\in\Omega,\quad 0<r(x)<r_0.
 \label{eq:posterior_localization}
\end{equation}
\end{assumption}
The substantive requirement is a controlled constant throughout the
relevant tube and scale range. For fixed finite data and
\(\sigma_{\min}>0\), compactness always allows the uninformative value
\(\operatorname{diam}(\mathcal M)^2/(m\sigma_{\min}^2)\). The flat
model in Appendix~\ref{subsec:flat_model} instead gives an order-one
constant in a specified high-codimension regime.

\subsection{From posterior localization to normal attraction}

\begin{lemma}[Positive-reach normal inequality]
\label{lemma:positive_reach_normal_inequality}
Under Assumption~\ref{assumption:local_manifold_geometry}, for
\(z,\mu\in\mathcal M\) and every unit normal \(u\in N_z\mathcal M\),
\begin{equation}
 \langle u,\mu-z\rangle\le\frac{\|\mu-z\|^2}{2\tau}.
 \label{eq:positive_reach_normal_inequality}
\end{equation}
\end{lemma}
\begin{proof}
For $0<a<\tau$, the normal-tube property gives $\Pi_{\mathcal M}(z+au)=z$ \citep{leobacher2021existence}. Hence
\[
 a^2\le\|z+au-\mu\|^2
 =a^2-2a\langle u,\mu-z\rangle+\|\mu-z\|^2.
\]
Rearrange and let $a\uparrow\tau$.
\end{proof}

\begin{lemma}[Posterior alignment]
\label{lemma:posterior_alignment}
Under Assumptions~\ref{assumption:local_manifold_geometry}
and~\ref{assumption:posterior_localization},
\begin{equation}
 \langle u_x,\bar\mu(x)-z_x\rangle
 \le\frac{C_{\mathrm{loc}}m}{2\tau}
       \max\{\sigma_{\min}^2,r(x)^2/c\},
 \qquad x\in\Omega,\quad 0<r(x)<r_0.
 \label{eq:posterior_alignment}
\end{equation}
\end{lemma}
\begin{proof}
Average \eqref{eq:positive_reach_normal_inequality} over the nonnegative precision-normalized weights, then apply \eqref{eq:posterior_localization}:
\[
 \langle u_x,\bar\mu(x)-z_x\rangle
 \le\frac1{2\tau}\sum_{j,i}\widetilde w_{ij}(x)\|\mu_j-z_x\|^2
 \le\frac{C_{\mathrm{loc}}m}{2\tau}\max\{\sigma_{\min}^2,r(x)^2/c\}.
\]
\end{proof}

Define the geometric constants:
\begin{equation}
\begin{split}
 \varepsilon_0&:=\frac{C_{\mathrm{loc}}m}{2\tau}\sigma_{\min}^2,\qquad
 \kappa_0:=\frac{C_{\mathrm{loc}}mr_0}{2c\tau},\\
 \varepsilon_{\mathcal M}&:=\min\{\varepsilon_0,r_{\min}\}.
\end{split}
 \label{eq:C_manifold_definition}
\end{equation}

\begin{assumption}[Contraction gap and consistent tube]
\label{assumption:local_contraction_gap}
We require
\begin{align}
 \kappa_0&<1,
 \label{eq:kappa0_less_than_one}\\
 \varepsilon_{\mathcal M}&<r_0.
 \label{eq:tube_radius_consistency}
\end{align}
\end{assumption}
The gap asks that the curvature-induced normal bias be weaker than
the direct inward displacement. It is favored by large reach and
small \(m/c\), provided \(C_{\mathrm{loc}}\) is controlled.

The standing conditions also imply
\(\varepsilon_{\mathcal M}=\varepsilon_0<r_{\min}\). Indeed,
\[
 \varepsilon_0
 =\frac{C_{\mathrm{loc}}m}{2\tau}\sigma_{\min}^2
 =\kappa_0\frac{r_{\min}^2}{r_0}.
\]

If \(\varepsilon_0\ge r_{\min}\), division by \(r_{\min}>0\)
would give \(r_{\min}\ge r_0/\kappa_0>r_0\), and then
\(\varepsilon_{\mathcal M}=r_{\min}>r_0\), a contradiction.
The minimum notation records the two radial regimes, but creates no
extra admissible regime. The quantity
\(\varepsilon_0\) is an upper bound on endpoint distance, not a
lower bound on the actual error or a rate of convergence in time.

\begin{lemma}[Strict inward score]
\label{lemma:inward_score}
Under the three local assumptions, along the flow in \(\Omega\) whenever
\(0<r(t)<r_0\),
\begin{equation}
 \dot r(t)\le
 \begin{cases}
 -\beta(x(t))(r(t)-\varepsilon_0),&0<r(t)\le r_{\min},\\
 -\beta(x(t))(1-\kappa_0)r(t),&r_{\min}<r(t)<r_0.
 \end{cases}
 \label{eq:piecewise_radial_distance_dynamics}
\end{equation}
Here and below \(r(t):=r(x(t))\). In particular,
\begin{equation}
 \langle x-z_x,\nabla\log p_{MN}(x)\rangle<0
 \quad\text{for }x\in\Omega,\quad
       \varepsilon_{\mathcal M}<r(x)<r_0.
 \label{eq:strict_inward_score}
\end{equation}

There is no critical point in this intermediate contraction region. For every
\(\rho\in(\varepsilon_{\mathcal M},r_0)\), a trajectory in \(\Omega\)
that enters \(\{r\le\rho\}\) cannot subsequently leave it.
\end{lemma}
\begin{proof}
The distance gradient is $u_x$ in the reach tube, so
\begin{equation}
 \dot r=\beta(x)\bigl(\langle u_x,\bar\mu(x)-z_x\rangle-r\bigr).
 \label{eq:radial_distance_dynamics}
\end{equation}
For $r\le r_{\min}$, posterior alignment bounds the first term by $\varepsilon_0$. For $r>r_{\min}$, it bounds it by
$C_{\mathrm{loc}}mr^2/(2c\tau)=\kappa_0r^2/r_0<\kappa_0r$.
These are the two radial inequalities. As $\varepsilon_{\mathcal M}=\varepsilon_0<r_{\min}$, both give $\dot r<0$ for $\varepsilon_{\mathcal M}<r<r_0$, excluding critical points there. On $r=\rho$ the derivative points strictly inward. At a hypothetical first exit from $r\le\rho$, continuity would instead require a nonnegative outward derivative. This contradiction proves invariance.
\end{proof}

\subsection{Off-tube instability and saddle avoidance}

Local contraction leaves one global question: could ascent stop at an
attracting critical point outside the tube? Define
\begin{equation}
 C_x:=\operatorname{Cov}_{w(x)}(a_{ij}(x)).
 \label{eq:component_score_vector}
\end{equation}

Explicitly, since \(\sum_{j,i}w_{ij}(x)a_{ij}(x)=\nabla\log p_{MN}(x)\),

\[
 C_x=\sum_{j,i}w_{ij}(x)
       (a_{ij}(x)-\nabla\log p_{MN}(x))(a_{ij}(x)-\nabla\log p_{MN}(x))^\top\succeq0.
\]

For a nonzero positive semidefinite matrix, use
\(r_{\mathrm{eff}}(C)=\operatorname{tr}(C)/\lambda_{\max}(C)\).

\begin{assumption}[Off-tube shell-rank geometry]
\label{assumption:off_tube_geometry}
There are \(\alpha_d\in(0,1]\) and \(R_d>0\), with
\(R_d<\alpha_dd\), such that every critical point \(x\in\Omega\)
with \(r(x)\ge r_0\) satisfies
\begin{equation}
 \operatorname{tr}(C_x)\ge\alpha_dd\,\beta(x),\qquad
 r_{\mathrm{eff}}(C_x)\le R_d.
 \label{eq:shell_trace_assumption}
\end{equation}
\end{assumption}
The trace bound ensures \(C_x\ne0\). It supplies enough score
variation; the rank bound concentrates that variation in sufficiently
few directions to create an ascent-unstable direction.

\begin{lemma}[Hessian decomposition of the unified log density]
\label{lemma:hessian_decomposition}
For every \(x\),
\begin{equation}
 \nabla^2\log p_{MN}(x)=C_x-\beta(x)I.
 \label{eq:hessian_decomposition}
\end{equation}

\end{lemma}
\begin{proof}
Differentiating the normalized component weights gives
$\nabla w_{ij}=w_{ij}(a_{ij}-\nabla\log p_{MN})$, while
$Da_{ij}=-\sigma_i^{-2}I$. Therefore
\[
 \nabla^2\log p_{MN}
 =\sum_{j,i}w_{ij}a_{ij}a_{ij}^{\top}
   -(\nabla\log p_{MN})(\nabla\log p_{MN})^{\top}-\beta I
 =C_x-\beta I.
\]
\end{proof}

\begin{lemma}[Instability of off-tube critical points]
\label{lemma:off_tube_instability}
Under Assumption~\ref{assumption:off_tube_geometry}, every off-tube
critical point in \(\Omega\) satisfies
\begin{equation}
 \lambda_{\max}(\nabla^2\log p_{MN}(x))
 \ge\left(\frac{\alpha_dd}{R_d}-1\right)\beta(x)>0.
 \label{eq:positive_hessian}
\end{equation}

It is therefore not an ambient local maximum of \(\log p_{MN}\).

\end{lemma}
\begin{proof}
The assumed trace is positive. By the effective-rank bound,
\[
 \lambda_{\max}(C_x)=\frac{\operatorname{tr}C_x}{r_{\mathrm{eff}}(C_x)}
 \ge\frac{\alpha_dd}{R_d}\beta(x).
\]
Subtracting $\beta I$ yields \eqref{eq:positive_hessian}. A positive Hessian eigenvalue at a critical point gives both an exponentially growing linearized direction for ascent and a direction of positive second variation. The point cannot be a local maximum.
\end{proof}

\paragraph{A sufficient trace mechanism.}
At a critical point, $\operatorname{tr}C_x=\sum_{j,i}w_{ij}\|x-\mu_j\|^2/\sigma_i^4$. If an active set carries at least $(1-\zeta_d)\beta(x)$ precision mass and satisfies $\|x-\mu_j\|^2\ge(1-\delta_d)d\sigma_i^2$, then
$\operatorname{tr}C_x\ge(1-\zeta_d)(1-\delta_d)d\beta(x)$. Effective rank is a separate requirement. The sphere model verifies both ingredients at its off-tube center.

\paragraph{Why unstable endpoints have zero probability.}
A positive eigenvalue of \(\nabla^2\log p_{MN}\) does not exclude initialization on a
stable set. We therefore use concrete sufficient conditions for the
exact flow: nondegenerate off-tube critical points and an absolutely
continuous initialization. The following lemma proves the required
avoidance, rather than assuming the behavior of the limiting point.
The limit map is Borel measurable because
\(x_\infty(x_0)=\lim_{n\to\infty}\phi_n(x_0)\) and each
\(\phi_n\) is continuous, so all endpoint events below are measurable.

\begin{lemma}[Almost-sure avoidance of off-tube critical points]
\label{lemma:sufficient_saddle_avoidance}
Suppose Assumption~\ref{assumption:off_tube_geometry} holds, every
point in
\begin{equation}
 \mathcal C_{\mathrm{off}}:=\{x\in\Omega:\nabla\log p_{MN}(x)=0,\ r(x)\ge r_0\}
 \label{eq:off_tube_critical_set}
\end{equation}

is hyperbolic, and \(\nu\) is absolutely continuous with respect to
ambient Lebesgue measure \(\operatorname{Leb}_d\), written
\(\nu\ll\operatorname{Leb}_d\): every zero-volume set has
\(\nu\)-probability zero. Then
\begin{equation}
 \mathbb P_{X_0\sim\nu}[r(x_\infty)\ge r_0]=0.
 \label{eq:saddle_escaping_property}
\end{equation}
For this gradient flow, hyperbolicity means that \(\nabla^2\log p_{MN}\) at the
critical point has no zero eigenvalue. The condition is imposed only
at off-tube critical points, not at the limiting points inside the tube.

\end{lemma}
\begin{proof}
\textbf{Finitely many unstable equilibria.}
The off-tube critical set is closed in compact $\Omega$. A nonsingular Hessian makes each point isolated by the inverse function theorem. A compact set of such isolated critical points is finite. Each has a positive Hessian eigenvalue by Lemma~\ref{lemma:off_tube_instability}, so its local stable manifold has dimension less than $d$ and zero ambient volume \citep{hirsch1970invariant}.

\textbf{Pulling stable sets back to initialization.}
The derivative of the finite-time flow satisfies
$\dot J=\nabla^2\log p_{MN}(\phi_t(x_0))J$, $J(0)=I$, and hence
\[
 \det D\phi_t(x_0)
 =\exp\!\left(\int_0^t\Delta\log p_{MN}(\phi_u(x_0))du\right)>0.
\]
Uniqueness makes $\phi_t$ injective, and it is a diffeomorphism onto its image with locally Lipschitz inverse. A trajectory converging to an equilibrium eventually stays in its local stable neighborhood. Thus
\begin{equation}
 W^s(x_*)\subseteq\bigcup_{n\ge0}\phi_n^{-1}(W^s_{\mathrm{loc}}(x_*)).
 \label{eq:global_stable_set_union}
\end{equation}
Each preimage is null, using a countable cover by neighborhoods on which the inverse is Lipschitz. The countable union over $n$ and finite union over equilibria remain null. Absolute continuity of $\nu$ gives the conclusion.
\end{proof}
Here, hyperbolicity provides a convenient sufficient condition for almost-sure avoidance of off-tube equilibria.

\subsection{Manifold convergence and optimizer localization}

\begin{theorem}[Almost-sure manifold-tube convergence]
\label{thm:manifold_convergence}
Let \(K,\nu,\Omega\) be as in
\eqref{eq:compact_initialization_superlevel}. Suppose Assumptions
\ref{assumption:local_manifold_geometry},
\ref{assumption:posterior_localization},
\ref{assumption:local_contraction_gap},
and \ref{assumption:off_tube_geometry} hold. In addition, suppose all
off-tube critical points in \(\Omega\) are hyperbolic and
\(\nu\ll\operatorname{Leb}_d\). Then every trajectory from \(K\) converges to a critical point,
\[
 x(t)\longrightarrow x_\infty\in\Omega,
 \qquad \nabla\log p_{MN}(x_\infty)=0,
\]

and its limit satisfies
\begin{equation}
 \operatorname{dist}(x_\infty,\mathcal M)
 \le\varepsilon_{\mathcal M}
 =\min\left\{\frac{C_{\mathrm{loc}}m}{2\tau}\sigma_{\min}^2,
                      \sigma_{\min}\sqrt c\right\}
 \qquad\nu\text{-almost surely}.
 \label{eq:final_manifold_distance}
\end{equation}

Under the standing consistency conditions,
\(\varepsilon_{\mathcal M}=\varepsilon_0\), so the distance bound is
\(C_{\mathrm{loc}}m\sigma_{\min}^2/(2\tau)\).
The conclusion is relative to this chosen reference manifold; it does
not identify it uniquely or assert convergence to a global maximizer.

\end{theorem}
\begin{proof}
Every trajectory converges to a critical point by Lemma~\ref{lemma:boundedness_and_convergence}. Lemma~\ref{lemma:inward_score} excludes a critical limit with $\varepsilon_{\mathcal M}<r<r_0$, and Lemma~\ref{lemma:sufficient_saddle_avoidance} excludes $r\ge r_0$ almost surely. Thus $r(x_\infty)\le\varepsilon_{\mathcal M}$. The consistency calculation above gives $\varepsilon_{\mathcal M}=\varepsilon_0$.
\end{proof}

\begin{proposition}[Location of local and global maximizers]
\label{prop:optimizers_in_manifold_tube}
Under the local and off-tube geometry assumptions of the theorem,
without hyperbolicity or an absolute-continuity requirement on initialization,
every ambient local maximizer
\(x^\star\in\Omega\) of \(\log p_{MN}\) satisfies
\(r(x^\star)\le\varepsilon_{\mathcal M}\). Moreover,

\begin{equation}
 \varnothing\ne\operatorname*{argmax}_{x\in\mathbb R^d}\log p_{MN}(x)
 \subseteq\{x:r(x)\le\varepsilon_{\mathcal M}\}.
 \label{eq:global_optimizers_manifold_localized}
\end{equation}

\end{proposition}
\begin{proof}
A local maximum has zero gradient and a negative semidefinite Hessian. The inward-score inequality rules out $\varepsilon_{\mathcal M}<r<r_0$; off-tube instability rules out $r\ge r_0$. A global maximum exists because $p_{MN}$ is positive and tends to zero at infinity. Its value is at least $p_K$, so it belongs to $\Omega$ and satisfies the same bound.
\end{proof}
This proposition localizes the maximizers of the fixed objective;
trajectory-level convergence is addressed separately by
Theorem~\ref{thm:manifold_convergence}.

\begin{remark}[Eventual entry, Gaussian starts, and sampling resolution]
\label{remark:eventual_tube_entry}
For every \(\rho\in(\varepsilon_{\mathcal M},r_0)\), almost every
trajectory reaches \(\{r\le\rho\}\) in finite time and then remains
there. This follows from convergence to a strictly smaller radius and
the inward boundary condition. If the theorem holds under the
conditional Gaussian initialization
\eqref{eq:truncated_gaussian_initialization}, the untruncated Gaussian
satisfies
\begin{equation}
 \mathbb P[r(x_\infty)\le\varepsilon_{\mathcal M}]
 \ge1-\delta_{\mathrm{init}}.
 \label{eq:untruncated_gaussian_probability_bound}
\end{equation}
Indeed, let \(E=\{X_0\in\Omega_{\mathrm{init}}\}\) and
\(A=\{r(x_\infty)\le\varepsilon_{\mathcal M}\}\). The unconditional
flow has a measurable limit for every initialization, and
\(\mathbb P(E)=1-\delta_{\mathrm{init}}\). Applying the theorem to
\(P_{X_0\mid E}\) gives \(\mathbb P(A\mid E)=1\).
Thus
\[
 \mathbb P(A)\ge\mathbb P(A\cap E)
 =\mathbb P(A\mid E)\mathbb P(E)=1-\delta_{\mathrm{init}}.
\]

For fixed finite data, a controlled posterior-localization constant
need not persist as \(\sigma_{\min}\to0\). Sample-associated modes
may develop under separation and smallest-scale dominance. Such modes
can still be close to \(\mathcal M\): memorization concerns allocation
along the manifold, not a contradiction to normal localization.
\end{remark}

\begin{corollary}[Vanishing tube radius]
\label{cor:vanishing_manifold_tube_radius}
For a family of finite mixtures with centers on a fixed
\(\mathcal M\), with fixed \(m,c,\tau,r_0\), suppose the theorem's
assumptions hold for each
member, on its corresponding \(\Omega\) and under its initialization
distribution. Then, for each member,
\begin{equation}
 r(x_\infty^{(\sigma_{\min})})
 \le\frac{C_{\mathrm{loc}}(\sigma_{\min})m}{2\tau}\sigma_{\min}^2
 <\frac c{r_0}\sigma_{\min}^2
 \quad\text{almost surely}.
 \label{eq:universal_quadratic_tube_bound}
\end{equation}

\end{corollary}
\begin{proof}
For each model the contraction gap gives
$C_{\mathrm{loc}}(\sigma_{\min})m/(2\tau)<c/r_0$.
Multiply by $\sigma_{\min}^2$ and use Theorem~\ref{thm:manifold_convergence}. The probability-one conclusion is for each model separately.
\end{proof}

Finite empirical data can require a nonzero posterior-moment residual.
Appendix~\ref{app:finite_sample_refinements} propagates that residual
through both the normal and tangential bounds. The zero-residual
convergence result above does not assert uniform validity at arbitrarily
small noise for a fixed finite cloud.

\subsection{Affine representations and the meaning of larger reach}

\begin{proposition}[Exact affine normal convergence]
\label{prop:affine_subspace_exact_normal_convergence}
Suppose the finite set of training centers lies in a closed affine subspace
\(\mathcal A=a_0+V\). Write \(z_x=\Pi_{\mathcal A}(x)\),
\(r_{\mathcal A}(x)=\|x-z_x\|\), and let \(P_{V^\perp}\) be the
orthogonal projector onto \(V^\perp\). Then

\begin{equation}
 P_{V^\perp}\nabla\log p_{MN}(x)=-\beta(x)(x-z_x),\qquad
 r_{\mathcal A}(x(t))\le r_{\mathcal A}(x(0))e^{-t/\sigma_{\max}^2}.
 \label{eq:affine_exact_normal_score}
\end{equation}

Every critical point, and hence every limiting point, belongs to
\(\mathcal A\).
\end{proposition}
\begin{proof}
Since every training center lies in $\mathcal A$, so does $\bar\mu(x)$. With
$n(t)=P_{V^\perp}(x(t)-a_0)=x(t)-\Pi_{\mathcal A}(x(t))$, the barycenter identity gives $\dot n=-\beta(x(t))n$. Consequently
\[
 \|n(t)\|=\|n(0)\|e^{-\int_0^t\beta(x(u))du}
 \le\|n(0)\|e^{-t/\sigma_{\max}^2}.
\]
This identity also holds for $n(0)=0$. At any critical point it forces $n=0$. Global finite-mixture convergence then puts every endpoint in $\mathcal A$.
\end{proof}

The endpoint is also in \(\operatorname{conv}(\mathcal S_M)\), by
\eqref{eq:critical_points_convex_hull}. This normal result alone says
nothing about allocation within the affine manifold. After proving
regional coverage, Corollary~\ref{cor:sample_resolution_nonmemorization}
returns to this geometry and proves training-set separation together with
coverage at the training resolution for a verified finite-cloud family.

\begin{lemma}[Reach controls the local Gaussian geometry]
\label{lemma:reach_gaussian_sandwich}
For \(x=z_x+r u_x\) with \(0<r<\tau\) and every
\(\mu\in\mathcal M\),
\begin{equation}
\begin{split}
 r^2+(1-r/\tau)\|\mu-z_x\|^2
 &\le \|x-\mu\|^2\\
 &\le r^2+(1+r/\tau)\|\mu-z_x\|^2.
\end{split}
 \label{eq:reach_gaussian_distance_sandwich}
\end{equation}
\end{lemma}
\begin{proof}
Applying the reach inequality to both $u_x$ and $-u_x$ gives
$|\langle u_x,\mu-z_x\rangle|\le\|\mu-z_x\|^2/(2\tau)$.
Insert this into $\|x-\mu\|^2=r^2+\|\mu-z_x\|^2-2r\langle u_x,\mu-z_x\rangle$.
\end{proof}

\begin{remark}[Geometry and representation]
\label{remark:reach_representation_design}
The relevant quantities are $C_{\mathrm{loc}}/\tau$, $r_0/\tau$ and the operating noise scales relative to reach. The distance sandwich connects curvature to Gaussian weights, but sample coverage and the scale schedule also affect their normalized moment. In the sphere model a valid comparison is $C_{\mathrm{loc}}=\tau C_{\mathrm{flat}}/(\tau-r_0)$, giving $\varepsilon_0=C_{\mathrm{flat}}m\sigma_{\min}^2/[2(\tau-r_0)]$. These bounds favor locally near-affine, well-covered representations with separated branches. Uniform rescaling changes all lengths equally; nonlinear encoders additionally change the noise metric. Representation comparisons therefore require consistent normalization and a decoder-aware data interpretation.

\end{remark}

\subsection{Learned score perturbations and stopping}
\label{subsec:learned_perturbation}
Let $p=p_{MN}$, $b=\nabla\log p$, and $s_\theta=b+e$. The same argument applies to the population target. Within the tube, write $n=(x-z_x)/r(x)$ for the outward unit normal and $P_T$ for projection onto $T_{z_x}\mathcal M$. Suppose
\[
 [\langle n,e\rangle]_+\le\beta(x)\delta_N,\qquad
 \|P_Te\|\le\beta(x)\delta_T .
\]
\par\begingroup
Adding the normal error to \eqref{eq:main_radial_contraction} bounds $\dot r$ by
$-\beta(r-\varepsilon_0-\delta_N)$ for $r\le r_{\min}$ and
$-\beta((1-\kappa_0)r-\delta_N)$ for $r>r_{\min}$. Both are strictly inward above
\[
 \varepsilon_e=\max\{\varepsilon_0+\delta_N,\ \delta_N/(1-\kappa_0)\}.
\]
If $\varepsilon_e<r_{\mathrm{stop}}<r_0$, these bounds ensure first entry for trajectories already in the tube.
\par\endgroup

For the retention calculation, $D\Pi_{\mathcal M}$ annihilates the normal component of $e$, so the projected-speed bound is at most
\[
 L_\Pi\beta(x)\bigl(\sqrt{C_{\mathrm{loc}}m}\max\{\sigma_{\min},r/\sqrt c\}+\delta_T\bigr).
\]
Dividing this speed by the corresponding inward margin and integrating over $r$ gives the perturbed boundary erosion in Proposition~\ref{prop:early_stopping_retention}. These local controls address the approach to the stopping tube. Entry from outside it additionally requires a global condition or a separate placement argument.

\par\begingroup
\paragraph{Probability transport and ascent.}
Replacing $b$ by $b+e$ adds $A_N\operatorname{div}(pe)$ to the source--sink residual in \eqref{eq:finite_source_sink_identity}. For a complete $C^1$ flow with bounded forward trajectories and an integrable added divergence of zero total mass, the mark construction extends with an additional TV budget
\[
 \varepsilon_{\mathrm{flux}}=\tfrac12 A_N\|\operatorname{div}(pe)\|_{L^1}
   =\tfrac12 A_N\mathbb E_p\bigl|\operatorname{div}e+b^\top e\bigr|.
\]
\par\endgroup
The occupation-density proof uses the positive and negative parts of the new total residual as in Lemma~\ref{lemma:source_sink_mark}. Quantitative coverage by that route thus needs control of the added flux as well as the geometry. Alternatively, direct verification of inward motion, tangential drift, and prior placement yields a coverage bound without estimating the neural flux, as in Appendix~\ref{sec:learned_circle}.
Where $\|e\|\le\eta\|b\|$ with $\eta<1$, the learned flow satisfies
$\frac{d}{dt}\log p(x(t))=\|b\|^2+b^\top e\ge(1-\eta)\|b\|^2$.
This is an ascent condition on the region where the relative-error bound holds.

\par\begingroup
\paragraph{Stopping.}
If a continuous distance estimate obeys $|\widehat r-r|\le\delta_r$, its first downcrossing of $r_{\mathrm{stop}}-\delta_r$ from above returns a state in the true interval $[r_{\mathrm{stop}}-2\delta_r,r_{\mathrm{stop}}]$. Requiring $r_{\mathrm{stop}}-2\delta_r>\varepsilon_e$ leaves a strict inward margin up to that crossing. Checking at discrete times also requires an overshoot allowance.
\par\endgroup
For a reverse VE-SDE, a practical alternative is to select a positive terminal noise level on validation data and stop when the schedule reaches it. In the single-center calculation of \citet[Appendix A.7.2]{zhou2026smoothing}, the squared residual per coordinate increases in conditional expectation when it is below the injected variance $\sigma_n^2-\sigma_{n+1}^2$. On a geometric grid this variance shrinks in proportion to $\sigma_n^2$, so the schedule sets a useful empirical resolution for late samples. A geometric first-entry certificate still requires control of the true distance and numerical overshoot.

\clearpage
\section{Diversity: Retention and Regional Coverage}
\label{app:diversity_details}
\label{subsec:manifold_diversity}
The convergence theorem locates individual endpoints. To obtain a generative guarantee, we must also relate the initialized trajectories to probabilities across the data manifold. We combine a geometric bound on refinement with a source--sink identity for the same empirical field.

\subsection{Reference distribution and output time}
Keep the connected reference manifold $\mathcal M$ fixed and specify a probability distribution $Q_{\mathcal M}$ on it. If it has a density relative to manifold volume, write
\begin{equation}
 Q_{\mathcal M}(U)=\int_U q_{\mathcal M}(z)\,d\operatorname{vol}_{\mathcal M}(z).
 \label{eq:reference_manifold_distribution}
\end{equation}
A density is not required by the coverage proof. Intrinsic distance $d_{\mathcal M}$ is the infimum of lengths of piecewise smooth curves on $\mathcal M$; it dominates Euclidean distance. For a Borel region $U$ and $\delta\ge0$, define
\begin{equation}
 U^{-\delta}=\{z\in U:d_{\mathcal M}(z,\mathcal M\setminus U)>\delta\}.
 \label{eq:intrinsic_set_erosion}
\end{equation}
Distance to the empty set is infinite. This safe interior protects region membership against a displacement of at most $\delta$.

Choose the normal-distance thresholds
\begin{equation}
 \varepsilon_{\mathcal M}=\varepsilon_0<r_{\mathrm{stop}}<r_{\min}
 <r_{\mathrm{tr}}<r_0<\tau,\qquad r_{\min}=\sigma_{\min}\sqrt c.
 \label{eq:diversity_radius_ordering}
\end{equation}
The transition radius starts the refinement analysis; the stopping radius specifies the returned sample. They are not radii of regions along the manifold. Define
\begin{align}
 \tau_{\mathrm{tr}}&=\inf\{t\ge0:r(x(t))\le r_{\mathrm{tr}}\},&
 Z_{\mathrm{tr}}&=\Pi_{\mathcal M}(x(\tau_{\mathrm{tr}})),
 \label{eq:transition_entry_time}\\
 \tau_{\mathrm{stop}}&=\inf\{t\ge0:r(x(t))\le r_{\mathrm{stop}}\},&
 Z_{\mathrm{stop}}&=\Pi_{\mathcal M}(x(\tau_{\mathrm{stop}})),
 \label{eq:diversity_stopping_time}\\
 X_{\mathrm{stop}}&=x(\tau_{\mathrm{stop}}).&&
 \label{eq:ambient_stopped_output}
\end{align}
Theorem~\ref{thm:manifold_convergence} makes these times finite almost surely. On exceptional infinite-time events, assign a fixed point of $\mathcal M$ to the corresponding outputs. Strict inward crossing makes entrance into $r\le\rho$ equal to the infimum of times with $r<\rho$. For $a>0$,
$\{\tau_\rho<a\}=\bigcup_{q\in\mathbb Q\cap[0,a)}\{r(\phi_q(X_0))<\rho\}$,
which proves measurability. Implementing this theoretical stop requires the reference distance or a justified surrogate.

\subsection{Tangential motion during refinement}
\begin{lemma}[Projected tangential velocity]
\label{lemma:projected_tangential_velocity}
Under the local geometry and posterior-localization assumptions, for $0<r(x(t))\le r_{\mathrm{tr}}<r_0$ in $\Omega$,
\begin{equation}
 \left\|\frac d{dt}\Pi_{\mathcal M}(x(t))\right\|
 \le L_\Pi\beta(x(t))\sqrt{C_{\mathrm{loc}}m
             \max\{\sigma_{\min}^2,r(x(t))^2/c\}},\qquad
 L_\Pi=\frac{\tau}{\tau-r_{\mathrm{tr}}}.
 \label{eq:projected_tangential_velocity}
\end{equation}
\end{lemma}
\begin{proof}
\textbf{Projection distortion.}
The reach inequality at the projections $z_x,z_y$ of two points in the tube gives
\[
 \langle x-y,z_x-z_y\rangle
 \ge\left(1-\frac{r(x)+r(y)}{2\tau}\right)\|z_x-z_y\|^2.
\]
Apply Cauchy--Schwarz, then take $y=x+hv$ and let $h\to0$. The projection is $C^1$ on the reach tube \citep{leobacher2021existence}, so
\begin{equation}
 \|D\Pi_{\mathcal M}(x)\|\le(1-r(x)/\tau)^{-1}\le L_\Pi.
 \label{eq:projection_derivative_reach_bound}
\end{equation}

\textbf{Removing normal motion.}
Projection is constant along a normal fiber, hence $D\Pi_{\mathcal M}(x)(x-z_x)=0$. The barycenter identity gives
$\dot z=\beta D\Pi_{\mathcal M}(x)(\bar\mu-z)$.
Jensen's inequality bounds $\|\bar\mu-z\|^2$ by the posterior second moment in \eqref{eq:posterior_localization}. Combining the two estimates proves the result.
\end{proof}

\begin{proposition}[Explicit retention under early stopping]
\label{prop:early_stopping_retention}
Under Theorem~\ref{thm:manifold_convergence} and \eqref{eq:diversity_radius_ordering}, almost surely,
\begin{equation}
 r(X_{\mathrm{stop}})\le r_{\mathrm{stop}},\qquad
 d_{\mathcal M}(Z_{\mathrm{tr}},Z_{\mathrm{stop}})\le\delta_{\mathrm{stop}},
 \label{eq:early_stopping_retention}
\end{equation}
where
\begin{equation}
\begin{split}
 \delta_{\mathrm{stop}}={}&L_\Pi\frac{\sqrt{C_{\mathrm{loc}}m/c}}{1-\kappa_0}
       (r_{\mathrm{tr}}-r_{\min})\\
 &+L_\Pi\sqrt{C_{\mathrm{loc}}m}\,\sigma_{\min}
       \log\frac{r_{\min}-\varepsilon_0}{r_{\mathrm{stop}}-\varepsilon_0}.
\end{split}
\label{eq:explicit_early_stopping_retention_radius}
\end{equation}
The same bound controls the entire projected refinement path:
\begin{equation}
 \int_{\tau_{\mathrm{tr}}}^{\tau_{\mathrm{stop}}}
       \left\|\frac d{dt}\Pi_{\mathcal M}(x(t))\right\|dt\le\delta_{\mathrm{stop}}.
 \label{eq:full_projected_refinement_length}
\end{equation}
\end{proposition}
\begin{proof}
Convergence to radius at most $\varepsilon_0<r_{\mathrm{stop}}$ implies finite entrance. If the entrance times coincide, the path length is zero. Otherwise inward invariance and first entrance imply $r_{\mathrm{stop}}<r(t)\le r_{\mathrm{tr}}$ during refinement, with strictly decreasing radius. Dividing the projected-speed bound by the radial contraction bound gives
\[
 \frac{\|\dot z\|}{-\dot r}\le
 \begin{cases}
 L_\Pi\sqrt{C_{\mathrm{loc}}m}\,\sigma_{\min}/(r-\varepsilon_0),&r\le r_{\min},\\
 L_\Pi\sqrt{C_{\mathrm{loc}}m/c}/(1-\kappa_0),&r>r_{\min}.
 \end{cases}
\]
Integrate over the decreasing radius, extending the entrance radius up to $r_{\mathrm{tr}}$ if necessary. The constant integral on $[r_{\min},r_{\mathrm{tr}}]$ and logarithmic integral on $[r_{\mathrm{stop}},r_{\min}]$ give \eqref{eq:explicit_early_stopping_retention_radius}. Intrinsic endpoint distance is bounded by curve length.
\end{proof}

\subsection{The source--sink identity and a point on the same path}
\label{subsec:source_sink_coverage}
Let
\begin{equation}
 p_\sigma^Q(x)=\int_{\mathcal M}\mathcal N(x;z,\sigma^2I)Q_{\mathcal M}(dz).
 \label{eq:source_sink_smoothings}
\end{equation}
This is the density of $Z+\sigma G$ for independent $Z\sim Q_{\mathcal M}$ and $G\sim\mathcal N(0,I_d)$. It is a reference smoothing, rather than the distribution of the optimizer at a fixed time. Take $N\ge2$, $\sigma_i=\sigma_{\min}e^{(i-1)h}$, $h>0$, and define
\begin{equation}
 A_N=h\sum_i\sigma_i^2,\qquad A_Np_{MN}=h\sum_i\sigma_i^2\widehat p_{\sigma_i}.
 \label{eq:source_sink_empirical_objective}
\end{equation}

\begin{proposition}[Multi-scale source--sink identity]
\label{prop:multiscale_source_sink}
For positive finite cutoffs,
\begin{equation}
 \Delta\int_{\sigma_{\min}}^{\sigma_{\max}}\sigma p_\sigma^Q\,d\sigma
 =p_{\sigma_{\max}}^Q-p_{\sigma_{\min}}^Q.
 \label{eq:continuous_source_sink_identity}
\end{equation}
For the actual finite mixture, let
\begin{equation}
 \mathcal R_N=A_N\Delta p_{MN}-(p_{\sigma_{\max}}^Q-p_{\sigma_{\min}}^Q),
 \qquad\epsilon_{\mathrm{Pois}}=\tfrac12\|\mathcal R_N\|_1.
 \label{eq:poisson_residual_definition}
\end{equation}
Then $\mathcal R_N$ is integrable with zero integral and
\begin{equation}
 \nabla\cdot(A_Np_{MN}\nabla\log p_{MN})
 =A_N\Delta p_{MN}=p_{\sigma_{\max}}^Q-p_{\sigma_{\min}}^Q+\mathcal R_N.
 \label{eq:finite_source_sink_identity}
\end{equation}
\end{proposition}
\begin{proof}
Differentiating a Gaussian gives $\partial_\sigma p_\sigma^Q=\sigma\Delta p_\sigma^Q$. Integration in $\sigma$ proves the first identity. Compact reference support and positive finite cutoffs justify differentiation under the integral. The second follows from $p_{MN}\nabla\log p_{MN}=\nabla p_{MN}$. Gaussian Laplacians are integrable with zero integral, and both reference smoothings integrate to one. These facts also prove the residual assertions.
\end{proof}
Here $\Delta p_{MN}$ is the density Laplacian; in particular,
\begin{equation}
 \Delta p_{MN}=p_{MN}\bigl(\Delta\log p_{MN}+\|\nabla\log p_{MN}\|^2\bigr).
 \label{eq:density_and_log_density_laplacians}
\end{equation}
The identity describes stationary flux along a fixed field. It does not prescribe time marginals. Related divergence-based constructions include Poisson flow and Beckmann transport \citep{xu2022poisson,cheuk2026beckmann}.

For distributions $P,Q$, use
\begin{equation}
 \operatorname{TV}(P,Q)=\sup_U|P(U)-Q(U)|,
 \label{eq:total_variation_definition}
\end{equation}
which equals half the $L^1$ distance when densities exist. For measurable $0\le f\le1$,
\begin{equation}
 \left|\int f\,dP-\int f\,dQ\right|\le\operatorname{TV}(P,Q).
 \label{eq:tv_bounded_test_function}
\end{equation}
Consequently applying the same conditional sampling kernel to two input distributions cannot increase TV, even when the input is retained in the output. Define the initialization mismatch by
\begin{equation}
 \epsilon_{\mathrm{init}}=\operatorname{TV}(\nu,p_{\sigma_{\max}}^Q(x)dx).
 \label{eq:source_initialization_mismatch}
\end{equation}

\begin{lemma}[A fine-scale reference point on the score path]
\label{lemma:source_sink_mark}
Let $\mathcal R_N^\pm=\max\{\pm\mathcal R_N,0\}$ and set
$\lambda_{\mathrm{mark}}=(p_{\sigma_{\min}}^Q+\mathcal R_N^-)/(A_Np_{MN})$.
For an independent $E\sim\operatorname{Exp}(1)$, define on the original flow
\begin{equation}
 T_{\mathrm{mark}}=\inf\left\{t\ge0:\int_0^t
 \lambda_{\mathrm{mark}}(\phi_u(X_0))du\ge E\right\},\qquad
 Y=\phi_{T_{\mathrm{mark}}}(X_0).
 \label{eq:auxiliary_score_mark}
\end{equation}
For every start, $T_{\mathrm{mark}}$ is finite almost surely, and under $X_0\sim\nu$,
\begin{equation}
 \operatorname{TV}(P_Y,p_{\sigma_{\min}}^Q(x)dx)
 \le\epsilon_{\mathrm{Pois}}+\epsilon_{\mathrm{init}}.
 \label{eq:marked_fine_scale_law}
\end{equation}
\end{lemma}
The proof in Appendix~\ref{app:source_sink_mark_proof} identifies $A_Np_{MN}$ as the occupation density for a corrected nonnegative source and sink. It requires bounded reference support, so it also applies to the affine construction. The auxiliary mark leaves the trajectory and returned output unchanged. The next argument places it in the refinement segment where geometry controls its relation to the output.

\subsection{From reference probability to output coverage}
Choose $\zeta\in(0,1)$ and the Gaussian norm thresholds
\begin{equation}
 a_d=\sqrt{Q_d(1-\zeta/3)},\qquad
 a_m=\sqrt{Q_m(1-\zeta/3)},\qquad b_c=\sqrt{Q_c(\zeta/3)}.
 \label{eq:source_sink_gaussian_quantiles}
\end{equation}
When $\sigma_{\min}a_d<\tau$, put
\begin{equation}
 \delta_{\mathrm{noise}}=\frac{\sigma_{\min}a_m}{1-\sigma_{\min}a_d/\tau},\qquad
 \delta_{\mathrm{cov}}=\delta_{\mathrm{noise}}+\delta_{\mathrm{stop}}.
 \label{eq:source_sink_coverage_resolution}
\end{equation}

\begin{lemma}[Gaussian projection and placement in the refinement segment]
\label{lemma:gaussian_manifold_projection}
For fixed $z\in\mathcal M$ and $y=z+\sigma_{\min}G$, consider
$\mathcal G_z=\{\|G\|\le a_d,\ \|P_{T_z\mathcal M}G\|\le a_m,\ \|P_{N_z\mathcal M}G\|\ge b_c\}$.
Its probability is at least $1-\zeta$, and on this event
\begin{equation}
 d_{\mathcal M}(z,\Pi_{\mathcal M}(y))\le\delta_{\mathrm{noise}},\qquad
 \sigma_{\min}b_c-\frac{\delta_{\mathrm{noise}}^2}{2\tau}
 \le r(y)\le\sigma_{\min}a_d.
 \label{eq:gaussian_manifold_projection_bound}
\end{equation}
Thus the conditions
\begin{equation}
 \sigma_{\min}a_d\le r_{\mathrm{tr}},\qquad
 r_{\mathrm{stop}}<\sigma_{\min}b_c-\frac{\delta_{\mathrm{noise}}^2}{2\tau}
 \label{eq:source_sink_band_conditions}
\end{equation}
place the perturbation between the two stopping surfaces.
\end{lemma}
Appendix~\ref{app:gaussian_projection_proof} proves the bounds by adding normal noise first, then projecting the segment that adds tangent noise. The parameter $\zeta$ affects both the probability and the admissible radii; it cannot be reduced arbitrarily with these radii fixed.

\begin{theorem}[Convergence and regional coverage]
\label{thm:source_sink_coverage}
Under Proposition~\ref{prop:early_stopping_retention}, the geometric schedule and \eqref{eq:source_sink_band_conditions}, the exact empirical flow converges with $r(x_\infty)\le\varepsilon_{\mathcal M}$ almost surely, and returns $X_{\mathrm{stop}}$ with $r(X_{\mathrm{stop}})\le r_{\mathrm{stop}}$. For every Borel $U\subseteq\mathcal M$,
\begin{equation}
 \mathbb P_\nu[Z_{\mathrm{tr}}\in U,Z_{\mathrm{stop}}\in U]
 \ge(1-\zeta)Q_{\mathcal M}(U^{-\delta_{\mathrm{cov}}})
      -\epsilon_{\mathrm{Pois}}-\epsilon_{\mathrm{init}}.
 \label{eq:source_sink_regional_coverage}
\end{equation}
For $z\in\operatorname{supp}(Q_{\mathcal M})$ and $\rho>\delta_{\mathrm{cov}}$,
\begin{equation}
 \mathbb P_\nu[Z_{\mathrm{stop}}\in B_{\mathcal M}(z,\rho)]
 \ge(1-\zeta)Q_{\mathcal M}(B_{\mathcal M}(z,\rho-\delta_{\mathrm{cov}}))
       -\epsilon_{\mathrm{Pois}}-\epsilon_{\mathrm{init}}.
 \label{eq:source_sink_ball_coverage}
\end{equation}
The latter also bounds $\mathbb P_\nu[\|X_{\mathrm{stop}}-z\|<r_{\mathrm{stop}}+\rho]$.
\end{theorem}
\begin{proof}
\textbf{A reference label for the actual path.}
Construct $Y$ and attach a label $Z$ with the kernel
\begin{equation}
 K_Q(y,dz)=\frac{\mathcal N(y;z,\sigma_{\min}^2I)}{p_{\sigma_{\min}}^Q(y)}Q_{\mathcal M}(dz).
 \label{eq:source_sink_reference_coupling}
\end{equation}
Call the augmented distribution $\mathbb P_{\mathrm{act}}$. Compare it with
$\mathbb P_{\mathrm{id}}(dz,dy)=Q_{\mathcal M}(dz)\mathcal N(y;z,\sigma_{\min}^2I)dy$.
Both use the same kernel given $Y$, so the mark lemma and TV contraction give, for every joint event $D$,
\begin{equation}
 |\mathbb P_{\mathrm{act}}((Z,Y)\in D)-\mathbb P_{\mathrm{id}}(D)|
 \le\epsilon_{\mathrm{Pois}}+\epsilon_{\mathrm{init}}.
 \label{eq:actual_ideal_joint_comparison}
\end{equation}
Only under $\mathbb P_{\mathrm{id}}$ is the reference label exactly distributed as $Q_{\mathcal M}$.

\textbf{A reference event with controlled probability.}
Let $\mathcal G$ be the Gaussian inequalities in Lemma~\ref{lemma:gaussian_manifold_projection}, expressed using $(Y-Z)/\sigma_{\min}$. Under the ideal distribution its conditional probability is at least $1-\zeta$ for every $Z=z$. Thus
\begin{equation}
 \mathbb P_{\mathrm{act}}[Z\in U^{-\delta_{\mathrm{cov}}},\mathcal G]
 \ge(1-\zeta)Q_{\mathcal M}(U^{-\delta_{\mathrm{cov}}})
       -\epsilon_{\mathrm{Pois}}-\epsilon_{\mathrm{init}}.
 \label{eq:source_sink_good_event_mass}
\end{equation}
All geometric implications of $\mathcal G$ are deterministic and therefore also hold for the actual labeled trajectory.

\textbf{Following the same mark to the output.}
On this event, $r_{\mathrm{stop}}<r(Y)\le r_{\mathrm{tr}}$ and
$d_{\mathcal M}(Z,\Pi_{\mathcal M}(Y))\le\delta_{\mathrm{noise}}$.
First entrance and inward invariance imply
$\tau_{\mathrm{tr}}\le T_{\mathrm{mark}}<\tau_{\mathrm{stop}}$.
The full projected path-length bound controls both subsegments adjacent to the mark, so each of $Z_{\mathrm{tr}}$ and $Z_{\mathrm{stop}}$ is within $\delta_{\mathrm{cov}}$ of $Z$. Safe-interior membership puts both in $U$. Taking probabilities proves \eqref{eq:source_sink_regional_coverage}. Auxiliary labeling has not changed either output distribution.

For the ball version, a label in $B_{\mathcal M}(z,\rho-\delta_{\mathrm{cov}})$ gives $d_{\mathcal M}(z,Z_{\mathrm{stop}})<\rho$. Finally,
$\|X_{\mathrm{stop}}-z\|\le r_{\mathrm{stop}}+d_{\mathcal M}(Z_{\mathrm{stop}},z)$.
This yields both remaining assertions.
\end{proof}

\subsection{Noise, sampling error and coverage resolution}
For distributions with finite first moments, define
\begin{equation}
 W_1(P,Q)=\inf_{\gamma\in\Gamma(P,Q)}\int\|u-v\|\,\gamma(du,dv),
 \label{eq:wasserstein_one_definition}
\end{equation}
where $\Gamma(P,Q)$ denotes joint distributions with marginals $P,Q$.
Let $B=\sup_{z\in\operatorname{supp}(Q_{\mathcal M})}\|z\|$.

\begin{proposition}[Probability-error decomposition and generic input bounds]
\label{prop:source_sink_input_errors}
Define the finite-grid error
\begin{equation}
\epsilon_{\mathrm{grid}}
:=
\frac12
\left\|
A_N\Delta p_{MN}
-
\bigl(\widehat p_{\sigma_{\max}}
      -\widehat p_{\sigma_{\min}}\bigr)
\right\|_1
\label{eq:source_sink_grid_error_definition}
\end{equation}
and the cutoff mismatch
\begin{equation}
\epsilon_{\mathrm{cut}}
:=
\operatorname{TV}
\bigl(\widehat p_{\sigma_{\max}}dx,p_{\sigma_{\max}}^Qdx\bigr)
+
\operatorname{TV}
\bigl(\widehat p_{\sigma_{\min}}dx,p_{\sigma_{\min}}^Qdx\bigr).
\label{eq:source_sink_cutoff_error_definition}
\end{equation}
Then
\begin{equation}
\epsilon_{\mathrm{Pois}}
\le
\epsilon_{\mathrm{grid}}+\epsilon_{\mathrm{cut}}.
\label{eq:source_sink_error_decomposition}
\end{equation}
For the finite geometric schedule, these terms admit the generic bounds
\begin{align}
\epsilon_{\mathrm{grid}}
&\le
dh\log\frac{\sigma_{\max}}{\sigma_{\min}}
+\frac h2\sqrt{2d},
\label{eq:source_sink_generic_grid_bound}\\
\epsilon_{\mathrm{cut}}
&\le
\frac{W_1(\widehat Q_M,Q_{\mathcal M})}{\sqrt{2\pi}}
\left(
\sigma_{\min}^{-1}+\sigma_{\max}^{-1}
\right).
\label{eq:source_sink_generic_cutoff_bound}
\end{align}
Consequently,
\begin{equation}
\begin{split}
\epsilon_{\mathrm{Pois}}
\le{}&
dh\log\frac{\sigma_{\max}}{\sigma_{\min}}
+\frac h2\sqrt{2d}\\
&+
\frac{W_1(\widehat Q_M,Q_{\mathcal M})}{\sqrt{2\pi}}
\left(
\sigma_{\min}^{-1}+\sigma_{\max}^{-1}
\right).
\end{split}
\label{eq:source_sink_explicit_residual_bound}
\end{equation}

For the truncated Gaussian
\eqref{eq:truncated_gaussian_initialization},
\begin{equation}
\epsilon_{\mathrm{init}}
\le
\delta_{\mathrm{init}}
+
\frac{\int\|z\|\,Q_{\mathcal M}(dz)}
     {\sqrt{2\pi}\sigma_{\max}}
\le
\delta_{\mathrm{init}}
+
\frac{B}{\sqrt{2\pi}\sigma_{\max}}.
\label{eq:source_sink_gaussian_initialization_bound}
\end{equation}
\end{proposition}
Appendix~\ref{app:source_sink_errors} proves this decomposition by first
comparing the finite scale sum with its continuous log-scale integral and
then comparing the empirical and reference smoothings only at the two
cutoffs. Thus finite-sample mismatch enters through smoothed endpoint
distributions rather than throughout the entire scale path.

The explicit bounds in
\eqref{eq:source_sink_generic_grid_bound}--\eqref{eq:source_sink_generic_cutoff_bound}
are distribution-free sufficient estimates. In particular, the
$O(dh\log(\sigma_{\max}/\sigma_{\min}))$ grid term comes from uniform
$L^1$ control of Gaussian scale derivatives and can be conservative in
high ambient dimension. The coverage theorem itself depends on the actual
errors $\epsilon_{\mathrm{Pois}}$ and $\epsilon_{\mathrm{init}}$, for which
structured models may admit substantially sharper control. These generic
bounds are useful for making the scale tradeoffs explicit: decreasing
$\sigma_{\min}$ improves geometric resolution but can increase finite-data
cutoff mismatch, whereas increasing $\sigma_{\max}$ improves initialization
overlap while requiring finer scale discretization.

\paragraph{Stopping policies.}
\label{subsec:diversity_parameter_effects}
For $a>1$, $0<b<1$, set $r_{\mathrm{tr}}=a\sigma_{\min}\sqrt c<r_0$ and $r_{\mathrm{stop}}=b\sigma_{\min}\sqrt c$, with $\varepsilon_0\le(b/2)\sigma_{\min}\sqrt c$. Substitution into the retention bound gives
\begin{equation}
 \delta_{\mathrm{stop}}\le
 \frac{\sqrt{C_{\mathrm{loc}}m}\sigma_{\min}}{1-a\sigma_{\min}\sqrt c/\tau}
 \left[\frac{a-1}{1-\kappa_0}+\log\frac2b\right].
 \label{eq:explicit_reach_noise_retention_scaling}
\end{equation}
Thus the retention scale is $O(\sqrt m\,\sigma_{\min})$ when the constants and geometric margins remain controlled. The Gaussian band conditions can hold by choosing $a>a_d/\sqrt c$, $b<b_c/\sqrt c$, and a sufficiently small relative noise scale.
For $r_{\mathrm{stop}}=2\varepsilon_0<r_{\min}$ instead,
\begin{align}
 r(X_{\mathrm{stop}})&\le \frac{C_{\mathrm{loc}}m}{\tau}\sigma_{\min}^2,
 \label{eq:quadratic_stopped_normal_accuracy}\\
 \delta_{\mathrm{stop}}&\le
 \frac{\sqrt{C_{\mathrm{loc}}m}\sigma_{\min}}{1-a\sigma_{\min}\sqrt c/\tau}
 \left[\frac{a-1}{1-\kappa_0}
   +\log\frac{2\tau\sqrt c}{C_{\mathrm{loc}}m\sigma_{\min}}\right].
 \label{eq:quadratic_stopped_retention}
\end{align}
This follows from $r_{\min}-\varepsilon_0\le r_{\min}$ in the logarithm. It gives quadratic normal accuracy at a logarithmically larger retention scale. In either policy, coverage additionally requires the stated Gaussian band and a probability-error budget smaller than the reference region mass. The general endpoint needs further tangential control (Appendix~\ref{app:endpoint_retention}); the affine construction in Appendix~\ref{app:resolved_endpoint_proof} supplies it at the training resolution.

\clearpage
\section{Source--Sink Proofs and Error Bounds}
\label{app:source_sink}

\subsection{A marked trajectory and its occupation density}
\label{app:source_sink_mark_proof}
\begin{proof}[Proof of Lemma~\ref{lemma:source_sink_mark}]
\textbf{Complete trajectories and finite marking.}
With $B_M=\max_j\|\mu_j\|$, the barycenter identity gives
$\|\nabla\log p_{MN}(x)\|\le\sigma_{\min}^{-2}(B_M+\|x\|)$.
The score and its negative have at most linear growth, so Gronwall's inequality gives a complete flow $\phi_t$ for all real $t$. It is a smooth diffeomorphism with inverse $\phi_{-t}$. Each forward path stays in its compact positive superlevel set. The continuous marking rate is strictly positive on that set, hence its integral along the path diverges and $T_{\mathrm{mark}}<\infty$ almost surely. Joint continuity of the integrated rate ensures measurability.

\textbf{Nonnegative source and sink.}
For this proof put $a=p_{\sigma_{\max}}^Q+\mathcal R_N^+$ and
$b=p_{\sigma_{\min}}^Q+\mathcal R_N^-$. Both have integral
$1+\epsilon_{\mathrm{Pois}}$, since the positive and negative parts of the residual have the same integral $\epsilon_{\mathrm{Pois}}$. The flux identity becomes
\begin{equation}
 \nabla\cdot(A_Np_{MN}\nabla\log p_{MN})=a-b,
 \qquad\lambda_{\mathrm{mark}}A_Np_{MN}=b.
 \label{eq:marked_stationary_balance}
\end{equation}
These are unnormalized source and sink densities of equal mass.

\textbf{Occupation density.}
Let $\mathsf T_t f$ transport an integrable nonnegative density $f$ along the score flow while removing each trajectory at its mark. The inverse-flow Jacobian and conditional survival probability give
\begin{equation}
 (\mathsf T_t f)(y)=f(\phi_{-t}(y))
 \exp\!\left[-\int_0^t(\Delta\log p_{MN}+\lambda_{\mathrm{mark}})
                    (\phi_{-u}(y))du\right].
 \label{eq:marked_characteristic_density}
\end{equation}
Consequently,
\begin{equation}
 \|\mathsf T_t f\|_1
 =\int f(x_0)e^{-\int_0^t\lambda_{\mathrm{mark}}(\phi_u(x_0))du}dx_0
 \le\|f\|_1.
 \label{eq:marked_mass_formula}
\end{equation}
Differentiate the characteristic formula for $f=A_Np_{MN}$. The product rule and \eqref{eq:marked_stationary_balance} give
$\frac d{dt}\mathsf T_t(A_Np_{MN})=-\mathsf T_t a$.
The rate is continuous, which suffices to differentiate the time integral; derivatives of $\mathcal R_N^-$ are not needed. Integrating gives
\begin{equation}
 A_Np_{MN}-\mathsf T_t(A_Np_{MN})=\int_0^t\mathsf T_u a\,du.
 \label{eq:marked_finite_occupation}
\end{equation}
The right-hand side increases pointwise. As each mark is finite, dominated convergence in \eqref{eq:marked_mass_formula} gives $\|\mathsf T_t(A_Np_{MN})\|_1\to0$. Monotone convergence therefore implies
\begin{equation}
 \int_0^\infty\mathsf T_t a\,dt=A_Np_{MN}\quad\text{almost everywhere}.
 \label{eq:marked_occupation_identity}
\end{equation}

\textbf{Distribution of the marked point.}
The mark kernel is
\[
 \mathsf K(x_0,U)=\int_0^\infty\mathbf1_U(\phi_t(x_0))
 \lambda_{\mathrm{mark}}(\phi_t(x_0))
 e^{-\int_0^t\lambda_{\mathrm{mark}}(\phi_u(x_0))du}dt.
\]
It has total mass one. Tonelli's theorem, a flow change of variables and the occupation identity give
\[
 \int a(x_0)\mathsf K(x_0,U)dx_0
 =\int_U\lambda_{\mathrm{mark}}(y)\int_0^\infty(\mathsf T_ta)(y)dt\,dy
 =\int_U b(y)dy.
\]
In measure notation this is $(a\,dx)\mathsf K=b\,dx$. Subtract the residual terms:
\[
 (p_{\sigma_{\max}}^Qdx)\mathsf K-p_{\sigma_{\min}}^Qdx
 =\mathcal R_N^-dx-(\mathcal R_N^+dx)\mathsf K.
\]
Both measures on the right have mass $\epsilon_{\mathrm{Pois}}$, so their difference on every event has magnitude at most that value. Finally, TV contracts under the same kernel, yielding
\[
 \operatorname{TV}(P_Y,p_{\sigma_{\min}}^Qdx)
 \le\operatorname{TV}(\nu,p_{\sigma_{\max}}^Qdx)+\epsilon_{\mathrm{Pois}}.
\]
This proves the claim for the unchanged initialized trajectories.
\end{proof}

\subsection{Projection of a Gaussian perturbation}
\label{app:gaussian_projection_proof}
\begin{proof}[Proof of Lemma~\ref{lemma:gaussian_manifold_projection}]
Write $G_T=P_{T_z\mathcal M}G$ and $G_N=P_{N_z\mathcal M}G$.
The squared norms of $G,G_T,G_N$ have the indicated chi-squared distributions. Each excluded tail has probability $\zeta/3$, so their union has probability at most $\zeta$.

\textbf{Tangential projection error.}
On the good event the segment
$\gamma(v)=z+\sigma_{\min}G_N+v\sigma_{\min}G_T$, $0\le v\le1$,
stays in the reach tube because $\|\gamma(v)-z\|\le\sigma_{\min}a_d<\tau$. Its initial projection is $z$, and its final projection is $q=\Pi_{\mathcal M}(y)$. By the projection derivative bound,
\[
 d_{\mathcal M}(z,q)\le\int_0^1
 \|D\Pi_{\mathcal M}(\gamma(v))\sigma_{\min}G_T\|dv
 \le\frac{\sigma_{\min}a_m}{1-\sigma_{\min}a_d/\tau}
 =\delta_{\mathrm{noise}}.
\]

\textbf{Normal placement.}
The reach inequality for all unit normals at $z$ gives
$\|P_{N_z\mathcal M}(q-z)\|\le\|q-z\|^2/(2\tau)$. Hence
\[
 r(y)\ge\|\sigma_{\min}G_N-P_{N_z\mathcal M}(q-z)\|
 \ge\sigma_{\min}b_c-\delta_{\mathrm{noise}}^2/(2\tau).
\]
The upper bound is $r(y)\le\|y-z\|\le\sigma_{\min}a_d$.
The band conditions place $y$ in the refinement segment.
\end{proof}

\subsection{Finite-grid, finite-sample and initialization errors}
\label{app:source_sink_errors}
\begin{proof}[Proof of Proposition~\ref{prop:source_sink_input_errors}]
\textbf{Log-scale quadrature.}
Write $\varphi_\sigma=\mathcal N(\cdot;0,\sigma^2I)$ and
$k_\sigma=\sigma^2\Delta\varphi_\sigma=(\|x\|^2/\sigma^2-d)\varphi_\sigma$.
For $t=\log(\sigma/\sigma_{\min})$ and $u=\|x\|^2/\sigma^2$,
\[
 \|k_\sigma\|_1\le\sqrt{2d},\qquad
 \partial_t k_\sigma=[(u-d)^2-2u]\varphi_\sigma,\qquad
 \|\partial_t k_\sigma\|_1\le4d.
\]
The first bound uses the variance $2d$ of $\|G\|^2$; the last uses its centered second moment and mean. Convolution with $\widehat Q_M$ does not increase $L^1$ norms. Thus $f(t)=\sigma^2\Delta\widehat p_\sigma$ is $4d$-Lipschitz in $L^1$.
Set $b=\log(\sigma_{\max}/\sigma_{\min})=(N-1)h$. The scale identity gives
$\int_0^b f(t)dt=\widehat p_{\sigma_{\max}}-\widehat p_{\sigma_{\min}}$.
Comparing each integral cell with its left endpoint costs at most $2dh^2$ in $L^1$. The extra upper endpoint of the finite mixture contributes at most $h\sqrt{2d}$. Therefore
\begin{equation}
 \tfrac12\|A_N\Delta p_{MN}-(\widehat p_{\sigma_{\max}}-\widehat p_{\sigma_{\min}})\|_1
 \le dhb+\tfrac h2\sqrt{2d}.
 \label{eq:source_sink_grid_error}
\end{equation}

\textbf{Empirical versus reference smoothings.}
For any centers $u,v$, integration along their joining segment gives
\[
 \|\varphi_\sigma(\cdot-u)-\varphi_\sigma(\cdot-v)\|_1
 \le\|u-v\|\sup_{\|e\|=1}\|\partial_e\varphi_\sigma\|_1
 =\sqrt{2/\pi}\,\|u-v\|/\sigma.
\]
Average over a coupling and take the infimum:
\begin{equation}
 \operatorname{TV}(\widehat p_\sigma dx,p_\sigma^Qdx)
 \le\frac{W_1(\widehat Q_M,Q_{\mathcal M})}{\sqrt{2\pi}\sigma}.
 \label{eq:source_sink_smoothed_wasserstein}
\end{equation}

Decompose the residual as
\[
\mathcal R_N
=
\Bigl[
A_N\Delta p_{MN}
-(\widehat p_{\sigma_{\max}}-\widehat p_{\sigma_{\min}})
\Bigr]
+
(\widehat p_{\sigma_{\max}}-p_{\sigma_{\max}}^Q)
-
(\widehat p_{\sigma_{\min}}-p_{\sigma_{\min}}^Q).
\]
Using $\operatorname{TV}(P,Q)=\tfrac12\|p-q\|_1$ for densities and the
triangle inequality gives
\[
\epsilon_{\mathrm{Pois}}
\le
\epsilon_{\mathrm{grid}}+\epsilon_{\mathrm{cut}}.
\]
Equation~\eqref{eq:source_sink_grid_error} gives
\eqref{eq:source_sink_generic_grid_bound}, while applying
\eqref{eq:source_sink_smoothed_wasserstein} at the two cutoffs gives
\eqref{eq:source_sink_generic_cutoff_bound}. Combining the two proves
\eqref{eq:source_sink_explicit_residual_bound}.

\textbf{Gaussian initialization.}
Conditioning a distribution on an event of mass $1-\delta_{\mathrm{init}}$ changes TV by exactly $\delta_{\mathrm{init}}$. The same translation estimate gives
\[
 \operatorname{TV}(\mathcal N(0,\sigma_{\max}^2I),p_{\sigma_{\max}}^Qdx)
 \le\frac{\int\|z\|Q_{\mathcal M}(dz)}{\sqrt{2\pi}\sigma_{\max}}
 \le\frac B{\sqrt{2\pi}\sigma_{\max}}.
\]
Add the truncation cost. For an untruncated Gaussian, this cost is zero.
\end{proof}

\paragraph{Continuous-scale specializations.}
For a bounded log-continuous mixture, $A_N$ is replaced by
$\int_{\sigma_{\min}}^{\sigma_{\max}}\sigma\,d\sigma$.
The quadrature error vanishes; when its center distribution is also $Q_{\mathcal M}$, the entire residual vanishes. Compact center support and positive finite cutoffs ensure the same analyticity, Gaussian decay and complete-flow properties. The local geometry must still hold for that instance.

\subsection{Why the scale weight matters}
\label{app:general_scale_weight}
For a positive $C^1$ weight $w(\sigma)$ aggregated against $d\sigma/\sigma$, integration by parts gives
\[
 \Delta\int_{\sigma_{\min}}^{\sigma_{\max}}w(\sigma)p_\sigma^Q\frac{d\sigma}{\sigma}
 =\left[\frac{w(\sigma)}{\sigma^2}p_\sigma^Q\right]_{\sigma_{\min}}^{\sigma_{\max}}
 -\int_{\sigma_{\min}}^{\sigma_{\max}}
        \left(\frac{w(\sigma)}{\sigma^2}\right)'p_\sigma^Q\,d\sigma.
\]
The choice $w(\sigma)\propto\sigma^2$ removes intermediate-scale terms and yields the endpoint balance used in the proof. Other weights still define fixed objectives, but this identity generally has additional source or sink terms. When integrating against $d\sigma$ instead, the corresponding weight is proportional to $\sigma$.

\clearpage
\section{Endpoint Non-Memorization at the Training-Set Resolution}
\label{app:affine_nonmemorization}
\label{app:resolved_endpoint_proof}

This appendix proves Corollary~\ref{cor:sample_resolution_nonmemorization}
using finite-mixture convergence, exact affine normal contraction, and the
source--sink marked-distribution lemma. Its main ingredients are a finite-grid
sign estimate at the training resolution and a coupling that preserves
regional probability. The reference box has compact support, as required
by the marked-distribution argument; the containing affine space may be unbounded.

\subsection{Construction and probability coupling}
Use the finite grid, reference box $D_L$, and output cores $C_{\mathbf k}$ of Section~\ref{subsec:main_nonmemorization}. Directly from the largest one-dimensional gap and the outer boundary distance,
\begin{equation}
 h_{\mathrm{cov},M}=\sup_{z\in D_L}\operatorname{dist}(z,\mathcal S_M)
 =\sqrt m\,\ell/3,
 \label{eq:resolved_grid_covering_radius}
\end{equation}
and the cores are
\begin{equation}
 C_{\mathbf k}=\{(u,0):|u_j-\ell k_j|\le\ell/12\text{ for all }j\}.
 \label{eq:resolved_grid_output_cells}
\end{equation}
The proof verifies one finite relative schedule on the full interval
\begin{equation}
 \sigma_{\min}\in[s_0\ell,2s_0\ell],\qquad
 \sigma_i=\sigma_{\min}e^{(i-1)h},\quad N\ge2.
 \label{eq:resolved_nonempty_bandwidth_window}
\end{equation}
Every trajectory converges, with
\begin{equation}
 x_\infty\in\mathcal A\cap\operatorname{conv}(\mathcal S_M),
 \label{eq:resolved_endpoint_affine_geometry}
\end{equation}
by the finite-mixture and affine results. We prove that the covered endpoints additionally satisfy
\begin{equation}
 \mathbb P[\operatorname{dist}(x_\infty,\mathcal S_M)\ge h_{\mathrm{cov},M}/4]
 \ge1-\epsilon.
 \label{eq:resolved_endpoint_training_separation}
\end{equation}

\paragraph{Additional consequences of the main corollary.}
At every covered output cell, the center marginal obeys
\begin{equation}
 \max_j\sum_i\widetilde w_{ij}(x_\infty)
 \le\left(1+e^{-1/(36s_0^2)}\right)^{-m}<0.529^m
 \quad\text{on }\{x_\infty\in\bigcup_{\mathbf k}C_{\mathbf k}\}.
 \label{eq:resolved_endpoint_center_nondominance}
\end{equation}
The proof constructs a coupling with \(Z\sim Q_{\mathcal M}\) such that
\begin{equation}
 \mathbb P\!\left[
 \|x_\infty-Z\|\le7h_{\mathrm{cov},M}/4,\
 \operatorname{dist}(x_\infty,\mathcal S_M)\ge h_{\mathrm{cov},M}/4
 \right]\ge1-\epsilon.
 \label{eq:resolved_endpoint_reference_coupling}
\end{equation}
It also preserves the reference grid-cell label with that probability.
Consequently,
\begin{equation}
 W_1(P_{x_\infty},Q_{\mathcal M})
 \le7h_{\mathrm{cov},M}/4+\epsilon\operatorname{diam}(D_L).
 \label{eq:resolved_endpoint_wasserstein}
\end{equation}
Thus the coupling controls endpoint discrepancy at the training-resolution
scale, with an explicit probability error.

\subsection{Coupling a marked path to a resolved reference cell}
\label{app:resolved_mark_transfer}

We first record the probability step used below. Work on
\(\mathcal A=\mathbb R^m\times\{0\}\), and write
\(Q_{\mathcal M}*\mathcal N(0,\sigma^2I_m)\) for convolution in its
\emph{intrinsic} coordinates. The ambient smoothing then factors as
this distribution times \(\mathcal N(0,\sigma^2I_c)\) in the normal coordinates.
This factorization concerns a smoothing, not the time marginal of the
actual optimization flow.

Suppose disjoint capture sets \(B_{\mathbf k}\subseteq\mathcal A\)
and output sets \(C_{\mathbf k}\subseteq\mathcal A\) have the following
verified property: a path through
\[
 \{x:\Pi_{\mathcal A}(x)\in B_{\mathbf k},\
             r_{\mathcal A}(x)\le r_{\mathrm{tr}}\}
\]
converges to \(C_{\mathbf k}\). Let \(Z\sim Q_{\mathcal M}\), and let
its reference cell label be \(\mathbf k(Z)\). Assume that membership
in \(B_{\mathbf k}\) agrees with this label. Define, only for this
comparison,
\begin{align*}
 e_{\mathrm{blur}}&:=\operatorname{TV}\!\left(
 Q_{\mathcal M}*\mathcal N(0,\sigma_{\min}^2I_m),Q_{\mathcal M}\right),\\
 e_{\mathrm{cell}}&:=Q_{\mathcal M}\!\left(\mathcal A\setminus
                               \bigcup_{\mathbf k}B_{\mathbf k}\right),\\
 e_{\mathrm{normal}}&:=\mathbb P[\sigma_{\min}\|G_c\|>r_{\mathrm{tr}}].
\end{align*}
Lemma~\ref{lemma:source_sink_mark} gives a mark \(Y\) on the actual
path with total-variation error at most
\(\epsilon_{\mathrm{Pois}}+\epsilon_{\mathrm{init}}\) from the
ambient reference smoothing. Couple it maximally to
\(Y^Q\sim p_{\sigma_{\min}}^Q(x)\,dx\). Then couple
\(\Pi_{\mathcal A}(Y^Q)\) maximally to \(Z\sim Q_{\mathcal M}\).
These constructions can be attached conditionally to the original path;
neither changes that path or its endpoint. The event
\begin{equation}
 Y=Y^Q,\qquad Z=\Pi_{\mathcal A}(Y^Q),\qquad
 Z\in\bigcup_{\mathbf k}B_{\mathbf k},\qquad
 r_{\mathcal A}(Y^Q)\le r_{\mathrm{tr}}
 \label{eq:resolved_joint_good_event}
\end{equation}
has probability at least
\begin{equation}
 1-\epsilon_{\mathrm{Pois}}-\epsilon_{\mathrm{init}}
   -e_{\mathrm{blur}}-e_{\mathrm{cell}}-e_{\mathrm{normal}}.
 \label{eq:resolved_joint_error_budget}
\end{equation}
On this event \(x_\infty\in C_{\mathbf k(Z)}\).

A maximal coupling places the common density $\min\{dP/d(P+Q),dQ/d(P+Q)\}$ on the diagonal, giving equality probability $1-\operatorname{TV}(P,Q)$. Conditional kernels attach these couplings to the existing trajectory. A union bound gives \eqref{eq:resolved_joint_error_budget}.

This differs from the Gaussian-posterior reference label
in \eqref{eq:source_sink_reference_coupling}. There the label comparison
pays a spatial Gaussian displacement. Here the reference has an intrinsic
density, and its small total-variation change under smoothing permits
an equality coupling except on a controlled probability error. Such a
comparison is \emph{not} available for an arbitrary reference distribution.

\subsection{Periodic comparison and the finite-cloud error}
\label{app:resolved_grid_signs}

Rescale lengths by \(\ell\), so the grid spacing is one. Write
\(s=\sigma_{\min}/\ell\); in these coordinates we again use
\(\sigma_i\) for the rescaled noise levels. Scores are multiplied by a
positive constant under this rescaling, and the time variable is
rescaled by \(\ell^2\); trajectories and endpoint statements are
otherwise unchanged. The following quantities are local to this proof.

Let \(\varphi_\sigma\) denote the one-dimensional normalized Gaussian
and consider the infinite paired grid
\begin{equation}
 g_\sigma(u):=\frac12\sum_{k\in\mathbb Z}
 \left[\varphi_\sigma(u-k-1/6)+\varphi_\sigma(u-k+1/6)\right].
 \label{eq:resolved_periodic_gaussian}
\end{equation}
This is a periodic comparison function, not a replacement density for
the finite sampler. Integrating against \(e^{-2\pi inu}\) over one
period, shifting each Gaussian, and summing the intervals gives its
Fourier series
\begin{equation}
 g_\sigma(u)=1+2\sum_{n\ge1}\cos(n\pi/3)
                  e^{-2\pi^2n^2\sigma^2}\cos(2\pi nu).
 \label{eq:resolved_periodic_fourier}
\end{equation}
The Gaussian Fourier integral follows, for example, by differentiating
its characteristic function and integrating by parts. Gaussian decay
makes the displayed series and its spatial derivatives absolutely and
uniformly convergent, so all differentiations below are legitimate.

For \(\sigma\ge1/2\), put \(q=e^{-2\pi^2\sigma^2}<1/100\).
Then
\[
 |g_\sigma-1|\le\frac{2q}{1-q},\qquad g_\sigma\ge\tfrac12,
 \qquad
 \sum_{n\ge2}n^2q^{n^2}\le\frac{4q^4}{1-4q}\le\frac q4.
\]
To verify the sum bound, use \(n^2\le4^{n-1}\) and
\(n^2\ge n+2\) for integers \(n\ge2\). For \(0<t<\pi\),
\(|\sin(nt)|\le n\sin t\); induction follows from the angle-addition
formula. Since \(\cos(\pi/3)=1/2\), the first harmonic therefore
controls the derivative:
\begin{equation}
 g_\sigma'(u)\le-\pi q\sin(2\pi u)<0
 \quad(0<u<1/2),
 \label{eq:resolved_periodic_derivative_sign}
\end{equation}
with the opposite sign on \((-1/2,0)\), repeated every period.
The maxima of this comparison lie between, not on, the two training
positions in each cell. Importantly, every fixed scale has the same
coordinate signs.

At a state \(x=(u,n)\), the infinite-array fixed-scale sum in
\(d=m+c\) dimensions is
\[
 \mathscr G_\sigma(u,n)
 =(2\pi\sigma^2)^{-c/2}e^{-\|n\|^2/(2\sigma^2)}
                                      \prod_{j=1}^m g_\sigma(u_j).
\]
The scale mixture does not generally factor across coordinates; we use
factorization only \emph{at a fixed scale}, then sum same-sign
contributions. Choose \(0<\theta\le1/12\) and use the Gaussian threshold
\(a_d\) already defined in \eqref{eq:source_sink_gaussian_quantiles}.
For \(\|n\|\le s a_d\),
\(1/12\le u_j-k_j\le1/2-\theta\), and arbitrary other coordinates,
the derivative of \(h\sum_i\sigma_i^2\mathscr G_{\sigma_i}\)
is negative with magnitude at least
\begin{equation}
 A(s,\theta):=
 \frac{\pi\log2}{2^{m+c+1}(2\pi)^{c/2}}
 s^{2-c}e^{-a_d^2/2-4\pi^2s^2}\sin(2\pi\theta)>0,
 \label{eq:resolved_periodic_sign_margin}
\end{equation}
provided \(\sigma_{\max}/s\ge2\) and \(h\le(\log2)/4\).
Indeed, all scales have the required sign. Restrict the sum to
\(s\le\sigma_i\le\sqrt2s\); its total \(h\)-weight is at least
\((\log2)/4\). On that band use \(\sigma_i^2\ge s^2\),
\(\sigma_i^{-c}\ge(2s)^{-c}\), the normal lower bound
\(e^{-a_d^2/2}\), \(q\ge e^{-4\pi^2s^2}\), and
\(\prod_{\ell\ne j}g_{\sigma_i}\ge2^{-(m-1)}\).
Also \(\sin(2\pi(u_j-k_j))\ge\sin(2\pi\theta)\).
Reflection gives the positive margin for negative coordinate offsets.
For \(c\ge2\), \(A(s,\theta)\) decreases with \(s\).

We next control the error caused by truncating the training array. For
\(d>1\), a geometric grid with \(h\le1\) satisfies
\begin{equation}
 h\sum_i\sigma_i^{-d}e^{-r^2/(2\sigma_i^2)}
 \le K_d r^{-d},\qquad
 K_d:=2^{d/2-1}\Gamma(d/2)+2d^{d/2}e^{-d/2}.
 \label{eq:resolved_discrete_kernel_tail}
\end{equation}
To see this, extend the positive sum to the whole geometric grid. The
log-scale integrand is unimodal. Its Riemann sum is at most its integral
plus \(2h\) times its maximum, by summing the variation over grid
intervals. The substitution \(v=r^2/(2\sigma^2)\) gives the integral
\(2^{d/2-1}\Gamma(d/2)r^{-d}\); the maximum is
\(d^{d/2}e^{-d/2}r^{-d}\).

Let \(J\ge\max\{1,\sqrt m/3\}\) be an integer padding width. Suppose
\(|k_j|\le L-J\), \(|u_j-k_j|\le1/2-\theta\), and
\(\|n\|\le s a_d\). Every omitted training position has tangential
distance at least \(J\) from \(u\). The error in any coordinate
derivative of the scaled actual density
\((2L+1)^m A_Np_{MN}\), compared with
\(h\sum_i\sigma_i^2\mathscr G_{\sigma_i}\), is at most
\begin{equation}
 C_{m,d}J^{1-c},\qquad
 C_{m,d}:=(2\pi)^{-d/2}K_d
   (3/2)^{m+d-1}\frac{2^{c-1}|S^{m-1}|}{c-1}.
 \label{eq:resolved_finite_array_gradient_error}
\end{equation}
Here \(|S^{m-1}|=2\pi^{m/2}/\Gamma(m/2)\); in particular
\(|S^0|=2\). This bound is uniform in the upper cutoff and in the
number of scales. It is the reason for the sufficient condition
\(c\ge2\) in this construction, not a claim of impossibility for
codimension one.

For the details, differentiating one weighted Gaussian gives
\(\partial_{u_j}(\sigma^2\mathcal N(x;\mu,\sigma^2I))
=(\mu_j-u_j)\mathcal N(x;\mu,\sigma^2I)\).
Equation~\eqref{eq:resolved_discrete_kernel_tail} bounds its scale sum
in absolute value by \((2\pi)^{-d/2}K_d\|x-\mu\|^{1-d}\), and hence
by the same constant times \(\|u-\mu_\parallel\|^{1-d}\).
The infinite paired grid has disjoint cubes of side \(1/3\) around
its positions, each weighted by \(2^{-m}\). On such a cube,
\(\|u-y\|\le(3/2)\|u-\mu_\parallel\|\) when the latter is at
least \(J\). All these cubes are outside the ball of radius \(J/2\).
Replacing the sum by the integral over those disjoint cubes gives
\begin{align*}
 2^{-m}\sum_{\mathrm{omitted}\ \mu}
                   \|u-\mu_\parallel\|^{1-d}
 &\le (3/2)^{m+d-1}
       \int_{\|y-u\|\ge J/2}\|y-u\|^{1-d}\,dy\\
 &= (3/2)^{m+d-1}\frac{2^{c-1}|S^{m-1}|}{c-1}J^{1-c}.
\end{align*}
This proves the finite-cloud derivative bound. All infinite sums are
absolute in the estimates used here. The actual mixture remains finite
and normalized throughout.

Choose \(J\) so large that
\begin{equation}
 C_{m,d}J^{1-c}<\tfrac12 A(2s_0,\theta).
 \label{eq:resolved_finite_sign_certificate}
\end{equation}
Then the actual score has the stated inward coordinate signs, uniformly
for \(s\in[s_0,2s_0]\), on every retained cell's slabs
\(1/12\le|u_j-k_j|\le1/2-\theta\). 

\subsection{A fine-cell capture controls the full future trajectory}
\label{app:resolved_endpoint_capture}

For retained indices \(\mathbf k\in\{-L+J,\ldots,L-J\}^m\), let
\[
 B_{\mathbf k}=\{(u,0):|u_j-k_j|\le1/2-\theta\ \text{for all }j\}.
\]
The closed cylinder over this cell with normal radius
\(r_{\mathrm{tr}}=s a_d\) is forward invariant. On every tangential
face, the corresponding coordinate derivative of \(\log p_{MN}\)
points strictly inward by \eqref{eq:resolved_finite_sign_certificate}.
On the normal boundary the exact affine identity gives
\(\dot n=-\beta(x)n\), so normal distance decreases. A first-exit
argument, also at intersections of faces, proves invariance.

Every path in this compact cylinder converges by the finite-mixture
lemma. Its limit has zero normal component and is critical. The same
coordinate signs exclude any critical point with an offset of magnitude
at least \(1/12\). Thus the endpoint lies in \(C_{\mathbf k}\).
Every coordinate of such an endpoint is at distance at least \(1/12\)
from every training coordinate in that dimension. Therefore
\begin{equation}
 \operatorname{dist}(x_\infty,\mathcal S_M)\ge\sqrt m/12,
 \qquad
 \|x_\infty-z\|\le7\sqrt m/12
 \quad(z\text{ in the full unit reference cell }\mathbf k).
 \label{eq:resolved_captured_endpoint_distances}
\end{equation}
This controls the endpoint by a \emph{sample-resolution} cell, not by
the center of an arbitrarily large data region. Uniqueness of the
endpoint inside each cell is neither needed nor asserted.
Maximizing the log density on each compact capture cylinder also gives
an interior local maximum: an inward derivative rules out a boundary
maximizer. Hence distinct retained cells contain distinct off-training
local maxima of the actual finite density, not only candidate reference
points.

The posterior bound is also explicit. At a covered endpoint, the
nearest training center belongs to its own cell. For one coordinate,
the squared-distance difference between the two positions \(k_j\pm1/6\)
is at most \(1/18\). At any fixed scale \(\sigma_i\ge s_0\), the
larger of their two normalized weights is at most
\((1+e^{-1/(36s_0^2)})^{-1}\). Taking the product over the \(m\)
coordinates bounds the largest conditional weight among its own
\(2^m\) centers. Including other cells only increases the denominator.
Finally, the precision-normalized center marginal is a convex combination
of the fixed-scale center posteriors. This proves
\eqref{eq:resolved_endpoint_center_nondominance}. The center maximizing
that marginal is still a nearest center, since each radial Gaussian
kernel decreases with distance. The same bound holds throughout every
output cell, including normal offsets.

\subsection{Explicit probability errors and a nonempty bandwidth window}
\label{app:resolved_nonempty_window}

We verify that the geometry and the probability requirements can hold
simultaneously for one finite cloud and a whole positive-bandwidth
interval. Continue with unit spacing. Fix \(\epsilon\in(0,1/4)\), set
\(\zeta=\epsilon/8\), and use the associated \(a_d\). Choose
\begin{equation}
 s_0\ge\max\left\{\frac12,
        \sqrt{\frac{\log(128m/\epsilon)}{2\pi^2}}\right\},
 \qquad\theta=\frac{\epsilon}{32m}.
 \label{eq:resolved_explicit_smoothing_floor}
\end{equation}
Choose a finite integer \(J\) satisfying
\eqref{eq:resolved_finite_sign_certificate}, and then \(L>J\) so large
that
\begin{equation}
 2L+1\ge\max\left\{\frac{32mJ}{\epsilon},\
                         \frac{256ms_0}{\epsilon}\right\}.
 \label{eq:resolved_explicit_array_size}
\end{equation}
These are finite choices: \(A(2s_0,\theta)>0\), and
\(J^{1-c}\to0\) for \(c\ge2\). This explicit order of choices
establishes nonemptiness; no incompatible lower and upper bandwidth
requirements are hidden in the statement.

Let the fixed relative cutoff ratio \(\Lambda>1\) satisfy
\begin{equation}
 \Lambda\ge\max\left\{2,\
       \frac{8\sqrt m(L+1/2)}{\sqrt{2\pi}\,\epsilon s_0}\right\}.
 \label{eq:resolved_relative_upper_cutoff}
\end{equation}
Choose finite \(N\) and
\(h=\log\Lambda/(N-1)\) with
\begin{equation}
 h\le\min\left\{1,\frac{\log2}{4},\
       \frac{\epsilon}{8(d\log\Lambda+\sqrt{2d}/2)}\right\}.
 \label{eq:resolved_finite_grid_choice}
\end{equation}
Use this same \(\Lambda,h,N\) for all \(s\in[s_0,2s_0]\), with
\(\sigma_i=s e^{(i-1)h}\). Thus \(\sigma_{\max}=\Lambda s\) may be much
larger than the data box. Its local influence has already been included
in the finite-array sign estimate, rather than excluded by an artificial
small upper cutoff.

\textbf{Reference cells and intrinsic smoothing.}
For the uniform distribution on a box of side \(2L+1\), translation by
\(v\in\mathbb R^m\) changes total variation by
\[
 1-\prod_{j=1}^m(1-|v_j|/(2L+1))_+
 \le\sum_j |v_j|/(2L+1).
\]
Averaging over \(v=sG_m\) gives
\begin{equation}
 e_{\mathrm{blur}}\le\frac{ms\sqrt{2/\pi}}{2L+1}\le\epsilon/8.
 \label{eq:resolved_intrinsic_blur_error}
\end{equation}
The probability excluded by the omitted boundary cells and the thin
strips between capture cells satisfies
\begin{equation}
 e_{\mathrm{cell}}\le\frac{2mJ}{2L+1}+2m\theta\le\epsilon/8.
 \label{eq:resolved_excluded_cell_mass}
\end{equation}
Indeed the exact retained fraction is
\([((2(L-J)+1)/(2L+1))(1-2\theta)]^m\), and
\(1-\prod_j(1-b_j)\le\sum_j b_j\). Also
\(e_{\mathrm{normal}}=\mathbb P[\|G_c\|>a_d]\le\zeta\), since the
\(c\)-dimensional norm is dominated by the \(d\)-dimensional one.

\textbf{The empirical fine smoothing is close to the reference smoothing.}
An elementary cellwise coupling sends each coordinate of a uniform
reference point to \(k-1/6\) on the left half-cell and to \(k+1/6\)
on the right. It has the correct empirical marginal and coordinate
error at most \(1/3\). Consequently
\begin{equation}
 W_1(\widehat Q_M,Q_{\mathcal M})\le\sqrt m/3.
 \label{eq:resolved_training_reference_coupling}
\end{equation}
For a sharper estimate at the lower cutoff, truncate the infinite
periodic product density to \(D_L\) and divide it by \((2L+1)^m\).
This is a probability density, because each coordinate interval contains
an integer number of periods and \(\int_0^1g_s=1\). On \(D_L\) the
finite-array smoothed density is no larger than this periodic density.
Their total-variation distance is therefore exactly the mass of the
finite smoothing outside \(D_L\).

The preceding coordinatewise coupling bounds that mass by
\(2m(s\sqrt{2/\pi}+1/3)/(2L+1)\): a point can leave the box only if
the coupled uniform coordinate is within \(|sG_1|+1/3\) of a boundary.
The periodic density differs from the uniform density by at most
\[
 \rho_m(s):=\frac12\left[
       \left(1+\frac{2q}{1-q}\right)^m-1\right],
 \qquad q=e^{-2\pi^2s^2}.
\]
Our choice of \(s_0\) gives \(q\le\epsilon/(128m)\) and
\(\rho_m(s)\le4mq\le\epsilon/32\). For example, use
\(2q/(1-q)\le3q\) and \(e^v-1\le2v\) for \(0\le v\le1\).
Adding \eqref{eq:resolved_intrinsic_blur_error}, and using \(s\ge1/2\),
yields
\begin{equation}
 \operatorname{TV}(\widehat p_s\,dx,p_s^Q\,dx)
 \le\rho_m(s)+\frac{4ms}{2L+1}\le\epsilon/16.
 \label{eq:resolved_fine_empirical_error}
\end{equation}
The ambient normal Gaussian is common to both measures, so their ambient
TV equals their intrinsic TV. This calculation improves the generic
\(W_1/\sigma_{\min}\) estimate for this particular resolved grid; it
does not assert that improvement for arbitrary training clouds.

\textbf{The source--sink and initialization errors.}
The residual decomposition used in
Proposition~\ref{prop:source_sink_input_errors} gives
\begin{align*}
 \epsilon_{\mathrm{Pois}}
 &\le dh\log\Lambda+\frac h2\sqrt{2d}
   +\operatorname{TV}(\widehat p_s\,dx,p_s^Q\,dx)
   +\operatorname{TV}(\widehat p_{\Lambda s}\,dx,p_{\Lambda s}^Q\,dx).
\end{align*}
The grid term is at most \(\epsilon/8\). At the upper cutoff,
\eqref{eq:source_sink_smoothed_wasserstein} and
\eqref{eq:resolved_training_reference_coupling} give TV at most
\(\sqrt m/(3\sqrt{2\pi}\Lambda s)\le\epsilon/16\).
For the \emph{untruncated} initialization
\(\nu=\mathcal N(0,(\Lambda s)^2I_d)\), the same Gaussian translation
calculation gives
\[
 \epsilon_{\mathrm{init}}
 \le\frac{\sqrt m(L+1/2)}{\sqrt{2\pi}\Lambda s}\le\epsilon/8.
\]
No compact-initialization condition is needed here: finite-mixture
convergence holds for every start, and the marked-distribution lemma is global.
Together with \eqref{eq:resolved_fine_empirical_error}, these bounds
imply
\begin{equation}
 \epsilon_{\mathrm{Pois}}+\epsilon_{\mathrm{init}}\le3\epsilon/8.
 \label{eq:resolved_source_error_budget}
\end{equation}
Thus the total error in \eqref{eq:resolved_joint_error_budget} is at
most \(3\epsilon/4<\epsilon\), uniformly on the entire window.

\subsection{Endpoint separation and regional coverage}
\label{app:resolved_endpoint_conclusions}

\begin{proof}[Proof of Corollary~\ref{cor:sample_resolution_nonmemorization}]
Take the finite constants and schedule just constructed. By finite-mixture
convergence and the affine normal identity, every trajectory has a
critical limit in \(\mathcal A\cap\operatorname{conv}(\mathcal S_M)\).
Apply the coupling in Appendix~\ref{app:resolved_mark_transfer} with
the capture cells verified above. Its good event has probability at
least \(1-\epsilon\), and on this event the actual endpoint lies in
\(C_{\mathbf k(Z)}\). The reference cell label is uniform on
\(\{-L,\ldots,L\}^m\), since \(Z\sim Q_{\mathcal M}\).
For any index collection \(I\), the two indicators
\[
 \mathbf1_{\{x_\infty\in\bigcup_{\mathbf k\in I}C_{\mathbf k}\}},
 \qquad\mathbf1_{\{\mathbf k(Z)\in I\}}
\]
agree on the good event. Their expectation difference has magnitude at
most \(\epsilon\), proving \eqref{eq:resolved_endpoint_cell_mass}.
Equivalently, the output-cell label distribution, with one extra label for points
outside all output cells, is within TV \(\epsilon\) of the uniform
reference label distribution. The reference assigns that extra label zero mass.

Restore \(\ell\) in \eqref{eq:resolved_captured_endpoint_distances}.
On the same good event,
\[
 \operatorname{dist}(x_\infty,\mathcal S_M)\ge\sqrt m\ell/12,
 \qquad \|x_\infty-Z\|\le7\sqrt m\ell/12.
\]
Each training coordinate has covering radius \(\ell/3\) on its
interval: the largest adjacent gap is \(2\ell/3\), and the outer
boundary has the same \(\ell/3\) distance. Coordinatewise nearest
points give \(h_{\mathrm{cov},M}=\sqrt m\ell/3\), with equality
at a grid-cell corner. This proves the stated separation and coupling
constants. A reference point in \(U^{-7h_{\mathrm{cov},M}/4}\) cannot
leave \(U\) under that distance bound. Subtracting the failure
probability proves the \emph{joint} regional inequality
\eqref{eq:resolved_endpoint_population_coverage}.

For the count of populated cells, let \(I\) contain those with probability
less than \(1/[2(2L+1)^m]\). Their total deficit exceeds
\(|I|/[2(2L+1)^m]\), which \eqref{eq:resolved_endpoint_cell_mass}
bounds by \(\epsilon\). Thus at least
\((1-2\epsilon)(2L+1)^m\) cells have the claimed mass. Each cell
contains endpoints with positive training distance. In addition, every
retained capture cylinder contains an open set of positive Gaussian
initialization probability, so it supplies an endpoint in its own
output cell independently of the quantitative count.

Finally, both the endpoint and \(Z\) lie in \(D_L\), by the convex-hull
identity for the former. Their coupled expected distance is at most
\(7h_{\mathrm{cov},M}/4+\epsilon\operatorname{diam}(D_L)\), proving
\eqref{eq:resolved_endpoint_wasserstein}. The posterior bound was proved
in Appendix~\ref{app:resolved_endpoint_capture}. All events are
measurable because the limit map is the pointwise limit of continuous
integer-time flow maps.
\end{proof}

\paragraph{A positive training-distance margin for every start.}
For each fixed certified cloud and relative schedule, a smaller closed bandwidth interval $I\subset(s_0,2s_0)$ of positive length admits a uniform $\eta>0$ with $\operatorname{dist}(x_\infty,\mathcal S_M)\ge\eta$ for every start. To see this in unit-spacing coordinates, put $t_i=e^{(i-1)h}$ and denote the density at scales $s t_i$ by $p_{MN}^{(s)}$. Each function
$f_j(s)=\langle\mu_j,\nabla\log p_{MN}^{(s)}(\mu_j)\rangle$ is analytic on $(0,\infty)$, and the finite Gaussian expansion gives
\[
 s^2\nabla\log p_{MN}^{(s)}(\mu_j)
 \longrightarrow-\frac{\sum_i t_i^{-d}}{\sum_i t_i^{2-d}}\mu_j
 \quad(s\to\infty).
\]
The centered grid contains no zero training center, so no $f_j$ is identically zero. Their finitely many zeros on $[s_0,2s_0]$ leave such an interval $I$. The training-center gradient norms have a strictly positive minimum over $j$ and $s\in I$. Uniform continuity extends this nonvanishing property to common neighborhoods of the centers, excluding critical points there. Every endpoint is critical. Restoring $\ell$ rescales the margin; this every-start margin need not be comparable to $h_{\mathrm{cov},M}$, unlike the explicit high-probability margin.

\paragraph{A continuous population limit.}
Choose $\epsilon_n\downarrow0$ and a certified
$s_{0,n},J_n,L_n$ for each $n$, with $L_n\to\infty$.
Setting $\ell_n=(2L_n+1)^{-1}$ keeps the reference box fixed at
$[-1/2,1/2]^m\times\{0\}$, and
\eqref{eq:resolved_endpoint_wasserstein} gives convergence to its
continuous reference distribution in $W_1$.
Each finite model uses a positive bandwidth fixed throughout generation
and retains a positive training-distance threshold
$\sqrt m/[12(2L_n+1)]$, proportional to its training covering radius.

\paragraph{Small perturbations of a fixed certified cloud.}
The sign inequalities have a strict positive margin on a finite union
of compact slabs. For fixed finite \(L,N\) and positive cutoffs, the
finite Gaussian density and its derivatives depend continuously on the
training positions, uniformly on those compact sets. Therefore sufficiently
small perturbations \emph{within the same affine space} preserve the
capture and core-exclusion signs. Moving every center by less than
\(h_{\mathrm{cov},M}/8\) preserves a separation of at least
\(h_{\mathrm{cov},M}/8\), where this displayed radius denotes the
unperturbed grid value. The residual is also continuous in \(L^1\) under
these perturbations, by Gaussian translation continuity and finiteness of
the mixture. Hence the spare error margin preserves the cell-mass and
coverage bounds under sufficiently small within-subspace perturbations,
with either a slightly enlarged failure budget or the original budget after
starting from a smaller one. Thus the guarantees persist on an open
neighborhood of each certified finite cloud.

\paragraph{Finite-resolution interpretation.}

For a finite Gaussian mixture, an open \(m\)-dimensional critical plateau
on \(\mathcal A\) is impossible: the restricted density is analytic;
vanishing derivatives on an open set would make it constant on the
connected affine space, contradicting Gaussian decay at infinity.
If its critical points are all isolated, compactness of their convex-hull
location makes their number finite. In that common nondegenerate case,
the endpoint distribution is discrete. Neither fact prevents the many-cell,
finite-resolution coverage and continuous \emph{limit of distributions} proved
above. No continuum of endpoints for one fixed empirical mixture is
assumed.

\paragraph{Interpretation of the bandwidth window.}
The sufficient lower floor controls sample-scale contrast and the
fine-smoothing approximation. A larger floor weakens the useful periodic
modulation and therefore requires a smaller finite-array perturbation in
\eqref{eq:resolved_finite_sign_certificate}. For a fixed finite array this
creates an upper restriction on the certified floor. Choosing \(J\) and then
\(L\) as above leaves a nonempty interval between these requirements.
The constants are conservative, and weak modulation may lead to slow final
tangential convergence; nevertheless, the result establishes an exact
positive-bandwidth window for the finite-cloud family.

\subsection{Compatibility with a finite reach}
For a curved instance satisfying Theorem~\ref{thm:manifold_convergence},
the additional normal-accuracy requirement is

\begin{equation}
 \varepsilon_{\mathcal M}
 =\frac{C_{\mathrm{loc}}m}{2\tau}\sigma_{\min}^2
 \le\varepsilon
 \quad\Longleftarrow\quad
 \sigma_{\min}\le
 \sqrt{\frac{2\tau\varepsilon}{C_{\mathrm{loc}}m}}.
 \label{eq:goldilocks_geometric_upper_bound}
\end{equation}

Here \(\varepsilon\) is a prescribed normal tolerance. Increasing reach
relaxes this sufficient upper bound only when the actual localization
constant and the other hypotheses remain controlled. For example, an
independently established smoothing window \([\sigma_0,2\sigma_0]\)
meets this tolerance throughout if
\(C_{\mathrm{loc}}\le C_*\) and
\(\tau\ge2C_*m\sigma_0^2/\varepsilon\), along with the tube and
landscape conditions. This is a compatibility calculation, not a claim
that arbitrary curved data inherit the affine retention condition.
In the affine case the normal error is exactly zero, so only the
sample-wise and regional requirements remain. Large reach does not by
itself supply separation between distinct data regions.

\subsection{Comparison with the zero-noise empirical diffusion endpoint}
\label{subsec:diffusion_endpoint_comparison}

We compare with exact empirical reverse diffusion initialized from its
correct terminal noisy marginal. At noise level $\sigma$, its prescribed
marginal is $\widehat p_\sigma$ \citep{song2020score}. For a uniform
training index $J$ independent of $G\sim\mathcal N(0,I_d)$,
\[
Y_\sigma\overset{\mathrm d}=\mu_J+\sigma G.
\]
Consequently, for every fixed $\eta>0$,
\[
 \mathbb P[\operatorname{dist}(Y_\sigma,\mathcal S_M)>\eta]
 \le
 \mathbb P[\sigma\|G\|>\eta]
 \longrightarrow0,
 \qquad
 P_{Y_\sigma}\Longrightarrow\widehat Q_M
 \quad(\sigma\to0).
\]
The distance bound follows from
$\operatorname{dist}(\mu_J+\sigma G,\mathcal S_M)
\le\sigma\|G\|$, while weak convergence follows by dominated convergence
against bounded continuous test functions. Thus the exact empirical
zero-noise diffusion marginal recovers the empirical center distribution
$\widehat Q_M$.

\par\begingroup
\paragraph{Positive-cutoff diffusion followed by the same flow.}
Let $T(x)=\lim_{t\to\infty}\phi_t(x)$ be the endpoint map of the exact unified-score flow. The point $Y$ in Lemma~\ref{lemma:source_sink_mark} lies on the trajectory of $X_0$, so $T(Y)=T(X_0)$. Applying the same measurable map to the two laws in \eqref{eq:marked_fine_scale_law} gives
\begin{equation}
 \operatorname{TV}\!\left(\mathcal L(T(X_0)),T_\#(p_{\sigma_{\min}}^Q(x)\,dx)\right)
 \le\epsilon_{\mathrm{Pois}}+\epsilon_{\mathrm{init}}.
 \label{eq:cutoff_endpoint_transfer}
\end{equation}
\par\endgroup
The second law starts from the positive-cutoff diffusion marginal of $Q_{\mathcal M}$ and refines with the same noise-unconditioned field. For $Q_{\mathcal M}=\widehat Q_M$, the starting law is the empirical diffusion marginal at $\sigma_{\min}$. Both constructions share its cross-scale smoothing \citep{zhou2026smoothing} and its positive bandwidth. The zero-noise comparison above follows the empirical diffusion field all the way to the training distribution.

Regional coverage describes allocation of probability at the chosen resolution and can also hold for a discrete reference law. Corollary~\ref{cor:main_nonmemorization} adds separation from the training set for the same endpoints. Those endpoints can be atomic for a fixed cloud. The stationary-point equation places them in the convex hull of the centers, which lies in the reference support of the affine construction. This gives new on-support points at a controlled distance from the training samples. When local reach is large relative to the smoothing neighborhood, the normal bias bound suggests how interpolation can produce new points near the local data geometry. Empirical studies of local low-dimensional structure in diffusion denoisers and score Jacobians \citep{chen2024subspaces,ventura2025manifolds} motivate this affine model as a local geometric case.

Corollary~\ref{cor:main_nonmemorization} describes a different
positive-bandwidth regime. The smoothing floor remains fixed throughout
optimization, yet the same infinite-time endpoints satisfy a
training-distance margin proportional to the training resolution together
with regional and cell-mass coverage. In the constructive affine family,
positive smoothing can therefore retain sample-resolution diversity without
forcing the empirical-score endpoints back onto the training centers.

A positive-cutoff posterior-mean readout provides a related geometric
comparison. Direct differentiation gives
\[
 y+\sigma^2\nabla\log\widehat p_\sigma(y)
 =
 \frac{\sum_j\mu_j\mathcal N(y;\mu_j,\sigma^2I)}
      {\sum_j\mathcal N(y;\mu_j,\sigma^2I)}
 \in\operatorname{conv}(\mathcal S_M).
\]
Hence, when the training centers lie in the affine space $\mathcal A$, this
readout also lies in $\mathcal A$. The unified-score construction studied
here additionally controls the infinite-time endpoint distribution itself:
for the certified finite-cloud family, the endpoints remain separated from
the training set while retaining regional probability at the training
resolution. Along the fixed-box sequence constructed above, these endpoint
distributions further converge in $W_1$ to a continuous reference
distribution.

\clearpage
\section{Solvable Models}
\label{sec:solvable_models}

The flat calculation identifies the posterior moment used in normal contraction. The sphere exhibits its reach dependence, unstable off-tube geometry, and exact coverage under symmetry. The proper sphere mixtures are population instances of our framework; the translation-invariant flat model is a local reference, not a probability distribution.

\paragraph{Proper population mixtures.}
For compactly supported $Q_{\mathcal M}$ and a finite nonzero scale measure $\pi$ on $[\sigma_{\min},\sigma_{\max}]\subset(0,\infty)$, define
\begin{equation}
 p^{\mathrm{pop}}(x)=
 \frac{\int\sigma^2\int\mathcal N(x;\mu,\sigma^2I)Q_{\mathcal M}(d\mu)\pi(d\sigma)}
 {\int\sigma^2\pi(d\sigma)}.
 \label{eq:proper_population_mixture}
\end{equation}
Tonelli's theorem gives a positive normalized density. Bounded centers and positive cutoffs make the Gaussian power series locally uniformly convergent over the mixing variables, so the density and its logarithm are analytic. With $B=\sup\|\mu\|$, its tail is bounded by
$(2\pi\sigma_{\min}^2)^{-d/2}\exp[-(\|x\|-B)^2/(2\sigma_{\max}^2)]$
for $\|x\|\ge B$. Thus positive superlevels are compact, and the analytic convergence proof applies. Posterior sums become integrals in the geometric arguments; Appendix~\ref{app:population_extension} gives the full population convergence and coverage statements.

The unbounded log-scale measure on $(0,\infty)$ has infinite mixture normalizer and is used only for reference kernels and posteriors away from the data. The flat model also has an improper translation-invariant center measure. Neither is substituted for a proper sampling density.

\subsection{Flat model: posterior moments and finite schedules}
\label{subsec:flat_model}

Take \(\mathcal M_{\flat}=\mathbb R^m\times\{0\}\) with Lebesgue
base measure. Write \(x=(x_\parallel,x_\perp)\),
\(r=\|x_\perp\|>0\), and \(z_x=(x_\parallel,0)\). Conditional on
any fixed \(\sigma\), the precision-weighted center posterior is
\(\mathcal N(x_\parallel,\sigma^2I_m)\). Its conditional second
moment around \(z_x\) is \(m\sigma^2\).

\begin{proposition}[Exact unbounded-reference localization constant]
\label{prop:flat_exact_cloc}
With scale measure \(d\sigma/\sigma\) on \((0,\infty)\), the
precision-weighted posterior is proper for every \(r>0\), and
\begin{equation}
 u:=\frac{r^2}{2\sigma^2}\sim\operatorname{Gamma}(c/2,1).
 \label{eq:flat_scale_gamma}
\end{equation}
For \(c>2\),
\begin{equation}
 \mathbb E_{\widetilde w}\|\mu-z_x\|^2
 =m\mathbb E_{\widetilde w}\sigma^2
 =\frac{mr^2}{c-2}.
 \label{eq:flat_exact_second_moment}
\end{equation}
Thus the distance-matched moment bound has the sharp constant
\begin{equation}
 C_{\mathrm{loc}}=\frac{c}{c-2}.
 \label{eq:flat_cloc_closed_form}
\end{equation}
\end{proposition}
\begin{proof}
At fixed $\sigma$, normalizing the tangent Gaussian gives
$\mu_\parallel\mid x,\sigma\sim\mathcal N(x_\parallel,\sigma^2I_m)$,
with conditional squared displacement $m\sigma^2$. Integrating the tangent coordinates leaves the scale weight
$\sigma^{-c}e^{-r^2/(2\sigma^2)}d\sigma/\sigma$.
Under $u=r^2/(2\sigma^2)$ this is proportional to $u^{c/2-1}e^{-u}du$, with finite Gamma normalizer for $r>0,c>0$. For $c>2$,
\[
 \mathbb E\sigma^2=\frac{r^2}2\mathbb E[u^{-1}]
 =\frac{r^2}2\frac{\Gamma(c/2-1)}{\Gamma(c/2)}
 =\frac{r^2}{c-2}.
\]
Multiplication by $m$ gives the result, and comparison with $mr^2/c$ gives the exact constant.
\end{proof}
The mode squared is \(r^2/c\), whereas the mean squared scale is
\(r^2/(c-2)\). Their ratio explains the moment constant without
identifying a mode with a mean.

\begin{proposition}[Cutoff dependence in codimensions one and two]
\label{prop:flat_failure}
For \(c=1,2\), with a truncated log-continuous schedule satisfying
\(\sigma_{\min}\le r/\sqrt c\) and \(\sigma_{\max}\ge2r\),
\begin{equation}
 \mathbb E_{\widetilde w}\|\mu-z_x\|^2\asymp m
 \begin{cases}
 r\sigma_{\max},&c=1,\\
 r^2\log(\sigma_{\max}/r),&c=2,
 \end{cases}
 \label{eq:flat_low_codimension_cutoff}
\end{equation}
with absolute comparison constants. Thus no upper-cutoff-independent
constant controls the distance-matched second moment as
\(\sigma_{\max}/r\to\infty\).
\end{proposition}
\begin{proof}
The same substitution truncates the Gamma integrals to
$u_-=r^2/(2\sigma_{\max}^2)\le1/8$ and
$u_+=r^2/(2\sigma_{\min}^2)\ge c/2$. Thus
\begin{equation}
 \mathbb E\sigma^2=\frac{r^2}2
 \frac{\int_{u_-}^{u_+}u^{c/2-2}e^{-u}du}
      {\int_{u_-}^{u_+}u^{c/2-1}e^{-u}du}.
 \label{eq:detailed_truncated_inverse_moment}
\end{equation}
For $c=1,2$, the denominator lies between the positive integral on $[1/8,1/2]$ and $\Gamma(c/2)$. For $c=1$, bounding $e^{-u}$ from below on $[u_-,1/2]$ and from above by one gives a numerator comparable to $u_-^{-1/2}$. For $c=2$, integration on $[u_-,1]$ and a bounded exponential tail give a numerator comparable to $\log(1/u_-)$. Substitution yields respectively $r\sigma_{\max}$ and $r^2\log(\sigma_{\max}/r)$. Multiply by $m$.
\end{proof}
This is a threshold for a cutoff-independent moment in this model,
not an impossibility theorem for finite schedules in low codimension.
Nor does it say that the contraction gap must fail whenever \(c\le2\).

\begin{lemma}[Discrete geometric schedules]
\label{lemma:discrete_schedule}
Let \(c\ge3\), \(\sigma_i=\sigma_{\min}\rho^{i-1}\),
\(\sigma_N=\sigma_{\max}\), \(\rho>1\), and
\[
 h:=\log\rho\le\frac1{27\sqrt c},\qquad
 \sigma_{\max}^2\ge\frac{8c}{c-2}\sigma_{\min}^2.
\]
Then the flat-model posterior satisfies, for
\(0<r\le\sigma_{\max}\sqrt{(c-2)/2}\),
\begin{equation}
 \mathbb E_{\widetilde w}\|\mu-z_x\|^2
 \le\bigl(1+30(h\sqrt c+e^{-c/2})\bigr)
        \frac{m\max\{r^2,4c\sigma_{\min}^2\}}{c-2}.
 \label{eq:discrete_schedule_bound}
\end{equation}
In particular, the choice
\[
 C_{\mathrm{loc}}:=
 (1+30(h\sqrt c+e^{-c/2}))\frac{4c}{c-2}
\]
is valid on this range.
\end{lemma}
\begin{proof}
It is enough to bound the scale mean, since the conditional center moment is $m\sigma^2$.

\textbf{Reduction to radii above the lower cutoff.}
Differentiating with respect to $v=r^2/2$ gives
$\frac d{dv}\mathbb E\sigma^2=-\operatorname{Cov}(\sigma^2,\sigma^{-2})
=\mathbb E\sigma^2\mathbb E\sigma^{-2}-1\ge0$.
Set $r_\dagger=2\sqrt c\,\sigma_{\min}$, which is in the admissible range by the scale-range assumption. It suffices to treat $r\ge r_\dagger$ and then use monotonicity.

\textbf{Riemann sums relative to their integrals.}
For $k=c,c-2$, let $f_k(t)=\exp(-kt-r^2e^{-2t}/2)$,
$I_k=\int_{\log\sigma_{\min}}^{\log\sigma_{\max}}f_k(t)dt$,
and $S_k=h\sum_i f_k(\log\sigma_i)$. Unimodality gives
$|S_k-I_k|\le3h\sup f_k$: two peak heights bound total variation and one accounts for the extra endpoint. Both modes $t_k^*=\log(r/\sqrt k)$ are at least $(\log2)/2$ from the integration boundaries.
For $|v|\le1/2$,
\[
 \log f_k(t_k^*)-\log f_k(t_k^*+v)
 =k[v+(e^{-2v}-1)/2]\le ekv^2.
\]
Integrating on $|v|\le\min\{1/(2\sqrt k),1/3\}$ gives
$I_k\ge\sup f_k/(3\sqrt k)$. Therefore, for $b=h\sqrt c\le1/27$,
\[
 \frac{S_{c-2}}{S_c}
 \le\frac{1+9b}{1-9b}\frac{I_{c-2}}{I_c}
 \le(1+27b)\,\mathbb E_{\mathrm{cont,trunc}}\sigma^2.
\]

\textbf{Truncation of the scale posterior.}
For $u\sim\operatorname{Gamma}(c/2,1)$, the upper cutoff in $u$ is
$u_+=r^2/(2\sigma_{\min}^2)\ge2c$. Removing the lower-$u$ part only decreases the conditional mean of $u^{-1}$. Moreover,
$\mathbb P(u>u_+)\le e^{-c}\mathbb E e^{u/2}=e^{-c}2^{c/2}\le e^{-c/2}$.
Hence
\[
 \mathbb E_{\mathrm{cont,trunc}}\sigma^2
 \le\frac{r^2}{c-2}\frac1{1-e^{-c/2}}.
\]
For $b\le1/27$ and $a=e^{-c/2}<1/4$,
$(1+27b)/(1-a)\le1+30(b+a)$, as is seen by multiplying by $1-a$.
This proves the bound for $r\ge r_\dagger$. Monotonicity replaces $r^2$ by $r_\dagger^2$ below that threshold. Finally,
$\max\{r^2,4c\sigma_{\min}^2\}\le4c\max\{r^2/c,\sigma_{\min}^2\}$,
giving the stated localization constant.
\end{proof}
The grid condition is sufficient, not necessary. E1 uses \(24\) scales
on two decades, with \(h\sqrt c\approx0.6\), outside this sufficient
range. Its downstream diagnostics do not certify this particular
uniform moment bound.

\begin{remark}[Limits of the flat idealization]
\label{remark:flat_model_limits}
For finitely many centers in an affine subspace, normal contraction
is exact by
Proposition~\ref{prop:affine_subspace_exact_normal_convergence}.
The translation-invariant flat calculation instead describes a local
posterior and is not a normalizable data distribution. In particular,
its unbounded-scale singular field should not be substituted into the
finite-mixture global-flow theorem. Curvature and off-tube instability
are addressed by the sphere model next.
\end{remark}

\subsection{Sphere model: curvature, localization, and exact regional coverage}
\label{subsec:sphere_model}

Let \(\mathcal M_\circ=S^m(R)\subset V:=\mathbb R^{m+1}\times\{0\}
\subset\mathbb R^d\), with uniform reference distribution, reach \(R\), and codimension \(c=d-m\).
Use the proper population density \(p^{\mathrm{pop}}\) from
\eqref{eq:proper_population_mixture} for this uniform distribution and the stated
bounded scale schedule; reference-limit kernels are distinguished explicitly. Write \({P_V}\) for the orthogonal
projector onto \(V\), not for the nonlinear sphere projection
\(\Pi_{\mathcal M}\). The scalar \(\varrho=\|{P_V} x\|\) denotes subspace
radius in the sphere calculations.

\begin{proposition}[Sharpness of the reach inequality]
\label{prop:reach_saturation}
For \(z,\mu\in S^m(R)\), set \(e=z/R\). Then
\[
 \langle e,\mu-z\rangle=-\frac{\|\mu-z\|^2}{2R}.
\]
Consequently the positive-reach upper bound is attained for the inward
normal \(u=-e\), including after averaging over any posterior.
\end{proposition}
\begin{proof}
For $\|\mu\|=\|z\|=R$, $\|\mu-z\|^2=2R^2-2\langle\mu,z\rangle$.
Dividing by $2R$ gives the identity for $e=z/R$ and its positive version for $-e$. Averaging preserves the equality.
\end{proof}
This shows sharpness of the normal chord coefficient \(1/(2\tau)\),
not sharpness of every subsequent bound on a trajectory.

\begin{proposition}[Sphere localization by an exact flat-model comparison]
\label{prop:sphere_localization}
Suppose \(m\ge2\). Use the same finite or truncated-continuous scale
schedule in the sphere and flat models. For any \(x\) with
\(0<r(x)<R\), write \(r=r(x)\), set \(\varrho=\|{P_V} x\|>0\), and
\(z_x=R{P_V} x/\varrho\). Then

\begin{equation}
 \mathbb E_{\widetilde w,\mathrm{sph}}\|\mu-z_x\|^2
 \le\frac R{\varrho}\,m\,
          \mathbb E_{\widetilde w,\mathrm{flat},r(x)}\sigma^2.
 \label{eq:sphere_flat_moment_comparison}
\end{equation}

Hence, if the flat-model bound is valid with constant
\(C_{\mathrm{flat}}\) on \(0<r<r_0<R\), the sphere moment bound
is valid with the choice
\begin{equation}
 C_{\mathrm{loc}}:=\frac{R}{R-r_0}C_{\mathrm{flat}}.
 \label{eq:sphere_valid_localization_constant}
\end{equation}
For the unbounded reference posterior and \(c>2\),
\begin{equation}
 \mathbb E_{\widetilde w,\mathrm{sph}}\|\mu-z_x\|^2
 \le\frac R{\varrho}\frac{mr^2}{c-2},
 \qquad
 \mathbb E_{\widetilde w,\mathrm{sph}}\|\mu-z_x\|^2
 =\frac R{\varrho}\frac{mr^2}{c-2}(1+o(1))
 \quad(r/R\to0,\ m,c\text{ fixed}).
 \label{eq:sphere_reference_local_limit}
\end{equation}

The local-limit equality does not assert the same leading constant
uniformly for arbitrary finite cutoffs.
\end{proposition}
\begin{proof}
We compare the sphere to a flat posterior at the same normal distance
and with exactly the same scale measure. All auxiliary quantities in
this proof are local to this comparison: set
\[
 \varrho:=\|{P_V} x\|,\quad k:=\varrho/R,\quad a:=m/2,\quad
 z_x:=R{P_V} x/\varrho,\quad v:=\|\mu-z_x\|^2.
\]

Because \(r(x)<R\), \(\varrho>0\): if \({P_V} x=0\), its distance to
any sphere point is \(\sqrt{R^2+\|(I-{P_V})x\|^2}\ge R\).

\textbf{Exact chord coordinates.}
Writing \(\theta\) for the angle between \(\mu\) and \(z_x\),
\[
 v=2R^2(1-\cos\theta),\qquad
 r^2=(\varrho-R)^2+\|(I-{P_V})x\|^2.
\]

Therefore
\begin{align*}
 \|x-\mu\|^2
 &=\varrho^2+R^2-2\varrho R\cos\theta+\|(I-{P_V})x\|^2\\
 &=r^2+2\varrho R(1-\cos\theta)=r^2+kv.
\end{align*}

This is why the argument covers nonradial ambient points as well.
Uniform spherical measure has angular factor
\(\sin^{m-1}\theta\,d\theta\). Since
\[
 dv=2R^2\sin\theta\,d\theta,\qquad
 \sin^2\theta=\frac v{R^2}\left(1-\frac v{4R^2}\right),
\]
its density in \(v\), up to a positive constant independent of
\(v\) and \(\sigma\), is
\[
 v^{a-1}\left(1-\frac v{4R^2}\right)^{a-1}\,dv,
 \qquad 0<v<4R^2.
\]
Define
\[
 \chi(v):=\begin{cases}
 (1-v/(4R^2))^{a-1},&0\le v<4R^2,\\
 0,&v\ge4R^2.
 \end{cases}
\]
Because \(m\ge2\), we have \(a\ge1\), so \(0\le\chi\le1\)
and \(\chi\) is nonincreasing. This sign is the only reason for the
restriction \(m\ge2\).

\textbf{The fixed-scale chord moment.}
At fixed \(\sigma\), the sphere posterior in \(v\) is proportional to
\(v^{a-1}e^{-kv/(2\sigma^2)}\chi(v)\,dv\). Without \(\chi(v)\), this is
a Gamma distribution of shape \(a\) and scale \(2\sigma^2/k\), whose mean
is \(2a\sigma^2/k=m\sigma^2/k\).
For any integrable random variable \(Y\) and an independent copy
\(Y'\), a bounded nonincreasing weight satisfies
\[
 2\operatorname{Cov}(Y,\chi(Y))
 =\mathbb E[(Y-Y')(\chi(Y)-\chi(Y'))]\le0.
\]
Every integrand is nonpositive. Thus, when \(\mathbb E\chi(Y)>0\),
\[
 \frac{\mathbb E[Y\chi(Y)]}{\mathbb E\chi(Y)}
 =\mathbb E Y+\frac{\operatorname{Cov}(Y,\chi(Y))}{\mathbb E\chi(Y)}
 \le\mathbb E Y.
\]
Apply this identity to the fixed-scale Gamma distribution. Its tilt has a
positive normalizer, and we obtain
\begin{equation}
 \mathbb E_{\mathrm{sph}}[v\mid\sigma]\le\frac m{k}\sigma^2.
 \label{eq:detailed_sphere_conditional_chord}
\end{equation}

\textbf{Comparing the scale posteriors.}
For the common scale measure \(\pi(d\sigma)\), integration in \(v\)
gives the sphere scale weight
\[
 \sigma^{-d}e^{-r^2/(2\sigma^2)}
 \left[\int_0^\infty v^{a-1}e^{-kv/(2\sigma^2)}\chi(v)\,dv\right]
 \pi(d\sigma).
\]
Set \(v=2\sigma^2u/k\) inside this integral and define
\[
 \omega_{\mathrm{sph}}(\sigma):=\int_0^\infty u^{a-1}e^{-u}\chi(2\sigma^2u/k)\,du.
\]

The bracket becomes \((2\sigma^2/k)^a\omega_{\mathrm{sph}}(\sigma)\). Since
\(2a=m\), the full scale weight is, up to a scale-independent factor,

\[
 \sigma^{-c}e^{-r^2/(2\sigma^2)}\omega_{\mathrm{sph}}(\sigma)\,\pi(d\sigma).
\]

The flat scale weight is exactly the same expression without
\(\omega_{\mathrm{sph}}(\sigma)\). Moreover, \(\omega_{\mathrm{sph}}\) is nonincreasing: increasing
\(\sigma\) decreases \(\chi(2\sigma^2u/k)\) at every \(u\).
The covariance inequality above, now with the increasing variable
\(\sigma^2\) and decreasing weight \(\omega_{\mathrm{sph}}\), gives

\begin{align*}
 \mathbb E_{\mathrm{sph}}\sigma^2
 &=\frac{\mathbb E_{\mathrm{flat},r}[\sigma^2\omega_{\mathrm{sph}}(\sigma)]}
         {\mathbb E_{\mathrm{flat},r}\omega_{\mathrm{sph}}(\sigma)}\\
 &\le\mathbb E_{\mathrm{flat},r}\sigma^2.
\end{align*}

Taking expectations in \eqref{eq:detailed_sphere_conditional_chord}
therefore yields
\[
 \mathbb E_{\mathrm{sph}}\|\mu-z_x\|^2
 =\mathbb E_{\mathrm{sph}}v
 \le\frac m{k}\mathbb E_{\mathrm{sph}}\sigma^2
 \le\frac R{\varrho}\,m\mathbb E_{\mathrm{flat},r}\sigma^2,
\]

which is \eqref{eq:sphere_flat_moment_comparison}.
If \(r<r_0\), then \(|\varrho-R|\le r<r_0\), so \(\varrho>R-r_0\).
A valid flat constant consequently gives

\begin{align*}
 \mathbb E_{\mathrm{sph}}\|\mu-z_x\|^2
 &\le\frac R{\varrho}C_{\mathrm{flat}}m
                         \max\{\sigma_{\min}^2,r^2/c\}\\
 &\le\frac R{R-r_0}C_{\mathrm{flat}}m
                         \max\{\sigma_{\min}^2,r^2/c\}.
\end{align*}

This proves the stated choice of \(C_{\mathrm{loc}}\).

\textbf{The unbounded reference and its local limit.}
For \(c>2\), the flat reference scale moment is finite, so the same
comparison applies and gives \(\mathbb E_{\mathrm{sph}}v
\le (R/\varrho)mr^2/(c-2)\).
For the asymptotic equality, integrating the precision weight over
\(\sigma\in(0,\infty)\) gives the chord density proportional to

\[
 v^{a-1}\chi(v)(r^2+kv)^{-d/2}\,dv.
\]
This follows by the same Gamma substitution as in the flat proof,
with \(r^2+kv\) in place of \(r^2\).
Substituting \(v=r^2u/k\) in its numerator and denominator yields
\[
 \mathbb E_{\mathrm{sph}}v
 =\frac{r^2}{k}
   \frac{\displaystyle\int_0^\infty
       u^a(1+u)^{-d/2}\chi(r^2u/k)\,du}
        {\displaystyle\int_0^\infty
       u^{a-1}(1+u)^{-d/2}\chi(r^2u/k)\,du}.
\]
As \(r/R\to0\), \(|\varrho-R|\le r\) implies \(k\to1\), and
\(\chi(r^2u/k)\to1\) for each fixed \(u\). Both integrands are
dominated by their versions with \(\chi=1\). The denominator is
integrable for \(c>0\), and the numerator for \(c>2\); these
are precisely the required hypotheses for dominated convergence.
Using the beta integral, their limiting ratio is

\begin{align*}
 \frac{B(a+1,c/2-1)}{B(a,c/2)}
 &=\frac{\Gamma(a+1)\Gamma(c/2-1)}
          {\Gamma(a)\Gamma(c/2)}\\
 &=\frac a{c/2-1}=\frac m{c-2}.
\end{align*}
Here \(B(p,q)=\int_0^\infty u^{p-1}(1+u)^{-p-q}\,du
=\Gamma(p)\Gamma(q)/\Gamma(p+q)\) for positive \(p,q\).
Since \(1/k=R/\varrho\), this proves the local-limit equality in
\eqref{eq:sphere_reference_local_limit}. It does not replace the
finite-range moment by the unbounded moment without a truncation bound.

\end{proof}

\begin{remark}[Usable finite-range constants]
\label{remark:sphere_explicit_constants}
For \(m\ge2,c\ge3\), assume the schedule hypotheses of
Lemma~\ref{lemma:discrete_schedule} and
\(r_0\le\sigma_{\max}\sqrt{(c-2)/2}\). One valid flat constant is
\[
 C_{\mathrm{flat}}
 :=\bigl(1+30(h\sqrt c+e^{-c/2})\bigr)\frac{4c}{c-2}.
\]
The sphere comparison then gives the valid uniform choice
\[
 C_{\mathrm{loc}}:=\frac{R}{R-r_0}C_{\mathrm{flat}}.
\]
For this choice, substituting \(\tau=R\) and
the definition of \(\varepsilon_0\) yields

\begin{align*}
 \varepsilon_0
 &=\frac{C_{\mathrm{loc}}m}{2R}\sigma_{\min}^2
 =\frac{C_{\mathrm{flat}}m}{2(R-r_0)}\sigma_{\min}^2,\\
 \kappa_0
 &=\frac{C_{\mathrm{loc}}mr_0}{2cR}
 =\frac{C_{\mathrm{flat}}mr_0}{2c(R-r_0)}.
\end{align*}
The contraction gap can therefore be checked without an implicit
substitution:
\[
 \kappa_0<1
 \quad\Longleftrightarrow\quad
 C_{\mathrm{flat}}mr_0<2c(R-r_0)
 \quad\Longleftrightarrow\quad
 \frac{r_0}{R}<\frac{2c}{C_{\mathrm{flat}}m+2c}.
\]
These are valid finite-schedule constants at every point of the tube,
including points outside the sphere's containing subspace.
The value \(c/(c-2)\) remains the flat unbounded-reference benchmark;
it must not replace a uniform finite-schedule constant without a
separate justification. The \(m=1\) experiments concern radial scores
and spectra, not an application of this \(m\ge2\) comparison.
\end{remark}

\begin{proposition}[Sphere-center spectrum: finite schedule and reference limit]
\label{prop:sphere_origin}
At \(x=0\), the center posterior is uniform and independent of scale,
so \(\nabla\log p^{\mathrm{pop}}(0)=0\). For any proper bounded scale schedule,

\begin{equation}
 \beta(0)=\mathbb E_w\sigma^{-2},\qquad
 C_0=\frac{R^2}{m+1}\mathbb E_w\sigma^{-4}\,{P_V}.
 \label{eq:sphere_finite_center_covariance}
\end{equation}

The Hessian eigenvalues are
\(\lambda_{\parallel}=R^2\mathbb E_w\sigma^{-4}/(m+1)-\beta(0)\)
on \(V\) and \(\lambda_{\perp}=-\beta(0)\) on \(V^\perp\).

For the unbounded reference and \(d>2\),
\(u=R^2/(2\sigma^2)\sim\operatorname{Gamma}((d-2)/2,1)\) under \(w\),
and exactly
\begin{equation}
 \beta(0)=\frac{d-2}{R^2},\quad
 C_0=\frac{(d-2)d}{(m+1)R^2}{P_V},\quad
 \frac{\operatorname{tr}(C_0)}{\beta(0)}=d,\quad
 r_{\mathrm{eff}}(C_0)=m+1.
 \label{eq:sphere_reference_center_covariance}
\end{equation}

For \(c\ge2\), the reference eigenvalues are
\begin{equation}
 \lambda_+=\frac{(d-2)(c-1)}{(m+1)R^2}
 \quad(m+1\text{ directions}),\qquad
 \lambda_-=-\frac{d-2}{R^2}
 \quad(c-1\text{ directions}).
 \label{eq:sphere_reference_spectrum}
\end{equation}
Thus the reference shell-rank constants are \(\alpha_d=1\) and
\(R_d=m+1\), with a strict gap when \(c\ge2\).
\end{proposition}
\begin{proof}
\textbf{Finite schedule.}
At the origin all centers are at distance $R$. The posterior center distribution is therefore uniform and independent of scale, with
$\mathbb E\mu=0$ and $\mathbb E\mu\mu^\top=R^2P_V/(m+1)$.
Thus the gradient vanishes and
$C_0=\mathbb E_w[\sigma^{-4}]\mathbb E[\mu\mu^\top]$.
Subtracting $\beta(0)I$ gives the two eigenspaces and eigenvalues.

\textbf{Unbounded reference.}
Here the raw scale weight is
$\sigma^{2-d}e^{-R^2/(2\sigma^2)}d\sigma/\sigma$.
Under $u=R^2/(2\sigma^2)$ it becomes Gamma with shape $(d-2)/2$.
Its first two moments give
\[
 \mathbb E_w\sigma^{-2}=(d-2)/R^2,\qquad
 \mathbb E_w\sigma^{-4}=(d-2)d/R^4.
\]
Substitution gives \eqref{eq:sphere_reference_center_covariance} and
\eqref{eq:sphere_reference_spectrum}. The positive eigenspace has dimension $m+1$, and the negative one dimension $c-1$; both signs are strict for $c\ge2$.
\end{proof}

\begin{remark}[Codimension one is schedule dependent]
\label{remark:knife_edge}
For \(c=1,d>2\), the reference eigenvalue on \(V\) is zero because
\(2\mathbb E[u^2]/d=\mathbb E[u]\). A finite schedule need not preserve
this sign: a single scale with \(\sigma^2>R^2/(m+1)\) makes
\(\lambda_{\parallel}<0\). For \((m,c)=(1,1)\), the formal Gamma
shape is zero, so even that unbounded posterior is not proper.
The zero reference curve in that numerical case is only a formal
benchmark. None of these observations rules out useful finite-scale
models in codimension one.
\end{remark}

\begin{lemma}[Reduction of the off-tube critical set]
\label{lemma:sphere_critical_reduction}
For a proper uniform-sphere mixture, every critical point lies in
\(V\cap\overline B_R(0)\). In this subspace, write
\(g(\varrho)=\langle\nabla\log p^{\mathrm{pop}}(\varrho e),e\rangle\), where \(e\in V\) is any
unit vector. This is independent of \(e\). If
\(g(\varrho)>0\) on \((0,R-r_0]\), the only off-tube critical point is
\(0\).

\end{lemma}
\begin{proof}
The barycenter lies in $V$, so
$(I-P_V)\nabla\log p^{\mathrm{pop}}=-\beta(I-P_V)x$.
Criticality forces $x\in V$, and the barycenter identity gives $\|x\|\le R$.
Rotational symmetry makes the gradient radial on $V$. For $x\in V$ of radius $\varrho\le R$, its distance to the sphere is $R-\varrho$.
Thus positivity of $g$ on $(0,R-r_0]$ excludes all nonzero off-tube critical points; the origin remains critical by symmetry.
\end{proof}

\begin{proposition}[Reference radial positivity and finite-range persistence]
\label{prop:sphere_radial_positivity}
For \(c\ge2\), the radial score of the unbounded reference kernel is
strictly positive on \((0,R)\). For each fixed \(0<r_0<R\), this
positivity on \((0,R-r_0]\), and hyperbolicity at the origin, persist
for sufficiently small lower cutoff, sufficiently large upper cutoff,
and then a sufficiently fine geometric discretization. No numerical
cutoff threshold is asserted here.
\end{proposition}
\begin{proof}
We first obtain a positive margin for the reference radial score, then
show that finite cutoffs and a fine grid preserve that margin. Positivity
at grid points alone would not give this conclusion near the origin.

\textbf{Integrating the reference scale.}
Because \(m\ge1\) and \(c\ge2\), we have \(d>2\). For \(a>0\),
the substitution \(u=a^2/(2\sigma^2)\) gives
\begin{align*}
 \int_0^\infty\sigma^{1-d}e^{-a^2/(2\sigma^2)}\,d\sigma
 &=\frac12\left(\frac{a^2}{2}\right)^{1-d/2}
                         \int_0^\infty u^{d/2-2}e^{-u}\,du\\
 &=2^{d/2-2}\Gamma(d/2-1)a^{2-d}.
\end{align*}
After including the Gaussian factor \((2\pi)^{-d/2}\), the multiplier
is a positive constant depending only on \(d\). Since constant
multipliers do not change a score, we may use the reference kernel
\[
 \mathscr P_\infty(y)
 :=\int_{S^m(R)}\|y-\mu\|^{2-d}\,Q_{\mathcal M}(d\mu).
\]
It is positive and smooth away from the sphere, and radial when
restricted to \(V\).

\textbf{Positivity from the subspace Laplacian.}
The Laplacian here is taken in the \((m+1)\)-dimensional subspace
\(V\), not in \(\mathbb R^d\). In an \(n\)-dimensional Euclidean
space, differentiation gives
\[
 \nabla\|y\|^p=p\|y\|^{p-2}y,\qquad
 \Delta\|y\|^p=p(p+n-2)\|y\|^{p-2}.
\]
Take \(p=2-d\), \(n=m+1\), and replace \(y\) by \(y-\mu\).
The coefficient becomes
\[
 (2-d)(2-d+m+1-2)
 =(2-d)(m+1-d)=(d-2)(c-1)>0.
\]
For \(y\in V\) with \(\|y\|<R\), differentiation under the compact
center integral is valid locally because \(\|y-\mu\|\) is bounded
away from zero. Thus
\begin{equation}
 \Delta_V\mathscr P_\infty(y)
 =(d-2)(c-1)\int\|y-\mu\|^{-d}\,Q_{\mathcal M}(d\mu)>0.
 \label{eq:detailed_sphere_subspace_laplacian}
\end{equation}
Write \(P_\infty(\varrho)=\mathscr P_\infty(\varrho e)\) for a unit vector \(e\in V\).
Smooth radial symmetry gives \(P_\infty'(0)=0\), and the radial Laplacian
formula reads

\[
 P_\infty''(\varrho)+\frac m{\varrho}P_\infty'(\varrho)=\Delta_V\mathscr P_\infty(\varrho e),
 \quad\text{or}\quad
 (\varrho^m P_\infty'(\varrho))'=\varrho^m\Delta_V\mathscr P_\infty(\varrho e).
\]

Integrating from zero to \(\varrho>0\), using
\(\lim_{u\downarrow0}u^m P_\infty'(u)=0\), gives

\[
 P_\infty'(\varrho)=\varrho^{-m}\int_0^\varrho u^m\Delta_V\mathscr P_\infty(ue)\,du>0.
\]

Since \(P_\infty(\varrho)>0\), the reference score
\(g_\infty(\varrho)=P_\infty'(\varrho)/P_\infty(\varrho)\) is positive for every \(0<\varrho<R\).

\textbf{A uniform margin including the origin.}
The reference center spectrum gives
\[
 g_\infty(0)=0,\qquad
 g_\infty'(0)=\lambda_+
 =\frac{(d-2)(c-1)}{(m+1)R^2}>0.
\]
Thus \(g_\infty(\varrho)/\varrho\) extends continuously to \(\varrho=0\), with
value \(\lambda_+\). For fixed \(r_0\in(0,R)\), define

\[
 \gamma:=\min_{0\le \varrho\le R-r_0}\frac{g_\infty(\varrho)}\varrho>0,
\]

using this continuous extension at zero. Compactness and strict
positivity justify the positive minimum.

\textbf{Removing tails and discretizing in \(C^2\).}
On the full ambient closed ball \(\|x\|\le R-r_0\), every center
is at distance at least \(r_0\), and the distances are also uniformly
bounded above. After omission of the mixture normalization, the scale
integrand is \(\sigma^2\mathcal N(x;\mu,\sigma^2I)d\sigma/\sigma\).
The Gaussian derivative identities
\[
 \nabla_x\mathcal N=-\frac{x-\mu}{\sigma^2}\mathcal N,
 \qquad
 \nabla_x^2\mathcal N
 =\left(\frac{(x-\mu)(x-\mu)^\top}{\sigma^4}
                      -\frac I{\sigma^2}\right)\mathcal N
\]
give the following bounds for the scale-integrand coefficients
relative to \(d\sigma\), including their first two spatial derivatives
(the unit cutoff is only for expressing integrability):
\[
 C\sigma^{-d-3}e^{-r_0^2/(2\sigma^2)}\quad(0<\sigma\le1),
 \qquad
 C\sigma^{1-d}\quad(\sigma\ge1),
\]
where \(C\) is finite and independent of \(x,\mu,\sigma\) on the
specified compact ball. The first bound is integrable at zero because
of Gaussian suppression; the second is integrable at infinity because
\(d>2\). Their tail integrals tend to zero. Hence the unnormalized
bounded-scale kernels, after multiplication by the same harmless
positive constant obtained by scale integration, converge to \(\mathscr P_\infty\)
uniformly with their first two derivatives as the lower cutoff tends
to zero and the upper cutoff tends to infinity.

The reference kernel has a positive minimum on this compact ball.
The identities
\[
 \nabla\log P=\frac{\nabla P}{P},\qquad
 \nabla^2\log P=\frac{\nabla^2P}{P}
                   -\frac{\nabla P\nabla P^\top}{P^2}
\]
therefore imply convergence of the log-kernel derivatives as well.
For fixed positive finite cutoffs, set \(t=\log\sigma\). The scale
integrand and its first two spatial derivatives are uniformly
continuous on the compact \((x,t)\)-domain. Their geometric-grid
sums multiplied by \(h\) are Riemann sums, so they also converge
uniformly with these derivatives as \(h\to0\). Neither multiplication
by \(h\) nor normalization of the density changes its score or log
Hessian.

\textbf{Transferring the sign to the full interval.}
Let \(g_{\mathrm{fin}}\) be a score obtained after the chosen
cutoffs and discretization. Symmetry ensures
\(g_{\mathrm{fin}}(0)=g_\infty(0)=0\). For \(\varrho>0\), the fundamental
theorem of calculus gives

\begin{align*}
 \left|\frac{g_{\mathrm{fin}}(\varrho)-g_\infty(\varrho)}\varrho\right|
 &=\left|\frac1\varrho\int_0^\varrho
                   (g_{\mathrm{fin}}'(u)-g_\infty'(u))\,du\right|\\
 &\le\sup_{0\le u\le R-r_0}
                   |g_{\mathrm{fin}}'(u)-g_\infty'(u)|.
\end{align*}

Choose the log-density Hessian approximation error below \(\gamma/2\). Then

\[
 \frac{g_{\mathrm{fin}}(\varrho)}\varrho
 \ge\frac{g_\infty(\varrho)}\varrho-\frac\gamma2
 \ge\frac\gamma2>0
 \quad(0<\varrho\le R-r_0).
\]

This also handles arbitrarily small positive \(\varrho\), where ordinary
uniform convergence of scores alone would not suffice.
Finally, the reference center eigenvalues \(\lambda_+>0\) and
\(\lambda_-<0\) are separated from zero. Taking the same Hessian
error below \(\tfrac12\min\{\lambda_+,-\lambda_-\}\) preserves their
signs, and hence hyperbolicity at the origin. All cutoffs and grid
sizes in this argument are qualitative existence choices, not an
explicit certificate for the schedules used in E2.

\end{proof}
This argument supplies an analytic reference result and a qualitative
finite-range existence result. It does not certify the particular
coarse schedules in E2; those remain numerical tests.

\begin{corollary}[Saddle avoidance in a proper sphere mixture]
\label{cor:sphere_saddle_avoidance}
For a bounded-scale uniform-sphere mixture with \(c\ge2\), assume \(g>0\) on
\((0,R-r_0]\) and \(\lambda_{\parallel}>0\) at the origin. Then the
only possible off-tube endpoint is the hyperbolic center, and an
absolutely continuous initialization avoids it almost surely.
\end{corollary}
\begin{proof}
The preceding reduction leaves only the origin off the tube. Its eigenvalues are $\lambda_\parallel>0$ on $V$ and $-\beta(0)<0$ on $V^\perp$, so it is hyperbolic and its stable manifold has dimension $c-1<d$. The proof of Lemma~\ref{lemma:sufficient_saddle_avoidance} applies to this proper population mixture, whose analytic and flow properties were established above.
\end{proof}
\paragraph{Compatibility in one proper model.}
For $m\ge2,c\ge3$, let
$\overline C_c=4c[1+30(1/27+e^{-c/2})]/(c-2)$.
It bounds $C_{\mathrm{flat}}$ whenever $h\sqrt c\le1/27$.
Choose $r_0$ so that $\overline C_cmr_0/[2c(R-r_0)]<1$, then choose $\sigma_{\min}$ small enough that
$\overline C_cm\sigma_{\min}^2/[2(R-r_0)]<\min\{r_0,\sigma_{\min}\sqrt c\}$.
The finite-range persistence proposition permits a sufficiently large upper cutoff and sufficiently fine grid giving radial positivity and center hyperbolicity, while the preceding uniform moment bound remains valid.
At the only off-tube critical point, $C_0$ is proportional to a rank-$(m+1)$ projector and $\lambda_\parallel>0$ implies $\operatorname{tr}C_0/\beta(0)>m+1$.
Thus $R_d=m+1$ and $\alpha_d=\min\{1,\operatorname{tr}C_0/[d\beta(0)]\}$ satisfy the actual shell-rank gap. This verifies compatibility of the geometric assumptions in a proper model; it does not certify the particular coarse numerical schedules.

\begin{proposition}[Exact regional probabilities and zero projected drift on a uniform sphere]
\label{prop:sphere_routing}
Consider a proper bounded-scale uniform-sphere mixture and an
initialization distribution invariant under \(O(m+1)\times O(c-1)\), with
\(\mathbb P[{P_V} X_0=0]=0\). Suppose the transition and stopping tubes
are reached almost surely. Then

\begin{equation}
 Z_{\mathrm{stop}}=Z_{\mathrm{tr}}
 =R\frac{{P_V} X_0}{\|{P_V} X_0\|}\quad\text{almost surely},\qquad
 Z_{\mathrm{tr}},Z_{\mathrm{stop}}\sim\operatorname{Unif}(S^m(R)).
 \label{eq:sphere_exact_projection_retention}
\end{equation}

The actual tangential retention radius is zero.
\end{proposition}
\begin{proof}
Rotations in $V$ fixing $P_Vx$ also fix its posterior barycenter, which lies in $V$. Thus $\dot y=b(t)y$ for $y=P_Vx$ and a scalar $b(t)$. The subspace $y=0$ is invariant; uniqueness prevents a path starting outside it from reaching it at finite time. Therefore $y(t)/\|y(t)\|$ is constant.
The unique nearest point on the sphere is $RP_Vx/\|P_Vx\|$, so the two projected outputs equal the initial direction scaled by $R$. Rotation-invariant initialization makes that direction uniform. Finite entrance is assumed in the proposition, completing the probability statement.
\end{proof}

\begin{corollary}[Concrete sphere coverage]
\label{cor:sphere_concrete_diversity}
Under Proposition~\ref{prop:sphere_routing}, for every \(z\in S^m(R)\)
and \(\rho>0\),
\begin{equation}
 \mathbb P[Z_{\mathrm{stop}}\in B_{\mathcal M}(z,\rho)]
 =\frac{\displaystyle\int_0^{\min\{\rho/R,\pi\}}
                 \sin^{m-1}\theta\,d\theta}
        {\displaystyle\int_0^\pi\sin^{m-1}\theta\,d\theta}.
 \label{eq:sphere_geodesic_ball_coverage}
\end{equation}
\end{corollary}
\begin{proof}
The preceding proposition makes $Z_{\mathrm{stop}}$ uniform. In colatitude $\theta$ around the ball center, the surface element is proportional to $\sin^{m-1}\theta\,d\theta$ and intrinsic distance is $R\theta$. The ratio of the cap integral to the full integral is the stated probability. At $\rho=\pi R$ the missing antipode has zero mass.
\end{proof}
When the general posterior and radius assumptions also hold, their
reach--noise retention bound is valid but conservative on the uniform
sphere: symmetry gives zero projected drift directly. The zero-drift
result itself needs no localization constant once finite entrance is
known. Nonuniform densities or finite center clouds need not preserve
this cancellation, which is why E1 tests genuine projected motion.

\begin{remark}[Population-to-empirical scope]
\label{remark:population_to_empirical}
The exact sphere statements concern a uniform population, not a fixed
random center cloud. A uniform empirical transfer requires separate
control of posterior moments and landscape perturbations. The
\(h_M^2\) extension is a conditional way to express a moment residual;
E1's nearest-center quantile is a sampling-resolution diagnostic, not
a proof of that residual bound.
\end{remark}

\subsection{Finite-schedule reference calculations for E4}
\label{subsec:e4_reference_calculations}
The E4 curves use the same precision-normalized weights as the preceding analysis. We record their finite-schedule forms to distinguish these references from both the unbounded-scale limit and fitted empirical curves.

\paragraph{The flat finite-schedule moment.}
For any fixed positive finite schedule with equal scale sampling, integrating the tangent Gaussian in the flat model gives
\begin{equation}
  \pi_i^{\mathrm{flat}}(r)
  =\frac{\sigma_i^{-c}\exp[-r^2/(2\sigma_i^2)]}
         {\sum_{a=1}^N\sigma_a^{-c}\exp[-r^2/(2\sigma_a^2)]},
  \qquad
  \mathbb E_{\widetilde w,\mathrm{flat},r}\|\mu-z_x\|^2
  =m\sum_{i=1}^N\pi_i^{\mathrm{flat}}(r)\sigma_i^2.
  \label{eq:e4_flat_finite_posterior}
\end{equation}
Thus the radius-dependent reference plotted in E4 is
\begin{equation}
  \mathcal C_{\mathrm{flat},N}(r)
  :=\frac{\sum_{i=1}^N\pi_i^{\mathrm{flat}}(r)\sigma_i^2}
          {\max\{\sigma_{\min}^2,r^2/c\}}.
  \label{eq:e4_flat_finite_ratio}
\end{equation}
This is an exact ratio for the translation-invariant flat posterior, not a normalizable flat data distribution. It is also a function of $r$, distinct from a uniform constant $C_{\mathrm{flat}}$ used in Proposition~\ref{prop:sphere_localization}. The sphere comparison supplies the factor $R/\varrho$ for a population moment; it does not make \eqref{eq:e4_flat_finite_ratio} an empirical upper bound or predict an exact finite-cloud median.

\paragraph{Small-noise radial bias on the uniform two-sphere.}
For the proper population $S^2(R)$ model used in E4, fix a finite relative schedule $1=q_1\le\cdots\le q_N<\infty$ and write $\sigma_i=sq_i$, with $s=\sigma_{\min}$. At $x=\varrho e\in V$, let $a_i=\varrho R/\sigma_i^2$. The variable $t=\langle e,\mu/R\rangle$ is uniform on $[-1,1]$ before conditioning. Direct integration gives
\begin{equation}
 \begin{split}
  I_i(\varrho)&:=\int_{S^2(R)}\mathcal N(\varrho e;\mu,\sigma_i^2I)
                                    Q_{\mathcal M}(d\mu)\\
  &=(2\pi\sigma_i^2)^{-d/2}
    e^{-(\varrho-R)^2/(2\sigma_i^2)}
    \frac{\sigma_i^2}{2\varrho R}
    \left(1-e^{-2a_i}\right).
 \end{split}
 \label{eq:e4_sphere_fixed_scale_integral}
\end{equation}
The mean radial coordinate conditional on scale is $R(\coth a_i-a_i^{-1})$. Consequently the normalized radial field is
\begin{equation}
  F_s(\varrho)
  =R\sum_{i=1}^N\pi_i^{\mathrm{sph}}(\varrho)
            (\coth a_i-a_i^{-1})-\varrho,
  \qquad
  \pi_i^{\mathrm{sph}}(\varrho)
  =\frac{I_i(\varrho)}{\sum_{a=1}^NI_a(\varrho)}.
  \label{eq:e4_sphere_exact_normalized_field}
\end{equation}
These are precision-normalized scale probabilities: the extra $\sigma_i^2$ in the unified density has canceled against the score precision. Since $\beta>0$, zeros of $F_s$ are zeros of the original radial score as well.

To examine the near-sphere branch, put $\varrho=R-bs^2/R$ with bounded $b$. For the fixed finite relative schedule, $s/R\to0$ implies $a_i\to\infty$ at every scale. Equations~\eqref{eq:e4_sphere_fixed_scale_integral}--\eqref{eq:e4_sphere_exact_normalized_field}, together with $\coth a_i=1+O(e^{-2a_i})$, give
\begin{equation}
  \pi_i^{\mathrm{sph}}(R-bs^2/R)
  =\frac{q_i^{-c}}{\sum_{a=1}^Nq_a^{-c}}+O((s/R)^2),
  \qquad c=d-2,
  \label{eq:e4_sphere_small_noise_weights}
\end{equation}
and hence, uniformly for $b$ in a fixed bounded neighborhood of the limiting root,
\begin{equation}
  \frac R{s^2}F_s(R-bs^2/R)
  =b-\frac{\sum_iq_i^{2-c}}{\sum_iq_i^{-c}}
       +O((s/R)^2).
  \label{eq:e4_scaled_radial_root}
\end{equation}
The expansion also holds after differentiation in $b$: it follows directly from the finite sums above, whose omitted exponential terms and their derivatives vanish faster than any power. The leading root is positive and simple, so there is a local stationary branch with
\begin{equation}
  R-\varrho_\star
  =\frac{s^2}{R}\left[
       \frac{\sum_iq_i^{2-c}}{\sum_iq_i^{-c}}+O((s/R)^2)\right].
  \label{eq:e4_sphere_finite_schedule_bias}
\end{equation}
For $m=2$, the leading constant can therefore be written as
\begin{equation}
  \frac{R|\varrho_\star-R|}{\sigma_{\min}^2}
  \longrightarrow C_{\mathrm{fs}}
  =\frac m2\frac{\sum_{i=1}^N q_i^{2-c}}{\sum_{i=1}^N q_i^{-c}}
  =1.199652317\ldots,
  \label{eq:main_e4_finite_schedule_bias}
\end{equation}
where the numerical value substitutes $c=10$, $N=64$, and $q_i=16^{(i-1)/63}$. This calculation concerns a proper population sphere with a fixed finite relative schedule, all of whose scales shrink in the limit. It is not an assertion that the general finite-cloud localization bound is attained, nor does it identify a sampled value of $C_{\mathrm{loc}}$ with $C_{\mathrm{fs}}$.

\subsection{Empirical localization from local sample mass}
\label{subsec:local_mass_localization}
The constant $C_{\mathrm{loc}}$ measures the spread of the centers that influence the score, relative to the effective noise scale at the query. On a well-sampled affine patch, a ball of radius $t$ contains a number of centers proportional to $t^m$ once $t$ exceeds the sampling resolution. If the noise floor covers that resolution, the posterior has enough local mass to normalize and distant centers receive exponentially small weights. The following result quantifies this argument with curvature and a finite schedule.

\par\begingroup
\begin{proposition}[Local mass sufficient for empirical posterior localization]
\label{prop:empirical_mass_localization}
Let $\mathcal M$ have dimension $m$, codimension $c\ge3$, and reach at least $\tau$, with the empirical centers on $\mathcal M$. Fix a reference subset $K\subseteq\mathcal M$ and a tube radius $r_0$ with $r_0/\tau\le1/(4m)$. Set $\alpha=1-r_0/\tau$ and $\beta_+=1+r_0/\tau$. For $z\in K$, let $F_z(t)=\widehat Q_M(\overline B(z,t))$ for closed ambient Euclidean balls. Suppose for some $a_z>0$, mass ratio $K_{\mathrm{mass}}$, and radii $h_{\mathrm{reg}}\le R_{\mathrm{loc}}$,
\[
 a_z t^m\le F_z(t)\quad(h_{\mathrm{reg}}\le t\le R_{\mathrm{loc}}),
 \qquad F_z(t)\le K_{\mathrm{mass}}a_z t^m\quad(t\ge h_{\mathrm{reg}}).
\]
Here $h_{\mathrm{reg}}$ is the scale above which the ball counts satisfy these comparisons. It is at least the fill distance on $K$. Assume the finite geometric schedule and $r_0$ satisfy Lemma~\ref{lemma:discrete_schedule}. Write
\[
 C_{\mathrm{scale}}=\bigl[1+30(h\sqrt c+e^{-c/2})\bigr]\frac{4c}{c-2},\qquad
 q_0^2=\max\{\sigma_{\min}^2,r_0^2/c\}.
\]
It suffices that
\[
 \sigma_{\min}\ge\sqrt{2\beta_+}\,h_{\mathrm{reg}},
 \qquad
 R_{\mathrm{loc}}\ge\sqrt{\frac{8C_{\mathrm{scale}}(m+2)}{\beta_+}}\,q_0 .
\]
Then \eqref{eq:main_posterior_localization} holds uniformly for $z_x\in K$, $0\le r(x)\le r_0$, with
\[
 C_{\mathrm{loc}}
 =4K_{\mathrm{mass}}\frac{m+2}{m}
   \left(\frac{\beta_+}{\alpha}\right)^{m/2}\frac{C_{\mathrm{scale}}}{\alpha}.
\]
This constant stays $O(1)$ if $K_{\mathrm{mass}}$ stays bounded. The lower mass inequality controls normalization, and the upper inequality controls the second moment. The radius $R_{\mathrm{loc}}$ needs to cover the effective smoothing neighborhood, which can be much smaller than the largest noise scale.
\end{proposition}
\par\endgroup
\begin{proof}
Fix a projection $z=z_x$ and normal distance $r$. By Lemma~\ref{lemma:reach_gaussian_sandwich}, the denominator and moment numerator at each scale can be compared with radial kernels of coefficients $\beta_+$ and $\alpha$, respectively, after factoring out the normal likelihood. Define
$\bar\sigma=\min\{\sigma_{\max},\sqrt{2C_{\mathrm{scale}}}\,q_0\}$.
For $\sigma\in[\sigma_{\min},\bar\sigma]$, integration by parts and the lower mass bound give
\[
 \int e^{-\beta_+t^2/(2\sigma^2)}\,dF_z(t)
 \ge a_z\int_{h_{\mathrm{reg}}}^{R_{\mathrm{loc}}}
          \frac{\beta_+}{\sigma^2}t^{m+1}e^{-\beta_+t^2/(2\sigma^2)}\,dt .
\]
The retained integral is at least half of the full integral. After changing variables to $\beta_+t^2/(2\sigma^2)$, the endpoints bound an interval containing $[1/4,2(m+2)]$. For a gamma variable of shape $(m+2)/2$ and unit scale the two tail losses are at most $1-e^{-1/4}$ and $1/4$, by comparison with an exponential and Markov's inequality. The lower bound is therefore $(a_z/2)\Gamma(1+m/2)(2\sigma^2/\beta_+)^{m/2}$.

For the numerator, integrate $t^2 e^{-\alpha t^2/(2\sigma^2)}$ against $dF_z(t)$ by parts. The nonpositive part of the resulting integral occurs for $t<\sqrt{2/\alpha}\,\sigma$ and can be dropped. Above that threshold the upper mass bound applies. Dropping the term $-2t$ from the derivative gives the upper bound
$K_{\mathrm{mass}}a_z(m+2)(\sigma^2/\alpha)\Gamma(1+m/2)(2\sigma^2/\alpha)^{m/2}$.

Including the ambient Gaussian normalization leaves scale weights proportional to $\sigma_i^{-c}e^{-r^2/(2\sigma_i^2)}$. Lemma~\ref{lemma:discrete_schedule} bounds their mean of $\sigma_i^2$ by $C_{\mathrm{scale}}\max\{\sigma_{\min}^2,r^2/c\}$, with the $r=0$ case by continuity. At least half of their mass is on $\sigma_i\le\bar\sigma$ by Markov's inequality. Dividing the moment bound by the denominator bound gives the stated constant. Finally, $(\beta_+/\alpha)^{m/2}/\alpha\le(4/3)e^{1/3}$ and the schedule assumption keeps $C_{\mathrm{scale}}$ bounded for $c\ge3$.
\end{proof}

\clearpage
\section{Numerical Verification: Full Settings and Results}
\label{subsec:numerical_verification}

The solvable models give reference calculations for posterior localization and
critical-point geometry. E1 evaluates regional probability and tangential
retention on spheres, including explicit twelve-region coverage budgets.
E2 studies off-tube critical structure and compares ordinary denoising score
matching (DSM) with the exact score on the same finite sphere mixtures.
E3 measures separation from the training points and regional coverage on
finite affine grids. E4 examines spherical bias and posterior localization.
Appendix~\ref{sec:learned_circle} verifies first-entry coverage for learned
circle fields using uniform numerical enclosures.

For the empirical mixtures, the score
\eqref{eq:unified_score_manifold} is evaluated by the exact finite
log-sum-exp over all scale--center pairs, up to floating-point error.
The remaining numerical approximations are the Euler time
discretization and event-location error in E1, and quadrature and
nonlinear root-finding error in E2. E3 records the numerical accuracy of
the endpoint computations, while E4 separates population quadrature from
sampled finite-cloud moment calculations. In the empirical E1 trajectories
and the E3 bandwidth sweep, we integrate by
\[
    x\leftarrow x+\eta\bigl(\bar\mu(x)-x\bigr),
\]
an Euler discretization of the time-reparameterized flow
\(\dot x=\bar\mu(x)-x\). Its continuous-time trajectories and hitting
locations coincide with those of
\eqref{eq:unified_gradient_ascent}; sensitivity to the step size
\(\eta\) is reported explicitly below.

\begin{lemma}[Ascent of the normalized Euler update]
\label{lemma:normalized_euler_ascent}
For the exact finite-mixture field, let
\begin{equation}
x^+=x+\eta\bigl(\bar\mu(x)-x\bigr).
\label{eq:app_normalized_euler_update}
\end{equation}
Then, in exact arithmetic,
\begin{equation}
\log p_{MN}(x^+)-\log p_{MN}(x)
\ge
\beta(x)\eta\left(1-\frac{\eta}{2}\right)
\|\bar\mu(x)-x\|^2.
\label{eq:app_normalized_euler_ascent}
\end{equation}
In particular, $0<\eta<2$ gives strict ascent unless $x$ is stationary.
If the iteration uses steps
$0<\underline\eta\le\eta_n\le\overline\eta<2$, then, from every starting
point, it has finite total length and converges to a stationary point of
$\log p_{MN}$.
\end{lemma}

\begin{proof}
For any displacement $\Delta$, the Gaussian density ratio and Jensen's
inequality give
\begin{align*}
\log\frac{p_{MN}(x+\Delta)}{p_{MN}(x)}
&=
\log\sum_{j,i}w_{ij}(x)
\exp\!\left(
\frac{\langle\mu_j-x,\Delta\rangle}{\sigma_i^2}
-\frac{\|\Delta\|^2}{2\sigma_i^2}
\right)\\
&\ge
\langle\nabla\log p_{MN}(x),\Delta\rangle
-\frac{\beta(x)}{2}\|\Delta\|^2.
\end{align*}
Substituting $\Delta=\eta(\bar\mu(x)-x)$ and using
$\nabla\log p_{MN}(x)=\beta(x)(\bar\mu(x)-x)$ proves
\eqref{eq:app_normalized_euler_ascent}.

For the convergence statement, let $\Delta_n=x_{n+1}-x_n$. The step
bounds and $\sigma_{\max}^{-2}\le\beta(x)\le\sigma_{\min}^{-2}$ imply
\[
\begin{aligned}
\log p_{MN}(x_{n+1})-\log p_{MN}(x_n)
&\ge \sigma_{\max}^{-2}
\left(\overline\eta^{-1}-\frac12\right)\|\Delta_n\|^2,\\
\|\nabla\log p_{MN}(x_n)\|
&\le\frac{\sigma_{\min}^{-2}}{\underline\eta}\|\Delta_n\|.
\end{aligned}
\]
All iterates remain in the compact initial superlevel set. The two
estimates imply vanishing steps and critical accumulation points. Applying
the analytic-gradient inequality on the compact accumulation set to the
same estimates gives $\sum_n\|\Delta_n\|<\infty$, and therefore
convergence to a stationary point.
\end{proof}

The step sizes used in E1 and E3 lie in this ascent regime. This result
justifies the normalized update as an ascent method for the exact objective;
it does not identify its finite-step hitting locations or output
probabilities with those of the continuous flow. The step-size and
termination diagnostics below address these remaining numerical effects.

\subsection{E1: Regional probability beyond symmetry}
\label{sec:e1}

\subsubsection{Continuous coverage on a uniform sphere}
\label{subsec:revision_e1}
\paragraph{Finite-grid error bound.}
Write $s=\log\sigma$, $L=\log(\sigma_{\max}/\sigma_{\min})$, $h=L/(N-1)$, and $F(s)=\partial_s p_{e^s}^Q$. For a common exact reference law, the residual is $h\sum_i F(s_i)-\int F(s)\,ds$. The inclusive sum differs from the trapezoidal rule by $h(F(s_1)+F(s_N))/2$. The $L^1$ norms of $F$ and $F''$ are bounded by $\sqrt{2d}$ and $\sqrt{48d^3+576d^2+992d}$, respectively. To see this, differentiate the Gaussian kernel with respect to $s$. Its first and third derivatives are the kernel multiplied by $Z-d$ and $(Z-d)^3-6Z(Z-d)+4Z$, where $Z=\|x-\mu\|^2/\sigma^2$. Under the kernel, $Z\sim\chi^2_d$. Cauchy--Schwarz, its moments, and averaging over $Q$ give the stated bounds. Applying the composite trapezoid error bound in $L^1$ then gives
{
\begin{equation}
 \epsilon_{\mathrm{Pois}}\le
 \frac{h\sqrt{2d}}{2}
 +\frac{L h^2\sqrt{48d^3+576d^2+992d}}{24}.
 \label{eq:revision_grid_error}
\end{equation}
}
The first term retains the endpoint contribution of the actual training grid.

\paragraph{Uniform-sphere settings.}
We use $Q=\operatorname{Unif}(S^2(1))$ in $\mathbb R^{12}$ and twelve equal-mass colatitude bands, with boundaries at third coordinates $-1+2j/12$ for $j=0,\ldots,12$. Both settings take $r_0=0.05$. The 1,024-level grid uses cutoffs $[10^{-3},10]$, transition and stop radii $(0.006,0.0015)$, and $\zeta=0.1$. The 16,384-level grid uses $[10^{-5},50]$, radii $(8\times10^{-5},5\times10^{-6})$, and $\zeta=0.001$. Initialization is a centered Gaussian of variance $\sigma_{\max}^2$ conditioned on the central radial shell omitting mass $10^{-3}$ or $10^{-6}$, respectively.

To bound initialization, let $\phi_\sigma$ be the centered ambient Gaussian density. For independent uniform sphere points $\mu,\nu$, integrating the square of the likelihood ratio $p_\sigma^Q/\phi_\sigma$ yields
\[
 \chi^2(p_\sigma^Q\Vert\phi_\sigma)
 =\mathbb E e^{\langle\mu,\nu\rangle/\sigma^2}-1
 =\frac{\sinh(\sigma^{-2})}{\sigma^{-2}}-1,
\]
since $\langle\mu,\nu\rangle$ is uniform on $[-1,1]$. Thus $\epsilon_{\mathrm{init}}\le\delta_{\mathrm{shell}}+\tfrac12\sqrt{\chi^2(p_{\sigma_{\max}}^Q\Vert\phi_{\sigma_{\max}})}$.

For both grids we check the schedule conditions in Lemma~\ref{lemma:discrete_schedule}, localization, contraction, and off-tube instability using Appendix~\ref{subsec:sphere_model}, and the Gaussian-placement conditions in \eqref{eq:source_sink_band_conditions}. With $p_*=\min_jQ(U_j^{-\delta_{\mathrm{cov}}})$, the probability that both transition and first entry lie in any specified $U_j$ is at least
\[
 \frac1{12}-\underbrace{(1/12-p_*)}_{\text{boundary erosion}}
          -\underbrace{\zeta p_*}_{\text{placement loss}}
          -\epsilon_{\mathrm{Pois}}-\epsilon_{\mathrm{init}}.
\]
The computed displacement radii are $0.010570$ and $0.00016643$. We erode the colatitude bands by these radii and integrate the remaining spherical area.

\begin{table}[!htb]
\centering
\small
\caption{Continuous first-entry coverage on a uniform sphere at twelve-region resolution. Losses are rounded up and lower bounds down. All probability columns are percentages of total mass. The reference mass is $8.33\%$ per region.}
\label{tab:revision_e1}
\begin{tabular}{@{}rrrrrr@{}}
\toprule
\shortstack{Noise\\levels} & \shortstack{Poisson\\error (\%)} &
\shortstack{Initialization\\error (\%)} & \shortstack{Boundary\\erosion (\%)} &
\shortstack{Placement\\loss (\%)} & \shortstack{Per-region\\lower (\%)} \\
\midrule
1,024 & 3.517 & 0.305 & 1.051 & 0.729 & 2.73 \\
16,384 & 0.255 & 0.009 & 0.017 & 0.009 & 8.04 \\
\bottomrule
\end{tabular}
\end{table}

\subsubsection{Regional probability and retention on nonuniform empirical spheres}
\paragraph{Setup.}
We use the sphere
\(S^2(1)\subset\mathbb R^{12}\), so \(m=2\) and \(c=10\), with
\(M=10^4\) centers drawn from either
(i) a von Mises--Fisher density with concentration
\(\kappa\in\{1,4,16\}\), or
(ii) a two-cluster mixture of antipodal \(\mathrm{vMF}(8)\)
components with masses \((p,1-p)\),
\(p\in\{0.5,0.9\}\). The geometric schedule contains \(24\) scales
on \([0.02,2]\). We take
\[
    r_0=0.3,
    \qquad
    r_{\mathrm{tr}}=0.2,
    \qquad
    r_{\mathrm{stop}}=0.03.
\]

\paragraph{Finite-grid residual.}
All centers lie in $\mathbb R^3\times\{0\}^9$. On the cylinder
$\|x_\perp\|\le0.058$ in the nine appended coordinates, the projected
source--sink residual is independent of the spherical point cloud.
Integrating \eqref{eq:finite_source_sink_identity} over this cylinder for
the 24-level schedule on $[0.02,2]$ gives a signed mass of approximately
$-0.168174$. The cylinder mass for a Gaussian at scale $\sigma$ is
$F_{\chi^2_9}(0.058^2/\sigma^2)$, and the scale derivative in the finite
sum is its derivative with respect to $\log\sigma$. The total residual
integrates to zero, so $\epsilon_{\mathrm{Pois}}$ is at least the magnitude
of the cylinder mass, which exceeds $1/12$. The grid error alone precludes
a positive bound through this twelve-region coverage budget.

\paragraph{Retention benchmark and trajectories.}
Using the population benchmark constant specified below, the
reference values associated with \eqref{eq:diversity_radius_ordering} are
\[
    r_{\min}
    =
    \sigma_{\min}\sqrt c
    \approx0.063,
    \qquad
    \varepsilon_0\approx5\times10^{-4}.
\]
For the unbounded local-limit benchmark
\(C_{\mathrm{loc}}=c/(c-2)=1.25\) from
Proposition~\ref{prop:sphere_localization}, the constants are
\[
 C_{\mathrm{loc}}/2=0.625,\qquad
 L_\Pi=1.25,\qquad \kappa_0=0.0375,\qquad \varepsilon_0=0.0005.
\]

Consequently \eqref{eq:explicit_early_stopping_retention_radius} evaluates to
\begin{align*}
 \delta_{\mathrm{stop}}^{\mathrm{pop}}
 &=1.25\left[
   \frac{0.5}{0.9625}(0.2-0.02\sqrt{10})
   +0.02\sqrt{2.5}\log\frac{0.02\sqrt{10}-0.0005}{0.03-0.0005}
          \right]\\
 &=0.1186336\ldots\approx0.119.
\end{align*}
Replacing the benchmark constant by
\(1.25\,R/(R-r_{\mathrm{tr}})=1.5625\) and recomputing the same
formula gives the curvature-inflated reference value \(0.134\).
Both benchmarks use population constants. We compare them with the measured
retention of trajectories for the realized finite cloud and coarse schedule.

We run \(10^5\) trajectories from the truncated Gaussian
initialization \eqref{eq:truncated_gaussian_initialization} with
\(\delta_{\mathrm{init}}=10^{-3}\) and default step size
\(\eta=0.15\). We record \(Z_{\mathrm{tr}}\) at the first crossing of
\(r(x)\le r_{\mathrm{tr}}\) and \(Z_{\mathrm{stop}}\) at the first
subsequent crossing of \(r(x)\le r_{\mathrm{stop}}\). Since every
center lies in \(\mathbb R^3\times\{0\}\), the Euler update closes
over the pair \(({P_V} x,\|(I-{P_V})x\|)\), which we exploit exactly.

For the population vMF model, rotational symmetry about the vMF axis
makes the regional probability ratio one-dimensional in the colatitude. A fixed
finite center cloud has only approximate rotational symmetry, so the
same statistic should be interpreted as an axial, finite-resolution
probability diagnostic. Let \(\mathcal P\) denote the prescribed
partition of the sphere. For each cell \(B\in\mathcal P\), let
\(\underline p_B\) be the lower endpoint of the two-sided \(95\%\)
Wilson interval for the transition probability estimated from the
sampled trajectories. We report
\begin{equation}
    \widehat\alpha_{\mathcal P}
    :=
    \min_{B\in\mathcal P}
    \frac{\underline p_B}{Q_{\mathcal M}(B)}.
    \label{eq:empirical_partition_routing_constant}
\end{equation}

For the vMF targets, \(\mathcal P\) consists of twelve equal-
\(Q_{\mathcal M}\)-mass colatitude bins, constructed using the
closed-form inverse CDF of the \(t=\cos\theta\) marginal. For the
cluster mixtures, \(\mathcal P\) consists of the two hemispheres
centered at the component poles. If \(\widehat{\mathbb P}\) denotes
the empirical trajectory distribution, then
\(\underline p_B\le
\widehat{\mathbb P}[Z_{\mathrm{tr}}\in B]\) for every cell and therefore
\begin{equation}
    \widehat{\mathbb P}[Z_{\mathrm{tr}}\in A]
    \ge
    \widehat\alpha_{\mathcal P}Q_{\mathcal M}(A)
    \qquad
    \text{for every union }A\text{ of cells of }\mathcal P.
    \label{eq:empirical_partition_domination}
\end{equation}
Thus \(\widehat\alpha_{\mathcal P}\) directly quantifies finite-resolution
regional routing on the partition.

For retention, we report the \(99\%\) and \(99.9\%\) quantiles of
\[
    d_{\mathcal M}
    \bigl(
        Z_{\mathrm{stop}},Z_{\mathrm{tr}}
    \bigr).
\]
These quantiles test the high-probability analogue of the deterministic
retention bound in
Proposition~\ref{prop:early_stopping_retention}. We also report the
sampling-resolution statistic \(\varepsilon_M^{\mathrm{loc}}\),
defined as the \(Q_{\mathcal M}\)-weighted \(99\%\) quantile of the
ambient Euclidean distance to the nearest training center. This is a quantile-based local coverage
radius, not a certificate of the posterior-moment residual \(h_M\) in
Appendix~\ref{app:finite_sample_refinements}. An ablation
at \(\kappa=4\) with
\(M\in\{10^3,5\cdot10^3,10^4\}\), using three independent center
seeds at each sample size, tests whether the observed finite-sample
resolution is consistent with the expected \(M^{-1/2}\) behavior on
a two-dimensional manifold.

A high-probability displacement bound gives a coverage bound by subtracting the probability of its failure event.

A population \(99.9\%\) retention guarantee would lose at most
\(10^{-3}\) in the coverage lower bound. For an exact empirical
statement, let \(\widehat\delta_{0.999}\) be the upper empirical
quantile of the unrounded displacements, that is, the smallest observed
threshold for which at least \(99.9\%\) of trajectories have displacement
at most that threshold. This order-statistic definition, rather than a
rounded table entry or an arbitrary interpolation convention, guarantees
\[
 \widehat{\mathbb P}[d_{\mathcal M}(Z_{\mathrm{stop}},Z_{\mathrm{tr}})
                 \le\widehat\delta_{0.999}]\ge0.999.
\]
For any Borel region \(U\), retain only partition cells entirely inside
its safe interior. The event-inclusion proof and
\eqref{eq:empirical_partition_domination} give
\begin{equation}
 \widehat{\mathbb P}[Z_{\mathrm{stop}}\in U]
 \ge\widehat\alpha_{\mathcal P}\,
 Q_{\mathcal M}\!\left(
   \bigcup_{\substack{B\in\mathcal P:\\
             B\subseteq U^{-\widehat\delta_{0.999}}}} B
 \right)-10^{-3}.
 \label{eq:empirical_partition_coverage_transfer}
\end{equation}
When the safe interior is itself a union of cells, the union equals
\(U^{-\widehat\delta_{0.999}}\). This bound applies to the empirical
trajectory distribution. Use the unrounded order statistic in the bound
because the printed p99 and p99.9 values are rounded summaries.

\paragraph{Results.}
Table~\ref{tab:e1_routing} summarizes the five target distributions.
At the tested partition resolution, the regional probability lower bound remains
uniformly bounded away from zero:
\(\widehat\alpha_{\mathcal P}\ge0.789\) across all vMF
concentrations and \(\widehat\alpha_{\mathcal P}\ge0.937\) for the
cluster mixtures. The binding cell lies at or near the high-density
pole region for the vMF targets and in one of the two component
hemispheres for the cluster targets; for \(p=0.9\), the measured mass
split is \(0.846\). The ratio curves are consistent with mild
broad-scale flattening: the transition distribution undersamples the highest-
density cells relative to \(Q_{\mathcal M}\), while lower-density
cells are oversampled. The high-density cells determine the reported lower bound; oversampling
the other cells is not penalized by the one-sided criterion. The
strictly positive values of
\(\widehat\alpha_{\mathcal P}\) provide direct finite-resolution
evidence for regional probability allocation beyond the exact symmetry setting
of Proposition~\ref{prop:sphere_routing}.

The retention quantiles are also consistent with the predicted scale.
Across all configurations, the p99.9 displacement is at most \(0.112\),
below both the population benchmark
\(\delta_{\mathrm{stop}}^{\mathrm{pop}}=0.119\) and the
curvature-inflated reference value \(0.134\). Together with
\eqref{eq:empirical_partition_coverage_transfer}, this shows that the
observed regional allocation is largely retained through refinement at the
tested resolution.

\begin{figure}[!htb]
    \centering
    \includegraphics[width=0.5\linewidth]{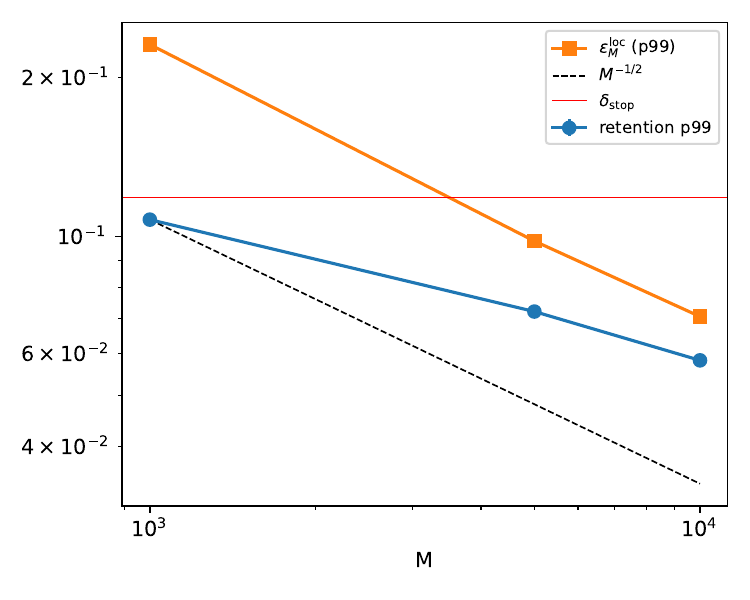}
    \caption{E1 ablation (\(\kappa=4\), three center seeds per \(M\)):
    retention quantile p99 and the measured \(99\%\) nearest-center
    coverage radius \(\varepsilon_M^{\mathrm{loc}}\) versus \(M\),
    with an \(M^{-1/2}\) guide and the idealized population benchmark
    \(\delta_{\mathrm{stop}}^{\mathrm{pop}}\). The displayed affine
    relation
    \(\mathrm{p99}\approx0.037+0.31\,\varepsilon_M^{\mathrm{loc}}\)
    is a descriptive fit over the three tested sample sizes.}
    \label{fig:e1_ablation}
\end{figure}

The \(M\)-ablation in Figure~\ref{fig:e1_ablation} examines the
observed dependence on sampling resolution. The measured
\(\varepsilon_M^{\mathrm{loc}}\) contracts by a factor \(3.27\) from
\(M=10^3\) to \(M=10^4\), consistent with the \(M^{-1/2}\) reference
rate \(\sqrt{10}\approx3.16\). Across the three tested sample sizes,
the descriptive affine fit is
\[
    \mathrm{p99}
    \approx
    0.037+0.31\,\varepsilon_M^{\mathrm{loc}}.
\]
The intercept is an extrapolation toward vanishing sampling-resolution
radius, while the second term records the observed finite-sample dependence;
the fit is descriptive rather than a decomposition asserted by the
theory. At \(M=10^3\), the empirical quantile
\(0.108\pm0.003\) over three center seeds nearly reaches the
\(0.119\) benchmark, whereas at \(M=10^4\) the seed-to-seed spread of
both \(\widehat\alpha_{\mathcal P}\) and the retention quantile is
below \(2\%\). These observations are consistent with the sampling-
resolution limitation discussed in
Appendix~\ref{app:finite_sample_refinements}.

Reducing the step size by a factor of three,
\(\eta=0.15\to0.05\), with identical centers and initializations,
changes \(\widehat\alpha_{\mathcal P}\) by \(0.04\) and the
retention quantile by \(6\%\). These differences matter at the scale of the
narrowest numerical margins. At both step sizes, partition-level coverage
remains bounded away from zero, and the upper retention quantiles remain
on the scale predicted by
\eqref{eq:explicit_early_stopping_retention_radius}.
\begin{table}[!htb]
\centering
\caption{E1: finite-resolution probability and retention across target
 densities (\(M=10^4\), \(10^5\) trajectories; every trajectory
 reached both stopping surfaces;
 \(\delta_{\mathrm{stop}}^{\mathrm{pop}}=0.119\)).
 \(\widehat\alpha_{\mathcal P}\) is the minimum, over twelve
 equal-\(Q_{\mathcal M}\)-mass colatitude bins, of the marginal
 two-sided \(95\%\) Wilson lower endpoint for the empirical bin mass
 divided by its target mass; the binding bin is reported in
 \(t=\cos\theta\). For the cluster mixtures, the partition consists
 of the two component hemispheres, and the measured mass split is also
 reported. The Wilson intervals are conditional on the realized center
 cloud and are not a simultaneous confidence band.}
\label{tab:e1_routing}
\label{tab:main_e1}
\begin{tabularx}{\linewidth}{@{}l c >{\centering\arraybackslash}X c c c@{}}
\toprule
Density & \(\widehat\alpha_{\mathcal P}\) & binding bin &
split & p99 & p99.9 \\
\midrule
vMF, \(\kappa=1\)  & \(0.789\) & \([0.93,1.00]\) & --- & \(0.039\) & \(0.052\) \\
vMF, \(\kappa=4\)  & \(0.799\) & \([0.98,1.00]\) & --- & \(0.058\) & \(0.106\) \\
vMF, \(\kappa=16\) & \(0.852\) & \([0.989,0.995]\) & --- & \(0.046\) & \(0.085\) \\
clusters, \(p=0.5\) & \(0.987\) & binding hemisphere & \(0.503\) & \(0.056\) & \(0.109\) \\
clusters, \(p=0.9\) & \(0.937\) & heavier-component hemisphere & \(0.846\) & \(0.059\) & \(0.112\) \\
\bottomrule
\end{tabularx}
\end{table}

\begin{figure}[!htb]
    \centering
    \includegraphics[width=0.5\linewidth]{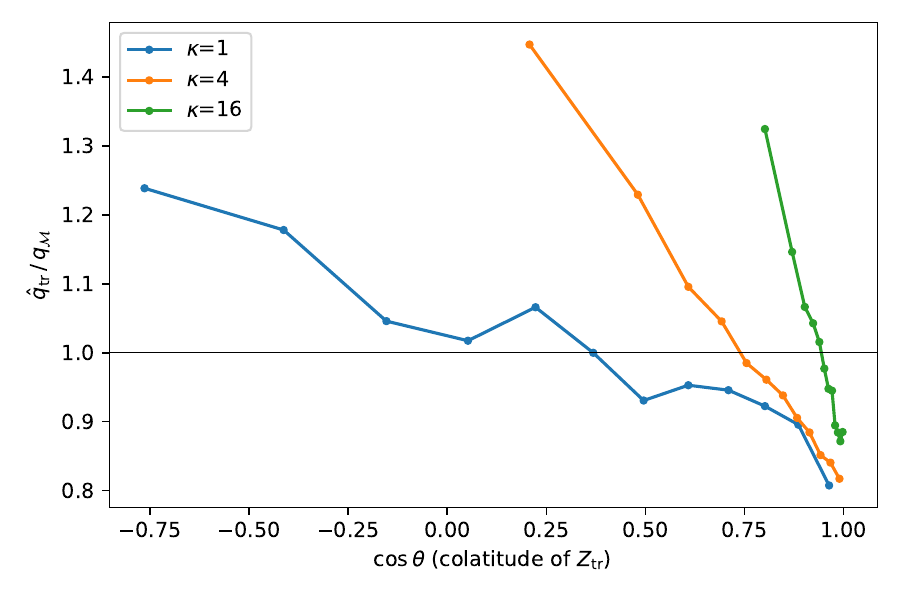}
    \caption{E1: binned density ratio
        \(\widehat q_{\mathrm{tr}}/q_{\mathcal M}\) versus
        \(\cos\theta\) on equal-\(Q_{\mathcal M}\)-mass bins, each
        plotted at the \(\cos\theta\) value splitting its target mass
        in half, for vMF targets with
        \(\kappa\in\{1,4,16\}\). The ratio is smallest in or near the
        high-density pole region and exceeds one in the tails, a pattern
        consistent with mild flattening by the broad scales.}
    \label{fig:e1_vmf_ratio}
\end{figure}

\clearpage
\subsection{E2: Off-tube critical structure}
\label{sec:e2}

\subsubsection{Verified off-tube instability and learned trajectories on finite spheres}
\label{subsec:revision_e2}
For each of three independent clouds, we draw 512 unit-sphere points with density proportional to $e^{u_3}$, where $u_3$ is the third coordinate. Appending nine zeros embeds the cloud in $\mathbb R^{12}$. We train noise-unconditioned MLPs with three width-128 SiLU layers using the corrupted state and $\log(1+\|x\|)$ as inputs. DSM corruptions use 24 log-spaced levels in $[0.05,2]$. The exact target is the empirical unified score for that same cloud and grid.

\paragraph{Off-tube assumption for the exact field.}
At a critical point, $x_\perp=0$ because the normal score outside the sphere's three-dimensional span equals $-\beta(x)x_\perp$. The covariance of component scores is therefore supported in three dimensions. For the unit-sphere cloud, off-tube critical points with $r_0=0.2$ have radius at most $0.8$ because they belong to the convex hull of the centers. At criticality $\operatorname{tr}(C_x)/\beta(x)$ is the precision-weighted average of $\|x-\mu\|^2/\sigma_i^2$. For each center at distance $t$, the conditional scale weights are proportional to $\sigma_i^{-12}e^{-t^2/(2\sigma_i^2)}$. We enclose the corresponding mean of $t^2/\sigma_i^2$ in intervals of width $0.001$ on $[0.199,1.801]$, allowing for the stored centers' rounding. Each lower bound exceeds $10.27$, and averaging over centers preserves the bound. With $\alpha_d=0.5$ and $R_d=3$, this gives $R_d<\alpha_d d=6<10.27$ in Assumption~\ref{assumption:off_tube_geometry}.

\paragraph{Exact and learned trajectories.}
Each paired comparison uses the same 512 initial states from the central $99.9\%$ radial shell of $\mathcal N(0,2^2 I_{12})$. Both fields use the exact precision normalization, a positive rescaling that preserves their continuous paths. We simulate 600 Euler updates with step size $0.03$ and interpolate along the crossing step for entry into the distance-$0.03$ tube. We compare the projected first-entry frequencies against twelve equal-probability bands of the continuous vMF law, counting failures in a separate bin. The relative score RMSE is $\sqrt{\mathbb E\|s_\theta-b\|^2/\mathbb E\|b\|^2}$ on independent noisy queries.

\begin{table}[!htb]
\centering
\small
\caption{Ordinary DSM on the finite sphere mixtures. Values are mean $\pm$ sample standard deviation across three cloud and training seeds. Exact-score regional TV is $0.08\pm0.03$.}
\label{tab:revision_sphere_dsm}
\begin{tabular}{@{}lrrr@{}}
\toprule
Updates & Relative score RMSE & Tube entry (\%) & Regional TV \\
\midrule
1k & $0.684\pm0.007$ & $7.9\pm3.1$ & $0.92\pm0.03$ \\
5k & $0.358\pm0.004$ & $38.3\pm2.8$ & $0.62\pm0.03$ \\
20k & $0.256\pm0.005$ & $99.9\pm0.2$ & $0.07\pm0.03$ \\
\bottomrule
\end{tabular}
\end{table}

Smaller overall score errors accompany smaller sampled outward errors.
Appendix~\ref{sec:learned_circle} derives first-entry coverage bounds from
uniform enclosures of learned fields on a circle.

\subsubsection{Population radial scores and empirical critical-point diagnostics}
\paragraph{Setup.}
We probe the off-tube geometry of the multi-scale score field in two
complementary ways, with $R=1$ throughout.

First, for the population uniform measure on
$S^m(1)\subset\mathbb R^{m+c}$, we evaluate the radial score
$g(\varrho)$ of Lemma~\ref{lemma:sphere_critical_reduction} by summing
over a finite geometric scale schedule and applying Gauss--Legendre
quadrature in colatitude. The baseline uses 200 scales on $[0.01,2]$
and 400 quadrature nodes; we also test a coarser 40-scale schedule on
$[0.01,2]$ and a wider 200-scale schedule on $[0.005,4]$.
We evaluate positivity on $(0,R-r_0]$ with $r_0=0.3$.

Second, for empirical mixtures with $M=2000$ centers on
$S^2(1)\subset\mathbb R^{12}$, sampled either uniformly or from
$\mathrm{vMF}(\kappa=4)$, we search for zeros of the exact score using
a damped Gauss--Newton (Levenberg--Marquardt) iteration. At each
$\sigma_{\min}\in\{0.2,0.1,0.05,0.02\}$, we use 24 geometrically
spaced scales on $[\sigma_{\min},2]$, keeping the center cloud fixed
throughout the coarse-to-fine sweep.

At every level, we draw 2048 fresh off-tube initializations: the
direction is uniform on the ambient unit sphere, the ambient radius is
sampled uniformly from $[0.02,2R]$, and only points with $r(x)>r_0$
are retained. Deduplicated roots from the preceding noise level are
additionally used as warm starts. Writing
$s(x)=\nabla\log p_{MN}(x)$ and
$J(x)=\nabla^2\log p_{MN}(x)$, each LM step solves
\begin{equation}
  \left[J^\top J+\lambda\,\bar d_J I\right]p
  =J^\top s,
  \qquad
  \bar d_J=\frac{1}{d}\operatorname{tr}(J^\top J),
  \label{eq:e2_lm_update}
\end{equation}
and proposes $x^+=x-p$. The proposal is accepted when the normalized
residual decreases. The damping starts at $10^{-6}$, is multiplied by
$0.3$ after acceptance and by $10$ after rejection, and the search is
run for at most 150 iterations. A numerical damping floor of \(10^{-14}\) is used, and runs with
\(\|x\|>3\sigma_{\max}\) or \(\lambda>10^{10}\) are marked as
diverged or stalled. Convergence is declared when
\begin{equation}
  \frac{\|\nabla\log p_{MN}(x)\|}{\beta(x)}
  =
  \|\bar\mu(x)-x\|
  \le10^{-10}.
  \label{eq:e2_root_tolerance}
\end{equation}
Converged candidates are greedily deduplicated in ambient Euclidean
distance with tolerance $10^{-5}$; we additionally sweep tolerances
from $10^{-7}$ to $10^{-2}$.

At each located critical point we evaluate the analytic log-density
Hessian
\begin{equation}
  \nabla^2\log p_{MN}(x)=C_x-\beta(x)I
  \label{eq:e2_hessian_identity}
\end{equation}
and record its spectrum. We also measure the precision-normalized
posterior spread
\begin{equation}
  \Sigma_x
  :=
  \sum_{j=1}^{M}\sum_{i=1}^{N}
  \widetilde w_{ij}(x)
  (\mu_j-x)(\mu_j-x)^\top.
  \label{eq:empirical_precision_spread}
\end{equation}
This quantity captures the geometry of the active centers but is
distinct from the component-score covariance $C_x$ in
Assumption~\ref{assumption:off_tube_geometry}. At a critical point,
\begin{equation}
  C_x
  =
  \beta(x)\sum_{j,i}\widetilde w_{ij}(x)
  \frac{(\mu_j-x)(\mu_j-x)^\top}{\sigma_i^2}.
  \label{eq:e2_Cx_sigma_relation}
\end{equation}
At the population center of the uniform sphere, center and scale
posteriors factorize, so $C_x\propto\Sigma_x$ and their effective
ranks coincide.

For a complementary center-utilization diagnostic, define the
center-marginal responsibility
\begin{equation}
  q_j(x)
  =
  \frac{\sum_i\sigma_i^2
  \mathcal N(x;\mu_j,\sigma_i^2I)}
  {\sum_{\ell,i}\sigma_i^2
  \mathcal N(x;\mu_\ell,\sigma_i^2I)},
  \qquad
  \operatorname{ESS}(x)
  =
  \frac{1}{\sum_j q_j(x)^2}.
  \label{eq:e2_center_ess}
\end{equation}
Finally, when all centers lie on $S^m(R)$,
\begin{align}
  \operatorname{tr}(\Sigma_x)
  &=
  R^2-\|x\|^2
  -2\left\langle
  x,\frac{\nabla\log p_{MN}(x)}{\beta(x)}
  \right\rangle,
  \label{eq:e2_sigma_trace_general}\\
  \operatorname{tr}(\Sigma_x)
  &=
  R^2-\|x\|^2
  \qquad\text{at a critical point}.
  \label{eq:sphere_sigma_trace_identity}
\end{align}
We use this identity as a numerical consistency check.

\begin{figure}[!htb]
  \centering
  \includegraphics[width=\linewidth]{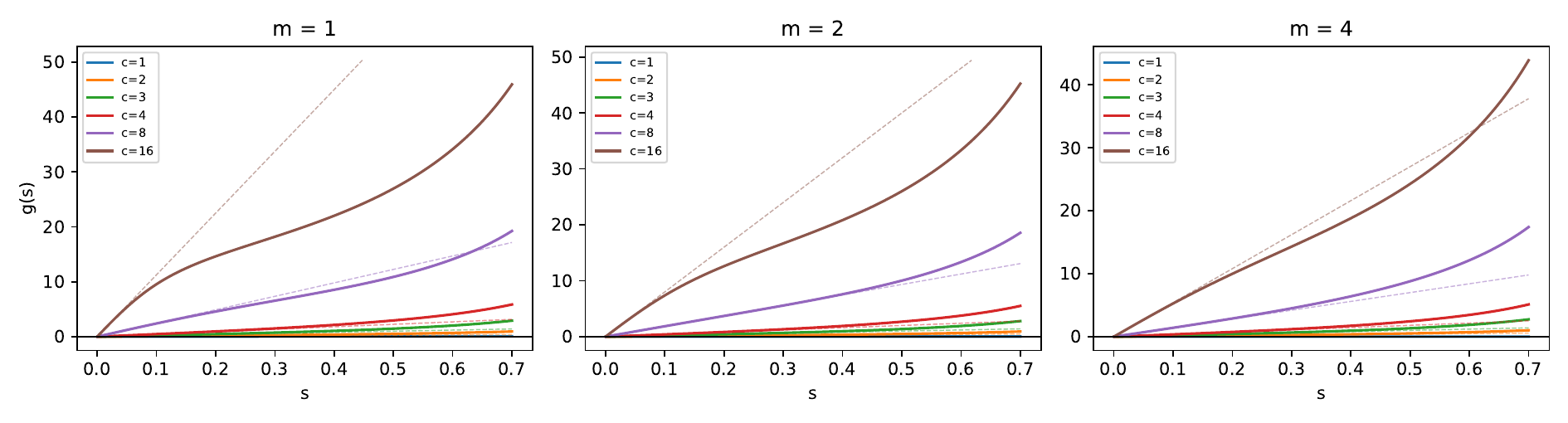}
  \caption{\textbf{Population off-tube radial score.}
  Radial score $g(\varrho)$ on
  $[10^{-3},R-r_0]$ for
  $m\in\{1,2,4\}$ and
  $c\in\{1,2,3,4,8,16\}$.
  The finite scale sum is combined with Gauss--Legendre quadrature in
  colatitude. Dashed lines show the post-hoc small-$\varrho$ reference
  $k(m,c)\varrho$, where
  $k=(c-1)(c+m-2)/(m+1)$.
  The score is positive at every tested grid node.}
  \label{fig:e2_radial}
\end{figure}

\paragraph{(i) Population radial-score positivity.}
For every tested pair
$(m,c)\in\{1,2,4\}\times\{1,2,3,4,8,16\}$ and all three scale
schedules, $g(\varrho)>0$ at every node of the 120-point grid on
$[10^{-3},R-r_0]$ (Figure~\ref{fig:e2_radial}). The minimum always
occurs at the smallest tested radius, consistent with linear vanishing
near the origin for $c\ge2$.

The post-hoc small-$\varrho$ slope, estimated from the first six grid
nodes, approaches
\begin{equation}
  k(m,c)
  =
  \frac{(c-1)(c+m-2)}{m+1},
  \label{eq:e2_reference_slope}
\end{equation}
which equals the unstable eigenvalue $\lambda_+$ in
Proposition~\ref{prop:sphere_origin} at $R=1$. The agreement is within
$0.15\%$ for $c\ge8$; for example, the measured and predicted values
are $112.38$ and $112.5$ at $(m,c)=(1,16)$ and $14.0006$ and $14$
at $(m,c)=(4,8)$. At moderate codimension, the finite-schedule excess
decreases as the scale range widens: for $(m,c)=(1,2)$, the ratio of
measured to predicted slope decreases from $1.70$ to $1.25$ when the
schedule is extended from $[0.01,2]$ to $[0.005,4]$.

The $c=1$ cases show the expected schedule sensitivity. Their measured
slope remains positive for all three tested schedules, but decreases
toward the continuum knife edge as the range widens; at $m=4$, for
example, it decreases from $4.3\times10^{-3}$ to
$1.4\times10^{-4}$. For $c\ge2$, these results support the radial
positivity mechanism used to exclude noncentral off-tube critical
points in the sphere model.

\begin{figure}[t]
  \centering
  \includegraphics[width=\linewidth]{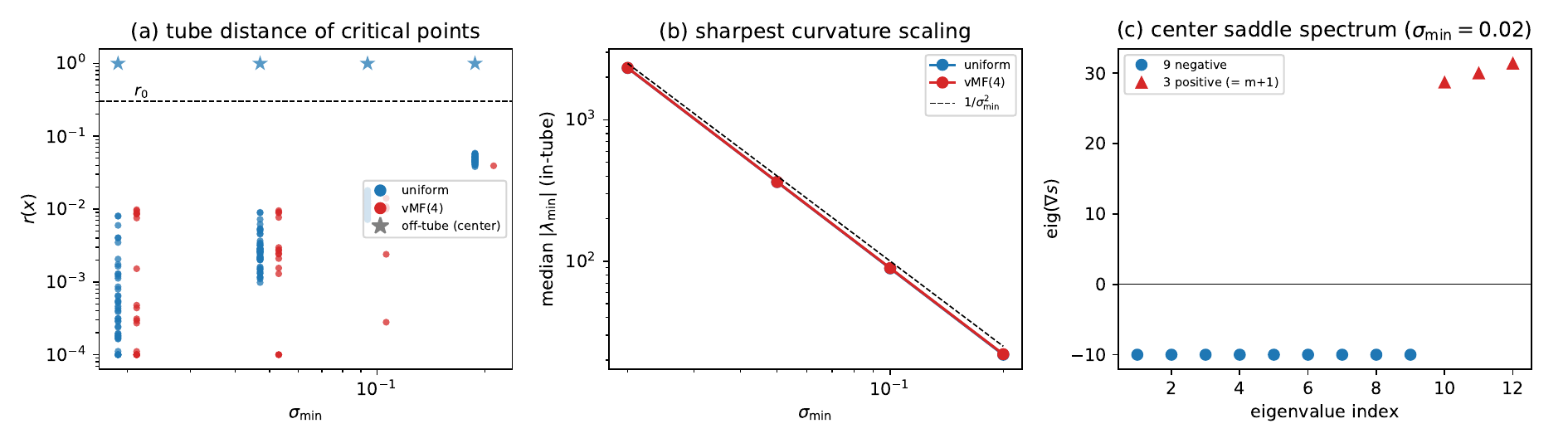}
  \caption{\textbf{Critical structure of empirical sphere mixtures.}
  $M=2000$ centers lie on
  $S^2(1)\subset\mathbb R^{12}$.
  (a) Tube distance of all located deduplicated critical points across
  the $\sigma_{\min}$ sweep; the only located off-tube point is the
  near-center saddle of the uniform cloud.
  (b) Median magnitude of the most negative Hessian eigenvalue for
  in-tube critical points, compared with the
  $1/\sigma_{\min}^2$ scale.
  (c) Hessian spectrum of the near-center uniform-cloud saddle, with
  three positive and nine negative eigenvalues.}
  \label{fig:e2_crit}
\end{figure}

\paragraph{(ii) Empirical critical points.}
For the uniform cloud, the search locates one deduplicated off-tube
critical point at every $\sigma_{\min}$ and no second off-tube root
(Figure~\ref{fig:e2_crit}a). It is a finite-sample perturbation of the
population center, with
$\|x\|\approx3\times10^{-3}$ and hence $r(x)\approx R$.
Across the sweep, this root receives $685$--$742$ converged solver
starts and remains a single root throughout the deduplication-tolerance
sweep.

Its local geometry closely matches the population prediction. The
log-density Hessian has $m+1=3$ positive and nine negative eigenvalues
(Figure~\ref{fig:e2_crit}c), giving the predicted three-dimensional
ascent-unstable subspace. The posterior-spread diagnostics give
\begin{equation}
  \operatorname{tr}(\Sigma_x)=0.99999\approx R^2,
  \qquad
  r_{\mathrm{eff}}(\Sigma_x)=2.90\approx m+1,
  \label{eq:e2_center_spread_results}
\end{equation}
with
$r_{\mathrm{eff}}(\Sigma_x)/(m+1)=0.97$.
The center-marginal ESS is approximately $1999$ out of $2000$, showing
that the near-center saddle is supported by essentially the entire
uniform cloud rather than a single training center or a small local
cluster.

Every other located critical point lies well inside the tube. For the
uniform cloud, $r(x)\le0.059$; for the vMF cloud,
$r(x)\le0.040$ at $\sigma_{\min}=0.2$ and
$r(x)\le0.015$ at all sharper scales. Their center ESS ranges from
approximately $1$ for sample-dominated modes to approximately $176$.
The median most negative Hessian eigenvalues across the four lower
cutoffs are
\begin{equation}
  -21.8,\;-88.7,\;-362,\;-2.33\times10^3,
  \label{eq:e2_intube_curvature_values}
\end{equation}
close to the finest-scale reference
$-1/\sigma_{\min}^2=
-25,-100,-400,-2500$.

For the $\mathrm{vMF}(4)$ cloud, no off-tube critical point is located
at any tested $\sigma_{\min}$. The score at the origin instead points
toward the density pole. At $\sigma_{\min}=0.2$, the located critical
structure consists only of the density mode near the pole, at polar
angle $3.1^\circ$. At finer scales, additional in-tube finite-sample
structures appear predominantly in lower-density regions; at
$\sigma_{\min}=0.02$, $16$ of the $20$ located points lie beyond
$50^\circ$ from the pole.

\paragraph{Summary and scope.}
The two parts give complementary evidence for the global landscape
picture. Population radial scores remain positive throughout the tested
off-tube grid, while the off-tube critical point located in the finite
uniform mixture has precisely the predicted ascent-unstable geometry.
Breaking uniform symmetry with a vMF cloud removes the center critical
point, and the search locates no replacement outside the tube.

The empirical root search is not exhaustive, particularly at the finest
vMF setting where cold-start convergence becomes sparse; coarse-to-fine
continuation substantially improves the root yield. We therefore use
E2 as numerical support for the off-tube instability mechanism rather
than as a census of all finite-mixture critical points. Together with E1,
these results support the geometric ingredients of the convergence and
coverage analysis in finite, nonuniform settings.

\subsection{E3: Positive-bandwidth separation and sample-resolution diversity}
\label{sec:e3}

\subsubsection{Finite-grid verification and cutoff comparison}
\label{subsec:revision_e3}
\paragraph{Setting and readout.}
The 82 centers are $(k\pm1/6,0)$ for $k=-20,\ldots,20$ in $\mathbb R^{51}$, with reference law uniform on the segment $[-20.5,20.5]\times\{0\}$. The grid spacing parameter is $\ell=1$. Consecutive center gaps are $1/3$ or $2/3$, and each boundary gap is $1/3$, so $h_{\mathrm{cov},M}=1/3$. We use the separation margin $h_{\mathrm{cov},M}/4=1/12$ from Corollary~\ref{cor:main_nonmemorization}. We evaluate three floors $s=\sigma_{\min}/\ell=0.50,0.55,0.60$ with $N=32{,}768$ log-spaced levels and $\sigma_{\max}=50$ throughout. The floors meet $s\ge1/2$ and were selected using a preliminary bandwidth sweep.

For each setting, we take three seeds with 1,024 samples per method and seed. The full optimization flow starts from $\mathcal N(0,\sigma_{\max}^2 I_{51})$ and uses the exact empirical unified score. The comparison draws from the exact empirical diffusion marginal $\widehat p_{\sigma_{\min}}$ and takes the single-scale Tweedie readout in Appendix~\ref{subsec:diffusion_endpoint_comparison}. All outputs count, including edge outcomes. We evaluate the finite field using the normal contraction and the one-dimensional tangential equation, and cross-check the endpoint classifications under tighter solver settings.

\begin{table}[!htb]
\centering
\small
\setlength{\tabcolsep}{7pt}
\caption{Finite-grid E3. Distances are Euclidean distances to the nearest training point. Cutoff denotes the exact cutoff marginal followed by the Tweedie readout.}
\label{tab:revision_affine}
\begin{tabular}{@{}rcccc@{}}
\toprule
 & \multicolumn{2}{c}{Mean distance} & \multicolumn{2}{c}{Distance $\ge1/12$ (\%)} \\
\cmidrule(lr){2-3}\cmidrule(lr){4-5}
$\sigma_{\min}$ & Flow & Cutoff & Flow & Cutoff \\
\midrule
$0.50$ & $0.1651$ & $0.1321$ & $100.00$ & $64.78$ \\
$0.55$ & $0.1586$ & $0.1342$ & $91.28$ & $64.88$ \\
$0.60$ & $0.1663$ & $0.1360$ & $100.00$ & $65.85$ \\
\bottomrule
\end{tabular}
\end{table}

\paragraph{Regional mass and separation.}
The three reference regions separated by $-6.5$ and $6.5$ have masses $(14,13,14)/41$. Empirical TV for the flow is $0.0114$ at all three floors. At the finer 41-cell resolution, empirical TV is $0.094$, $0.094$, and $0.142$, versus $0.049$, $0.051$, and $0.045$ for cutoff plus Tweedie. The flow increases the training-set distance while retaining coarse regional mass, and the readout better matches cell masses at the finer resolution in this comparison.

For these finite configurations we check critical points directly on the line. The center convex hull contains all critical points. We enclose the tangential score and its derivative using all centers and scales, with interpolation remainders and interval bounds for each subinterval. Normal Hessian eigenvalues are negative on the line. Tangential minima are hyperbolic saddles, and their attraction sets have probability zero under the Gaussian prior. At $s=0.50$ and $0.60$, the intervals containing possible maxima are all separated by at least $1/12$ from the training set. At $s=0.55$, an edge pair near $\pm18.908$ falls below that margin. The interval $[-18.5,18.5]$ is forward invariant in the tangential coordinate, as reflection at each boundary leaves only extra inward centers. It contains only separated possible maxima.

For the probability check, apply \eqref{eq:cutoff_endpoint_transfer} with the empirical reference. From the cutoff law, we account for the normal tail, strips of width $0.001$ on either side of the two coarse boundaries, and the finite-grid discrepancy in the tangential marginal. On the retained cylinder, the one-dimensional score points inward at the protected region boundaries, as verified by bounding the omitted lattice tails against the periodic score. The total variation upper bounds are $0.05545$, $0.05408$, and $0.05286$. The same calculation for the invariant interval at $s=0.55$ gives at least $85.65\%$ mass on margin-separated endpoints. Together with the all-mode check at the other two floors, the joint lower bounds for landing in any fixed coarse region and meeting the margin are $26.16\%$, $24.43\%$, and $26.42\%$ at the three floors.

The critical-point check uses analytic interpolation remainders, outward interval evaluation, and explicit binary64 finite-sum and transform error allowances, including a per-Gaussian elementary-operation allowance of $10^{-11}$. The guarantees rely on these numerical enclosures for the specified finite fields. Appendix~\ref{app:resolved_endpoint_proof} establishes a sufficient construction over a bandwidth window under additional hypotheses beyond $s\ge1/2$.

\subsubsection{Bandwidth sweep on a 24-level schedule}
\paragraph{Setup and exact field evaluation.}
We take $\mathcal A=\mathbb R\times\{0\}\subset\mathbb R^3$ and the interlaced grid \eqref{eq:resolved_grid_training_centers} with $L=20$ and $\ell=1$. Thus $M=82$, $Q_{\mathcal M}$ is uniform on $[-20.5,20.5]\times\{0\}$, and there are 41 reference cells with output cores $C_k=\{(u,0):|u-k|\le1/12\}$. The training covering radius and the separation threshold are, respectively, $h_{\mathrm{cov},M}=1/3$ and $h_{\mathrm{cov},M}/4=1/12$.
For each $s=\sigma_{\min}/\ell$ listed in Table~\ref{tab:e3_bandwidth}, we use
\begin{equation}
  \sigma_i=s\,128^{(i-1)/23},\quad i=1,\ldots,24,
  \qquad X_0\sim\mathcal N(0,(128s)^2I_3).
  \label{eq:e3_numerical_schedule}
\end{equation}
We fix the relative schedule across the sweep, so $s$ controls the lower
cutoff, the upper cutoff, and the initialization variance jointly.

At each setting we simulate $n_{\mathrm{traj}}=8192$ initializations using the update $x_{n+1}=x_n+\eta(\bar\mu(x_n)-x_n)$, with $\eta=0.2$ in the main sweep. Gaussian sums are evaluated in PyTorch single precision using stabilized log weights. At a fixed scale, the center sum on the Cartesian-product grid factorizes into one-dimensional sums. We combine the scale contributions using their precision-normalized responsibilities to evaluate the exact finite-mixture barycenter up to floating-point error.

\paragraph{Termination and numerical output.}
The implementation checks
\(\max_a\|\bar\mu(x_n^{(a)})-x_n^{(a)}\|\) every 50 iterations,
with threshold \(10^{-5}\) and a cap of 5000 updates. The residual is
evaluated before the checked update, and the returned state includes that
update. We denote the returned numerical output by \(X_{\mathrm{num}}\),
distinguishing it from the exact limit \(x_\infty\). Every reported run
reaches the threshold before the cap. In the main sweep the update counts
range from 101 to 2451, and the largest recorded maximum residual is
\(9.72\times10^{-6}\).

\paragraph{Metrics at the training resolution.}
Let $X_{\mathrm{num}}^{(a)}$ be the returned ambient outputs and $U^{(a)}=\Pi_{\mathcal A}(X_{\mathrm{num}}^{(a)})$ their affine projections. The separation statistic and output-core frequencies are
\begin{align}
  \widehat p_{\mathrm{sep}}
  &=\frac1{n_{\mathrm{traj}}}\sum_{a=1}^{n_{\mathrm{traj}}}
    \mathbf1\!\left\{\operatorname{dist}(X_{\mathrm{num}}^{(a)},\mathcal S_M)\ge1/12\right\},
    \label{eq:e3_separation_statistic}\\
  \widehat p_k
  &=\frac1{n_{\mathrm{traj}}}\sum_{a=1}^{n_{\mathrm{traj}}}
    \mathbf1\{U^{(a)}\in C_k\},\qquad
  \widehat p_{\mathrm{out}}=1-\sum_{k=-20}^{20}\widehat p_k.
    \label{eq:e3_core_frequencies}
\end{align}
The categorical reference assigns mass $1/41$ to each reference-cell label and zero to the outside-core label. As in \eqref{eq:resolved_endpoint_cell_mass}, reference labels describe the full cells of $Q_{\mathcal M}$, whereas output labels describe the smaller cores; the comparison is not with $Q_{\mathcal M}(C_k)$. We report
\begin{equation}
  \widehat\epsilon_{\mathrm{cell}}
  =\frac12\left[\sum_{k=-20}^{20}|\widehat p_k-1/41|
                                  +\widehat p_{\mathrm{out}}\right],
  \qquad
  \widehat n_{\mathrm{pop}}
  =\#\{k:\widehat p_k\ge1/82\}.
  \label{eq:e3_cell_tv_and_count}
\end{equation}
The populated-core fraction is $\widehat n_{\mathrm{pop}}/41$. All 8192 outputs enter the denominators; neither the separation statistic nor the core-label distribution is conditioned on the other. We also report empirical quantiles of the ambient distance to \(\mathcal S_M\).

Membership of a central core implies the training-distance margin, but the reverse implication does not hold for arbitrary locations between cells. We therefore retain separation and core mass as distinct diagnostics. For example, at $s=0.26$ their measured values are $0.9583$ and $0.9319$. Likewise, zero populated \emph{cores} at small bandwidth does not mean that outputs occupy only one full reference cell: sample-associated modes can remain distributed over the grid while missing the prescribed off-training cores.

\paragraph{Bandwidth sweep.}
Table~\ref{tab:e3_bandwidth} reports the main sweep. For the tested $s\le0.12$, no output meets the separation threshold or lies in a central core. At $s=0.13$, the separated fraction is $0.5875$; it reaches $0.9780$ at $s=0.14$. The sampled bandwidths from $0.15$ through $0.24$ retain 39 populated cores. At $s\in\{0.18,0.20,0.22,0.24\}$, separation is $0.9679$--$0.9734$, core-label TV is $0.0888$--$0.1060$, and the fifth-percentile distance is $0.0992$--$0.1228$, above $1/12$. Increasing \(s\) to \(0.30\) gives TV \(0.1623\), 35 populated cores, and separation \(0.8916\). Thus the strongest joint separation--coverage behavior occurs over an intermediate bandwidth range rather than varying monotonically with smoothing.

\begin{table}[!htb]
\centering
\small
\caption{\textbf{E3 bandwidth sweep.} Each row uses 8192 numerical outputs with $\eta=0.2$. The separation threshold is $1/12$; core mass is $1-\widehat p_{\mathrm{out}}$; $\widehat n_{\mathrm{pop}}$ counts the 41 cores whose estimated mass is at least $1/82$; $q_{0.05}$ is the empirical fifth-percentile distance to the training set. All rows meet the recorded residual threshold $10^{-5}$. Statistics are estimates for the simulated outputs, not confidence bounds for the exact flow.}
\label{tab:e3_bandwidth}
\begin{tabular}{@{}rrrrrrr@{}}
\toprule
$s$ & $\widehat p_{\mathrm{sep}}$ & core mass & $\widehat\epsilon_{\mathrm{cell}}$ & $\widehat n_{\mathrm{pop}}$ & $q_{0.05}$ & updates \\
\midrule
0.05 & 0.0000 & 0.0000 & 1.0000 & 0 & 0.00250 & 101 \\
0.08 & 0.0000 & 0.0000 & 1.0000 & 0 & 0.01096 & 151 \\
0.10 & 0.0000 & 0.0000 & 1.0000 & 0 & 0.02388 & 151 \\
0.11 & 0.0000 & 0.0000 & 1.0000 & 0 & 0.03499 & 251 \\
0.12 & 0.0000 & 0.0000 & 1.0000 & 0 & 0.05004 & 301 \\
0.13 & 0.5875 & 0.5875 & 0.4779 & 22 & 0.06941 & 751 \\
0.14 & 0.9780 & 0.9780 & 0.1325 & 38 & 0.09861 & 2451 \\
0.15 & 0.9753 & 0.9753 & 0.1237 & 39 & 0.10467 & 451 \\
0.16 & 0.9780 & 0.9780 & 0.1347 & 39 & 0.12682 & 301 \\
0.18 & 0.9734 & 0.9734 & 0.1060 & 39 & 0.12276 & 251 \\
0.20 & 0.9729 & 0.9729 & 0.0888 & 39 & 0.11978 & 201 \\
0.22 & 0.9701 & 0.9701 & 0.0934 & 39 & 0.11185 & 251 \\
0.24 & 0.9679 & 0.9679 & 0.0901 & 39 & 0.09919 & 301 \\
0.26 & 0.9583 & 0.9319 & 0.1173 & 37 & 0.09669 & 701 \\
0.28 & 0.9305 & 0.9305 & 0.1187 & 37 & 0.05256 & 301 \\
0.30 & 0.8916 & 0.8916 & 0.1623 & 35 & 0.00687 & 351 \\
\bottomrule
\end{tabular}
\end{table}

\begin{figure}[!htb]
  \centering
    \includegraphics[width=0.75\linewidth]{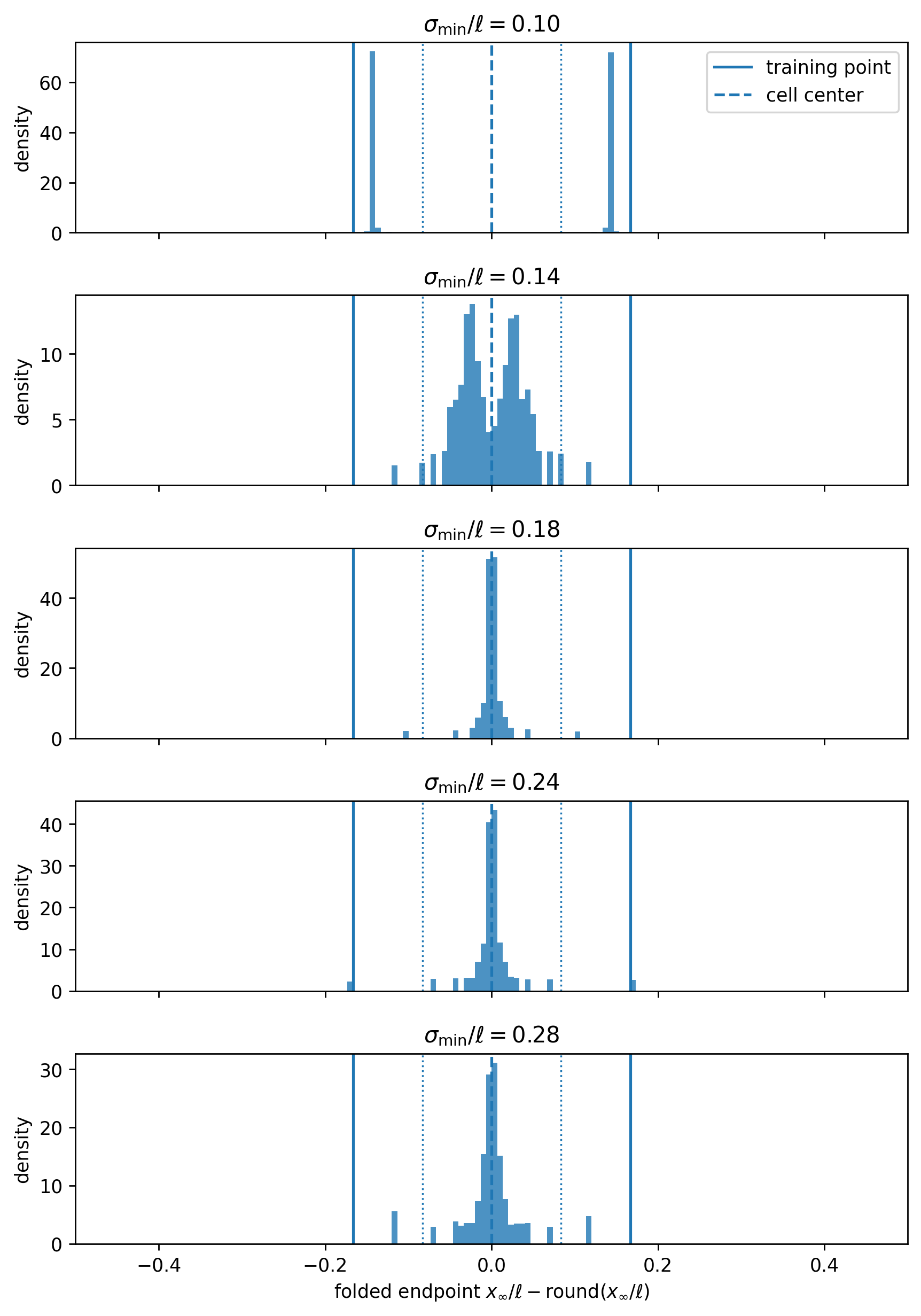}
  \caption{\textbf{Folded numerical outputs across the bandwidth sweep (E3).} The 150 equal-width histogram bins cover $[-1/2,1/2)$; training locations, cell centers, and output-core boundaries are marked. All full-grid outputs are pooled. This plot visualizes within-cell displacement only; diversity across cells is quantified by \eqref{eq:e3_cell_tv_and_count}. The finite-array histograms need not consist of a single exact central atom.}
  \label{fig:e3_folded_full}
\end{figure}

\paragraph{Folded visualization.}
For a returned tangential coordinate $u$, Figure~\ref{fig:e3_folded_full} displays
\begin{equation}
  v=\frac u\ell-\left\lfloor\frac u\ell+\frac12\right\rfloor
    \in[-1/2,1/2).
  \label{eq:e3_folded_coordinate}
\end{equation}
The training locations fold to $\pm1/6$, the cell center to zero, and core boundaries to $\pm1/12$. The displayed histograms pool all returned outputs across the full finite grid, without an interior-cell filter. At $s=0.10$ they retain two sample-associated peaks, and at $s=0.18$ and $0.24$ most mass is near the central part of the cores. Finite-array effects and variation between cells remain visible after folding. We use folding for visualization and compute separation and diversity metrics directly on the finite grid.

\paragraph{Step-size sensitivity.}
We repeat $s\in\{0.14,0.18,0.24\}$ with $\eta=0.1$, the same 8192 initializations at each bandwidth, and the same residual threshold. The seed is $\operatorname{round}(10^4s)$ for both step sizes; different bandwidths use different seeds. Table~\ref{tab:e3_stepsize} gives the paired results. The largest absolute changes are $0.00428$ in separation and $0.00574$ in core-label TV. At $s=0.18$ and $0.24$, the fifth-percentile training distances change by less than $4.4\times10^{-5}$, and the number of populated cores stays at 39. At $s=0.14$, however, the fifth percentile changes from $0.09861$ to $0.08411$, close to the $1/12$ threshold, and the populated-core count changes from 38 to 39. This setting requires 2451 and 4901 updates at the two step sizes. The qualitative separation--coverage pattern persists under step refinement, with the largest sensitivity at \(s=0.14\), where the lower-tail distance lies closest to the separation threshold.

\begin{table}[!htb]
\centering
\small
\caption{\textbf{E3 paired step-size comparison.} The initializations are matched within each bandwidth. The last column is the maximum normalized residual at the final recorded check. The $s=0.14$ lower-tail margin remains sensitive despite similar residuals, whereas the fifth percentiles at $s=0.18$ and $0.24$ change by less than $4.4\times10^{-5}$.}
\label{tab:e3_stepsize}
\begin{tabular}{@{}rrrrrrr@{}}
\toprule
$s$ & $\eta$ & $\widehat p_{\mathrm{sep}}$ & $\widehat\epsilon_{\mathrm{cell}}$ & $q_{0.05}$ & updates & max. residual \\
\midrule
0.14 & 0.2 & 0.9780 & 0.1325 & 0.09861 & 2451 & $9.72\!\times\!10^{-6}$ \\
0.14 & 0.1 & 0.9764 & 0.1268 & 0.08411 & 4901 & $9.78\!\times\!10^{-6}$ \\
0.18 & 0.2 & 0.9734 & 0.1060 & 0.12276 & 251 & $3.81\!\times\!10^{-6}$ \\
0.18 & 0.1 & 0.9700 & 0.1039 & 0.12278 & 451 & $9.54\!\times\!10^{-6}$ \\
0.24 & 0.2 & 0.9679 & 0.0901 & 0.09919 & 301 & $3.81\!\times\!10^{-6}$ \\
0.24 & 0.1 & 0.9636 & 0.0844 & 0.09914 & 501 & $9.54\!\times\!10^{-6}$ \\
\bottomrule
\end{tabular}
\end{table}

\begin{figure}[!htb]
  \centering
    \includegraphics[width=0.75\linewidth]{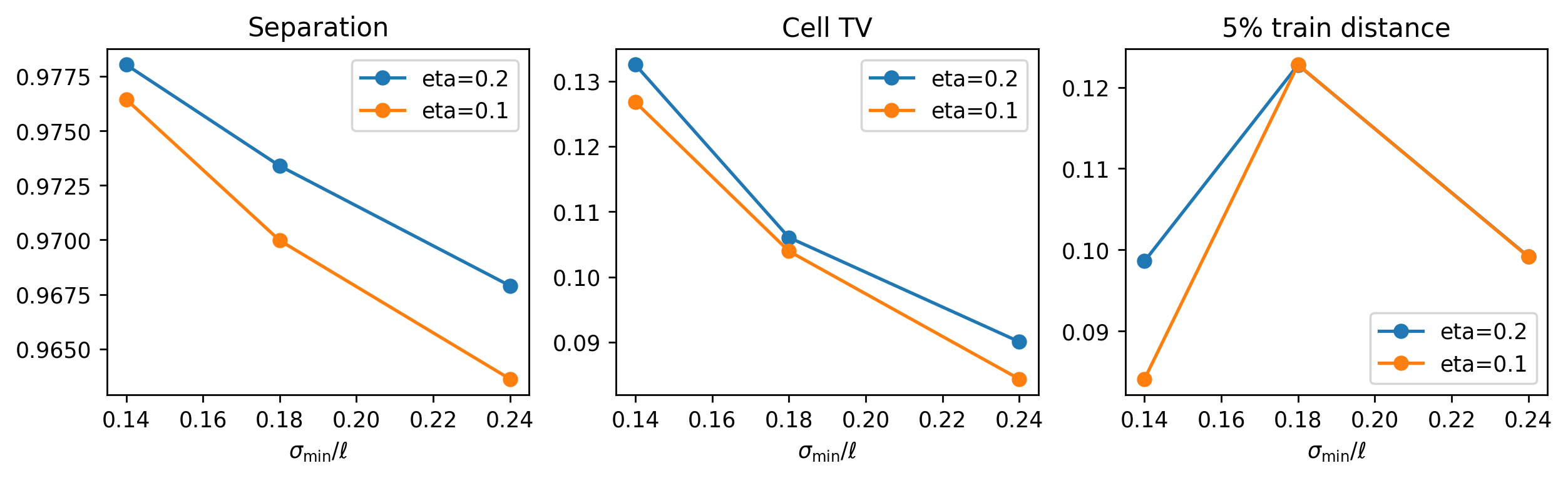}
  \caption{\textbf{Step-size sensitivity of the E3 numerical outputs.} The two step sizes use matched initializations at each bandwidth and the same stopping diagnostic. Separation and core-label TV change modestly at all three settings; the lower-tail distance at $s=0.14$ is more sensitive than at $s=0.18$ and $0.24$. This is a finite-step comparison, not an extrapolation to zero step size.}
  \label{fig:e3_stepsize_full}
\end{figure}

\paragraph{Relation to the constructive theorem.}
The sweep uses the geometry and statistics of
Corollary~\ref{cor:sample_resolution_nonmemorization}. All cutoffs in
this sweep satisfy \(s<1/2\), below the proof's sufficient floor
\(s_0\ge1/2\). The measured outputs illustrate simultaneous separation
and sample-resolution diversity for these numerical settings.

\newpage
\subsection{E4: Geometric scaling and posterior-localization diagnostics}
\label{sec:e4}

\subsubsection{Uniform localization on finite affine grids}
\label{subsec:revision_e4}
For the finite affine cloud in Appendix~\ref{subsec:revision_e3}, take $K=[-20.5,20.5]\times\{0\}$ and $r_0=8.786475\,\sigma_{\min}$. The ambient codimension is $50$ and the affine line has infinite reach. On the reference segment, $h_{\mathrm{reg}}=h_{\mathrm{fill}}=1/3$ and Proposition~\ref{prop:empirical_mass_localization} applies with $a_z=1/82$, $K_{\mathrm{mass}}=8$, and $R_{\mathrm{loc}}=12$. The ball-count checks reduce to counts on an interlaced rational grid. Closed balls of radius $1/3$ contain at least one center, and closed balls of each integer radius $j$ up to $12$ contain at least $2j$ centers at every projection. Monotonicity gives the lower bound between these radii. The minimum three-center span is one, so any interval of length $2t$ contains at most $4t+2\le 8t$ for $t\ge1/2$. For $1/3\le t<1/2$ it contains at most two, which also meets the upper bound.

The three $h_{\mathrm{fill}}/\sigma_{\min}$ ratios are $0.667$, $0.606$, and $0.556$. The finite schedule gives $C_{\mathrm{scale}}<4.30$, and the required mass-regularity radii are below $7.57$. The general result gives a common localization constant below $412$. A sharper calculation uses the interlaced structure and explicitly accounts for the missing mass near the finite grid's edges. Bounding the scale moment on 1,024 radial intervals with directed interval arithmetic gives $C_{\mathrm{loc}}<2.607$, $2.486$, and $2.432$, respectively, over the full tube above $K$. The sampled maxima $1.242$, $1.215$, and $1.204$ from 16,302 queries per setting illustrate the gap between sample diagnostics and a uniform bound.

\subsubsection{Spherical bias and empirical posterior localization}
E4 separates two complementary questions: whether the near-manifold bias
follows the finite-schedule population prediction, and whether a finite
empirical cloud exhibits controlled posterior moments at sampled tube
locations. The two diagnostics use their own matched scale schedules.

\paragraph{(i) Population bias: setup and computation.}
We use the proper population density \eqref{eq:proper_population_mixture} with $Q_{\mathcal M}=\operatorname{Unif}(S^2(R))$, $m=2$, $c=10$, and equal sampling of the 64 scales
\begin{equation}
  \sigma_i=\sigma_{\min}16^{(i-1)/63},\quad i=1,\ldots,64.
  \label{eq:e4_population_schedule}
\end{equation}
The radii are $R\in\{0.5,1,2\}$ and the lower cutoffs are $\sigma_{\min}\in\{0.01,0.015,0.02,0.03,0.04,0.06\}$. Thus the absolute schedule is held fixed when comparing different $R$ at one lower cutoff, and the relative schedule is held fixed throughout.

For $x=\varrho e\in V$, we evaluate the normalized radial field as $R\mathbb E_{\widetilde w}[\cos\theta]-\varrho$. The angular integral uses 1024 Gauss--Legendre nodes in $t=\cos\theta\in[-1,1]$, whose reference density is constant on $S^2$; the 64 noise scales are summed exactly. The calculation uses double precision and stabilized log weights. We search 600 equally spaced radii in $[0.7R,1.3R]$, select the sign-change bracket closest to $R$, and bisect for at most 120 iterations, stopping when the absolute field residual is below $10^{-14}$ or the bracket width is below $10^{-13}$. The resulting radius is denoted
\(\widehat\varrho_\star\), with reported bias
\(\widehat b_\star=|\widehat\varrho_\star-R|\). This directly estimates
the near-sphere stationary branch of the population field.

\begin{figure}[!htb]
  \centering
    \includegraphics[width=0.75\linewidth]{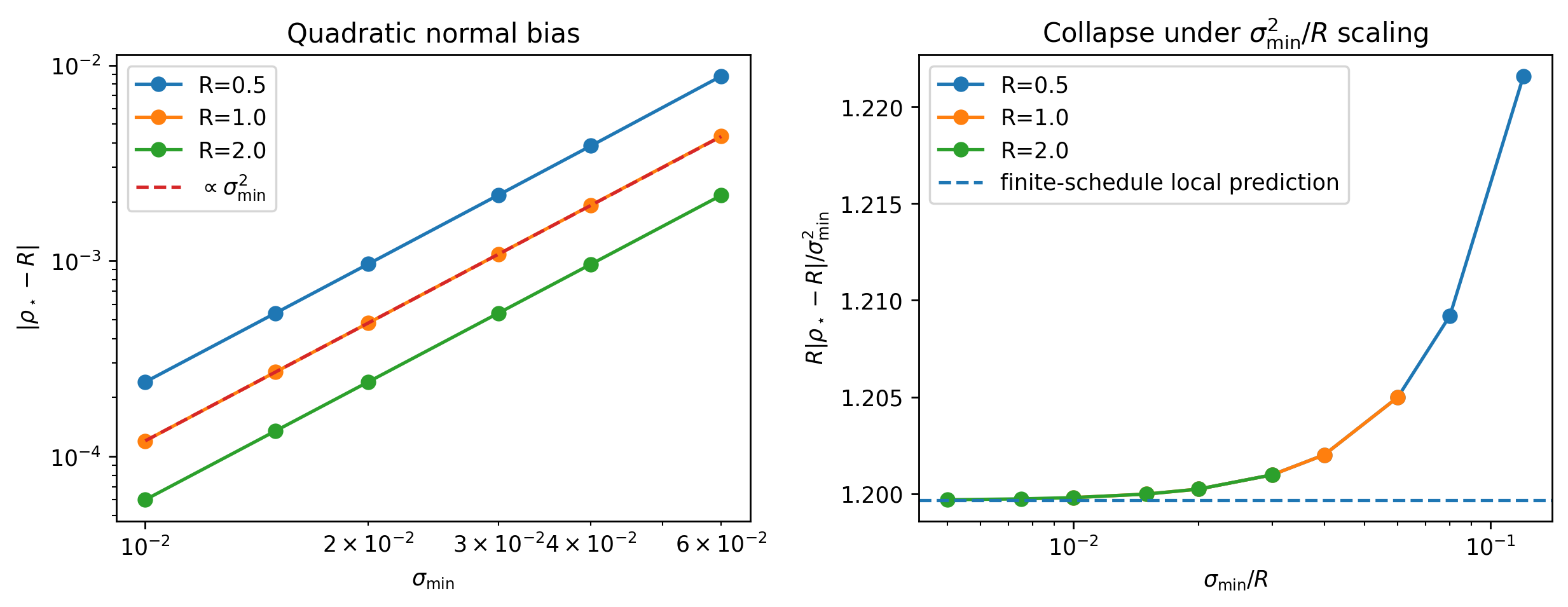}

  \caption{\textbf{Population spherical bias with a fixed relative noise schedule (E4).} Left: the located near-sphere bias for three sphere radii, with a quadratic guide. Right: the normalized bias approaches $C_{\mathrm{fs}}=1.199652317\ldots$, computed from the 64-scale schedule rather than fitted to the numerical outputs. The quadratic guide on the left is anchored to one numerical point and illustrates slope only; the horizontal line on the right is the analytic finite-schedule local reference. All reported biases come from population quadrature and radial root finding.}
  \label{fig:e4_scaling_full}
\end{figure}

\paragraph{Population bias: results and reference.}
For each $R$, ordinary least squares on the six pairs $(\log\sigma_{\min},\log\widehat b_\star)$ gives the slopes in Table~\ref{tab:e4_bias_scaling}. Figure~\ref{fig:e4_scaling_full} displays both the unnormalized biases and $R\widehat b_\star/\sigma_{\min}^2$. At the smallest relative cutoff $\sigma_{\min}/R=0.005$, the latter is $1.1996890$, compared with the prescribed-schedule prediction $C_{\mathrm{fs}}=1.1996523$ from Appendix~\ref{subsec:e4_reference_calculations}; their relative discrepancy is $0.00306\%$. Across the full sweep the discrepancy ranges up to $1.828\%$ at $\sigma_{\min}/R=0.12$, consistent with the local rather than exact finite-noise nature of the prediction. The nearly quadratic slopes, rescaled curves, and limiting constant together closely track the predicted finite-schedule noise--curvature dependence.

\begin{table}[!htb]
\centering
\small
\caption{\textbf{E4 population bias scaling.}
Slopes use all six tested lower cutoffs. Here
\(B(s)=R\widehat b_\star/s^2\) at fixed \(R\) and
\(s=\sigma_{\min}\). The last column reports the relative discrepancy
between \(B(0.01)\) and
\(C_{\mathrm{fs}}=1.199652317\ldots\).}
\label{tab:e4_bias_scaling}
\begin{tabular}{@{}rrrrr@{}}
\toprule
$R$ & fitted slope & $B(0.01)$ & $B(0.06)$ & error at $0.01$ (\%) \\
\midrule
0.5 & 2.0091 & 1.2002404 & 1.2215817 & 0.04902 \\
1.0 & 2.0022 & 1.1997992 & 1.2049867 & 0.01225 \\
2.0 & 2.0006 & 1.1996890 & 1.2009770 & 0.00306 \\
\bottomrule
\end{tabular}
\end{table}

\paragraph{(ii) Empirical posterior localization: setup.}
We draw a single cloud of $M=10^4$ uniform centers on $S^2(1)\subset\mathbb R^{12}$ and hold it fixed for the diagnostic. Here $R=1$, $\sigma_{\min}=0.05$, and the 24 geometric scales span $[0.05,0.8]$. At each radius $r$ we sample 512 independent base points $z\sim\operatorname{Unif}(S^2(1))$ and uniformly distributed unit directions $n_z\in N_z\mathcal M$, and evaluate $x=z+rn_z$. There are 14 logarithmically spaced radii between $r_{\min}/4$ and $0.30$, together with the additional radius $r=r_{\min}=0.05\sqrt{10}\approx0.158114$. Thus 15 radii and 7680 query locations are tested, all with $r<R$; the nearest-point projection is $z$. The normal space has one sphere-radial direction and nine directions orthogonal to the containing three-dimensional subspace. Sampling a normalized ten-dimensional Gaussian in this basis includes both inward and outward radial components, rather than only radial points in $V$.

The PyTorch implementation uses seed 999, one fixed empirical cloud, new queries at each radius, and single-precision log-sum-exp evaluation of the finite center--scale weights. It measures
\begin{equation}
  \widehat C_{\mathrm{loc}}(x)
  =\frac{\sum_{j,i}\widetilde w_{ij}(x)\|\mu_j-z\|^2}
         {m\max\{\sigma_{\min}^2,r^2/c\}},
  \qquad z=\Pi_{\mathcal M}(x),\quad m=2,\ c=10.
  \label{eq:e4_empirical_localization_ratio}
\end{equation}
For each radius we record the median, 90th and 99th percentiles, and sampled maximum. These statistics summarize 7680 sampled tube locations for one fixed empirical cloud over the tested radius range.

\begin{figure}[!htb]
  \centering
     \includegraphics[width=0.75\linewidth]{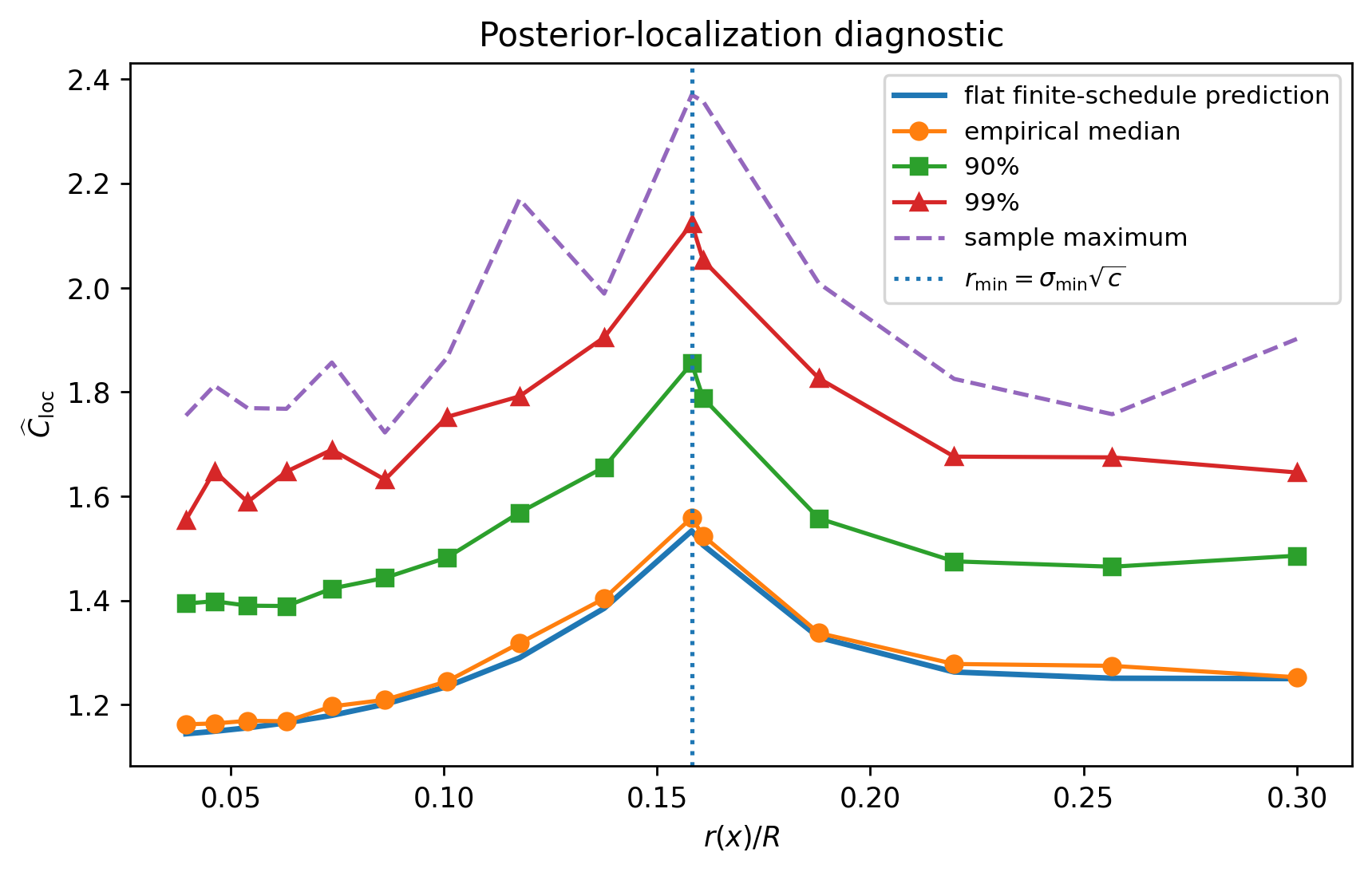}
  \caption{\textbf{Sampled posterior-localization diagnostic (E4).}
At each of 15 radii, 512 tube queries are evaluated against one fixed
cloud of \(10^4\) uniform spherical centers. The median closely follows
the radius-dependent flat reference for the same 24-scale schedule, while
the higher quantiles and sampled maximum show the variation across query
locations.}
  \label{fig:e4_cloc_full}
\end{figure}

\paragraph{Empirical localization: comparison and results.}
The reference curve is $\mathcal C_{\mathrm{flat},N}(r)$ from \eqref{eq:e4_flat_finite_ratio}, evaluated with the same 24-scale schedule as the empirical localization diagnostic. For each radius, let $\widehat C_{0.5}(r)$ denote the empirical median and define
\begin{equation}
  e_{\mathrm{med}}(r)
  =\frac{|\widehat C_{0.5}(r)-\mathcal C_{\mathrm{flat},N}(r)|}
          {\mathcal C_{\mathrm{flat},N}(r)}.
  \label{eq:e4_median_reference_discrepancy}
\end{equation}
The arithmetic mean of $e_{\mathrm{med}}$ over the 15 radii is $1.174\%$ and its maximum is $2.180\%$. The largest observed per-radius 99th percentile is $2.1233$, and the maximum over all queried ratios is $2.3701$. At $r=r_{\min}$, the flat reference is $1.533454$ and the empirical median is $1.558910$ (Table~\ref{tab:e4_localization_selected}). Figure~\ref{fig:e4_cloc_full} shows that the median follows the finite-schedule reference while the tails remain larger. The peak near \(r_{\min}\) coincides with the switch in the normalization \(\max\{\sigma_{\min}^2,r^2/c\}\).

\begin{table}[!htb]
\centering
\small
\caption{\textbf{E4 localization at selected radii.} All entries use $R=1$, one fixed cloud of $10^4$ centers, and 512 queries per radius. The row $r=0.158114$ corresponds to $r_{\min}$ before rounding. Aggregate discrepancies reported in the text use all 15 radii, not only these five rows.}
\label{tab:e4_localization_selected}
\begin{tabular}{@{}rrrrr@{}}
\toprule
$r$ & $\mathcal C_{\mathrm{flat},N}(r)$ & median & 99th percentile & sampled maximum \\
\midrule
0.039528 & 1.144095 & 1.162255 & 1.5547 & 1.7547 \\
0.117725 & 1.289768 & 1.317887 & 1.7915 & 2.1698 \\
0.158114 & 1.533454 & 1.558910 & 2.1233 & 2.3701 \\
0.219636 & 1.262671 & 1.278057 & 1.6759 & 1.8251 \\
0.300000 & 1.250007 & 1.252572 & 1.6455 & 1.9018 \\
\bottomrule
\end{tabular}
\end{table}

\subsection{First-entry coverage of learned scores on a circle}
\label{sec:learned_circle}
\paragraph{Exact target and assumptions.}
We use the uniform unit circle in $\mathbb R^2$ and the finite unified target with 64 log-spaced levels in $[0.12,0.18]$. At a query radius $\rho$ and fixed scale $\sigma_i$, the conditional squared chord moment is $2(1-I_1(t)/I_0(t))$, with $t=\rho/\sigma_i^2$ and $I_j$ modified Bessel functions. The bound $t^{-1}+t^{-2}$ on this moment, averaged over the precision weights, gives $C_{\mathrm{loc}}=2.86$ for the $r_0=0.18$ tube. Thus $\kappa_0<0.258$ and $\varepsilon_0<0.021$, below our stopping distance $0.05$. The origin is the only exact off-tube critical point. Its effective covariance rank equals the ambient dimension, so the rank-gap condition in Assumption~\ref{assumption:off_tube_geometry} is inapplicable there. Directly evaluating the Hessian gives eigenvalues at least $14.43\,\beta(0)$ and establishes repulsion.

To check ascent up to the stopping tube, write $A(t)=I_1(t)/I_0(t)$.
The Bessel identities give $A(t)\sim t/2$ at zero and
$A'(t)=1-A(t)/t-A(t)^2$. For a given $\sigma_i$, the difference
$A(t)-\sigma_i^2t$ is positive near zero. At a possible first zero with
$\rho=\sigma_i^2t\le0.95$, its derivative would be
$1-2\sigma_i^2-\rho^2>0$, which excludes that zero. Thus the exact radial
score is outward for $0<\rho\le0.95$. At $\rho\ge1.05$, the posterior
barycenter has norm at most one, so the exact radial score is inward.

\paragraph{Neural field.}
We fit three independent, unrestricted $2\to64\to64\to2$ tanh MLPs with 4,482 parameters each. After 5,000 ordinary DSM updates, auxiliary supervision on the exact score and its spatial derivatives tightens the margins of the learned field. The sampler uses $V_\theta=s_\theta/50$, a constant change of time. On the annulus $0.25\le\|x\|\le2.5$ outside the distance-$0.05$ tube, interval checks of the frozen weights establish, at every angle,
\[
 \dot d\le-0.35d,\qquad \|P_TV_\theta\|\le\eta_T ,
\]
where $d=|\|x\|-1|$ and the three upper bounds $\eta_T$ are $(4.16,3.76,4.36)\times10^{-3}$. The enclosures use the stored weights and binary64 matrix-product error bounds. The learned radial motion has the same sign as the exact radial score, giving ascent of the exact log density up to entry on this annulus. Central and annular interval checks also verify one repelling off-tube equilibrium per learned model within $\|x\|\le2.5$. They lie within $6.4\times10^{-4}$ of the origin, with minimum Jacobian eigenvalue real part at least $4.37$.

\paragraph{Coverage of the same learned outputs.}
We use the full prior $\mathcal N(0,0.8^2I_2)$ and return the first entry to the distance-$0.05$ tube. Let $B_j$ be one of twelve equal-angle regions, $f_R$ the Rayleigh density of the initial radius, and $D_\theta(\rho)$ the bound on pre-entry angular drift from radius $\rho$. For $\rho<0.95$, divide the angular speed bound $\eta_T/u$ by the inward radial speed $0.35(1-u)$ and integrate from $\rho$ to $0.95$. For $\rho>1.05$, divide by $0.35(u-1)$ and integrate from $1.05$ to $\rho$. Starts in the tube stop immediately, giving $D_\theta(\rho)=0$. Initial angle is uniform and independent of radius. Angles initially more than $D_\theta(\rho)$ from either boundary stay in $B_j$. Consequently, with $\tau_{\mathrm{stop}}$ the first-entry time,
{
\begin{equation}
 \mathbb P\{\tau_{\mathrm{stop}}<\infty,\ \arg X_{\tau_{\mathrm{stop}}}\in B_j\}
 \ge\int_{0.25}^{2.5}\left[\frac1{12}-\frac{D_\theta(\rho)}{\pi}\right]_+
 f_R(\rho)\,d\rho .
 \label{eq:circle_mlp_coverage}
\end{equation}
}
At least $94.47\%$ of the full prior is in the checked annulus. Its complement costs at most $0.461$ percentage points per region, and the checked drift costs at most $0.687$ further points. We evaluate the integral using lower sums over 1,024 radial strata on each side of the stopping tube.

For real-arithmetic Euler with step $h=10^{-4}$, its piecewise-linear interpolation has an additional defect at most $LhM\le2.67\times10^{-4}$, where $L$ bounds the field Jacobian and $M$ its magnitude. Replacing the radial lower speeds by their margins minus this defect, and adding it to the tangential upper speed, gives the numerical coverage column. The rule takes the first intersection of the interpolated path with the tube. Separate RK4 runs from 16,384 shared Gaussian starts give $100\%$ observed entry and regional TV near $0.01$ for each model.

\begin{table}[!htb]
\centering
\small
\caption{Per-region mass for the same three learned circle fields and first-entry rule. Continuous and Euler lower bounds include prior placement and neural drift. The Euler column also includes the interpolation defect. The observed column gives the smallest empirical region mass in the separate RK4 runs.}
\label{tab:revision_learned_coverage}
\begin{tabular}{@{}rrrr@{}}
\toprule
Seed & Continuous lower (\%) & Euler lower (\%) & Observed minimum (\%) \\
\midrule
$0$ & $7.21$ & $7.17$ & $7.98$ \\
$1$ & $7.28$ & $7.24$ & $7.92$ \\
$2$ & $7.18$ & $7.14$ & $7.94$ \\
\bottomrule
\end{tabular}
\end{table}

These bounds use prior placement and uniform enclosures of the learned fields.
At the end of ordinary DSM training, two checkpoints give positive lower
bounds of $0.78\%$ and $0.80\%$. The third yields successful sampled entries
but fails the sufficient normal-margin test. Auxiliary score and derivative
supervision tightens the verified margins on the same geometry.

\clearpage

\section{Finite-Sample Refinements and Output Resolution}
\label{app:modular_coverage}
The exact empirical model can have gaps between its training centers. This appendix records how an additional posterior-moment residual changes the normal and tangential bounds, and why coverage must be stated at an explicit resolution.

\subsection{A posterior-moment residual}
\label{app:finite_sample_refinements}
\label{subsec:finite_sample_diversity_scope}
Replace \eqref{eq:posterior_localization} by
\begin{equation}
 \sum_{j,i}\widetilde w_{ij}(x)\|\mu_j-z_x\|^2
 \le C_{\mathrm{loc}}m\max\{\sigma_{\min}^2,r(x)^2/c\}+h_M^2,
 \quad x\in\Omega,\quad0<r(x)<r_0.
 \label{eq:finite_sample_posterior_localization}
\end{equation}
Here $h_M\ge0$ has units of length and describes an assumed posterior residual, not automatically a fill distance or a sampling rate. Set $b_M=h_M^2/(2\tau)$. Averaging the reach inequality gives
\[
 \dot r\le\begin{cases}
 -\beta(x)(r-\varepsilon_0-b_M),&r\le r_{\min},\\
 -\beta(x)((1-\kappa_0)r-b_M),&r>r_{\min}.
 \end{cases}
\]
With $\kappa_0<1$, define
\begin{equation}
 \varepsilon_{\mathcal M,h}
 =
 \max\left\{\min\{r_{\min},\varepsilon_0+b_M\},
                  \frac{b_M}{1-\kappa_0}\right\},
 \label{eq:finite_sample_manifold_tube_radius}
\end{equation}
and assume $\varepsilon_{\mathcal M,h}<r_0$.
Above this radius each applicable radial bound is strictly inward. In the first regime either $r>\varepsilon_0+b_M$ or that regime is empty; in the second, $(1-\kappa_0)r>b_M$. Keeping the off-tube hypotheses of Theorem~\ref{thm:manifold_convergence} therefore gives
$r(x_\infty)\le\varepsilon_{\mathcal M,h}$ almost surely and the same localization for maximizers. Also,
\[
 \varepsilon_{\mathcal M,h}\le\varepsilon_{\mathcal M}
                  +\frac{h_M^2}{2\tau(1-\kappa_0)},
\]
by $\min\{r,a+b\}\le\min\{r,a\}+b$.

\begin{proposition}[Residual-aware tangential retention]
\label{prop:finite_sample_retention}
Under these residual-aware convergence conditions, choose
$\varepsilon_{\mathcal M,h}<r_{\mathrm{stop}}<r_{\mathrm{tr}}<r_0$ and set
\begin{equation}
 D_h(r)=\begin{cases}
 r-\varepsilon_0-b_M,&r\le r_{\min},\\
 (1-\kappa_0)r-b_M,&r>r_{\min}.
 \end{cases}
 \label{eq:residual_radial_factor}
\end{equation}
Then $r(X_{\mathrm{stop}})\le r_{\mathrm{stop}}$ almost surely and the projected refinement length is at most
\begin{equation}
 \delta_{\mathrm{stop},h}
 =L_\Pi\int_{r_{\mathrm{stop}}}^{r_{\mathrm{tr}}}
 \frac{\sqrt{C_{\mathrm{loc}}m\max\{\sigma_{\min}^2,r^2/c\}+h_M^2}}
      {D_h(r)}\,dr,
 \qquad L_\Pi=\frac{\tau}{\tau-r_{\mathrm{tr}}}.
 \label{eq:finite_sample_retention_integral}
\end{equation}
Under the corresponding Gaussian band conditions, the source--sink coverage theorem holds with $\delta_{\mathrm{cov}}=\delta_{\mathrm{noise}}+\delta_{\mathrm{stop},h}$ and the same probability errors.
\end{proposition}
\begin{proof}
The residual-aware normal bound gives finite entrance and $-\dot r\ge\beta D_h(r)>0$ between the two surfaces. Projection and Jensen's inequality now give
$\|\dot z\|\le L_\Pi\beta\sqrt{C_{\mathrm{loc}}m\max\{\sigma_{\min}^2,r^2/c\}+h_M^2}$.
Divide these inequalities and integrate over decreasing $r$, splitting at $r_{\min}$ if necessary. Each denominator is bounded away from zero on the relevant closed interval. The source--sink proof uses this full path-length bound and is otherwise unchanged. With $h_M=0$ and the original radius ordering, the two integrals evaluate to \eqref{eq:explicit_early_stopping_retention_radius}.
\end{proof}
A fixed positive $h_M$ prevents these sufficient bounds from supporting $r_{\mathrm{stop}}\to0$ at fixed reach. Thus both the normal error and the retention calculation must account for a nonvanishing posterior residual.

\subsection{Training distance and the resolution of coverage}
For a compactly supported reference distribution define
\begin{equation}
 h_{\mathrm{cov},M}=\sup_{z\in\operatorname{supp}(Q_{\mathcal M})}
                   \operatorname{dist}(z,\mathcal S_M).
 \label{eq:population_support_covering_radius}
\end{equation}
This is the geometric covering radius, distinct from the posterior residual $h_M$. The affine grid has the explicit value in \eqref{eq:resolved_grid_covering_radius}.

\begin{corollary}[Training distance for the stopped output]
\label{cor:source_sink_training_separation}
Under Theorem~\ref{thm:source_sink_coverage}, for every $\eta\ge0$,
\begin{equation}
\begin{split}
 \mathbb P[\operatorname{dist}(X_{\mathrm{stop}},\mathcal S_M)>\eta]
 \ge{}&(1-\zeta)Q_{\mathcal M}\bigl(\{z:\operatorname{dist}(z,\mathcal S_M)
                  >\eta+r_{\mathrm{stop}}+\delta_{\mathrm{cov}}\}\bigr)\\
 &-\epsilon_{\mathrm{Pois}}-\epsilon_{\mathrm{init}}.
\end{split}
 \label{eq:source_sink_ambient_training_separation}
\end{equation}
\end{corollary}
\begin{proof}
Distance to the training set is one-Lipschitz. A reference label at the displayed distance, within $\delta_{\mathrm{cov}}$ of the stopped projection, gives a projection at distance greater than $\eta+r_{\mathrm{stop}}$. The ambient output differs from that projection by at most $r_{\mathrm{stop}}$. Apply the same safe-interior event comparison as in the coverage theorem.
\end{proof}

Conversely, if a manifold-valued output has coverage
$\mathbb P[Z_{\mathrm{out}}\in U]\ge\alpha Q_{\mathcal M}(U^{-\delta})-\epsilon_{\mathrm{cov}}$
and training-set concentration
$\mathbb P[\operatorname{dist}(Z_{\mathrm{out}},\mathcal S_M)\le\eta]\ge1-\epsilon_{\mathrm{mem}}$,
then
\begin{equation}
 \alpha Q_{\mathcal M}(\{z:\operatorname{dist}(z,\mathcal S_M)>\eta+\delta\})
 \le\epsilon_{\mathrm{mem}}+\epsilon_{\mathrm{cov}}.
 \label{eq:memorization_coverage_compatibility}
\end{equation}
Indeed, apply coverage to $U=\{z:\operatorname{dist}(z,\mathcal S_M)>\eta\}$ and use the one-Lipschitz property to include the displayed reference set in $U^{-\delta}$. For $\alpha>0$ and zero errors this implies $h_{\mathrm{cov},M}\le\eta+\delta$ by continuity and the definition of support. This distinguishes coarse coverage from novelty: assigning each point of a uniformly populated circle to its nearest training center preserves coverage at half the training spacing, but exactly memorizes that finite set.

\subsection{From a finite stop to an endpoint}
\label{app:endpoint_retention}
Finite ambient path length and the projection derivative bound imply
\begin{equation}
 \int_{\tau_{\mathrm{stop}}}^\infty
 \left\|\frac d{dt}\Pi_{\mathcal M}(x(t))\right\|dt
 \le\frac{\tau}{\tau-r_{\mathrm{stop}}}
       \int_{\tau_{\mathrm{stop}}}^\infty\|\dot x(t)\|dt<\infty.
 \label{eq:projected_tail_finite_length}
\end{equation}
If this projected tail has a uniform bound $\delta_{\mathrm{tail}}$, the same reference coupling gives endpoint coverage with resolution $\delta_{\mathrm{cov}}+\delta_{\mathrm{tail}}$. Pathwise finiteness does not supply a small uniform bound. The sample-resolution result instead proves endpoint retention from finite-mixture gradient signs in the capture cells, so its endpoint guarantee does not require this additional tail assumption.

\clearpage
\section{Penalty-Guided Generation: Convergence, Geometry, and Experimental Details}
\label{app:control_details}

This appendix expands Section~\ref{subsec:optimization_control}. We first analyze the generic penalty-guided objective, then explain how the task gradient perturbs manifold attraction and resolved diversity. We finally record the concrete penalties used in the controllable-generation experiments. The theoretical statements use the exact unified score $\nabla\log p_{MN}$; replacing it by the learned state-only score $s_\theta$ gives an inexact-gradient implementation.

\subsection{Penalty-guided ascent and stationary points}
\label{subsec:conditional_controllable_generation}
\label{app:control_convergence_proof}

Let $E_y:\mathbb R^d\to\mathbb R_+$ be a differentiable task penalty,
with smaller values indicating better task agreement. In general, we do
\textit{not} require
\[
E_y(x)=0\quad\Longleftrightarrow\quad x\text{ satisfies the task}.
\]
For example, a squared measurement residual may vanish at exact feasibility,
whereas smooth inequality penalties and classifier losses generally need not. For a finite penalty weight $\lambda\ge0$, define
\begin{equation}
F_y(x)=\log p_{MN}(x)-\lambda E_y(x),
\qquad
\dot x=\nabla F_y(x)
=\nabla\log p_{MN}(x)-\lambda\nabla E_y(x).
\label{eq:app_penalty_objective}
\end{equation}
The task is imposed through a soft penalty relaxation rather than an
exact projection.

Assume that $E_y$ is nonnegative and real analytic. Along the exact flow,
\begin{equation}
\frac{d}{dt}F_y(x(t))
=\|\nabla F_y(x(t))\|^2\ge0.
\label{eq:penalty_objective_ascent}
\end{equation}
Since $F_y\le\log p_{MN}$ and
$\log p_{MN}(x)\to-\infty$ as $\|x\|\to\infty$, the superlevel set
\begin{equation}
\Omega_y(x_0)=\{x:F_y(x)\ge F_y(x_0)\}
\label{eq:controlled_superlevel_set}
\end{equation}
is compact and forward invariant. The same analytic-gradient argument as
in Lemma~\ref{lemma:boundedness_and_convergence} therefore gives global
forward existence, finite path length, and convergence to a stationary
point $x_{\infty,y}$, which satisfies
\begin{equation}
\nabla\log p_{MN}(x_{\infty,y})
=\lambda\nabla E_y(x_{\infty,y}).
\label{eq:app_controlled_stationarity}
\end{equation}
This is stationarity of the penalized objective, not a claim of global
optimality or exact satisfaction of a hard constraint.

For the explicit exact-score update
\begin{equation}
x_{n+1}
=x_n+\eta_n\left[
\nabla\log p_{MN}(x_n)-\lambda\nabla E_y(x_n)
\right],
\label{eq:app_penalty_discrete_update}
\end{equation}
choose $\overline\eta>0$ and $L_y>0$ such that $\nabla F_y$ is
$L_y$-Lipschitz on a compact convex set containing all trial segments
\[
\bigl\{x+t\nabla F_y(x):
       x\in\Omega_y(x_0),\ 0\le t\le\overline\eta\bigr\},
\]
and require
\begin{equation}
0<\underline\eta\le\eta_n\le\overline\eta<2/L_y
\qquad\text{for every }n.
\label{eq:app_penalty_step_bounds}
\end{equation}
Such choices are possible under the analytic assumption: first fix a
finite trial-step cap, bound the Hessian on a compact convex set containing
its trial segments, and then reduce $\overline\eta$ if necessary. This
smoothness condition covers the update segments, not merely the possibly
nonconvex superlevel set.

With $\Delta_n:=x_{n+1}-x_n$, the ascent estimate gives
\begin{equation}
\begin{aligned}
F_y(x_{n+1})-F_y(x_n)
&\ge
\left(\eta_n^{-1}-\frac{L_y}{2}\right)\|\Delta_n\|^2\\
&\ge
\left(\overline\eta^{-1}-\frac{L_y}{2}\right)\|\Delta_n\|^2.
\end{aligned}
\label{eq:main_control_ascent}
\end{equation}
Thus, by induction, all iterates stay in $\Omega_y(x_0)$ and the objective
values increase to a finite limit. Moreover,
\[
\|\nabla F_y(x_n)\|
=\frac{\|\Delta_n\|}{\eta_n}
\le\underline\eta^{-1}\|\Delta_n\|.
\]
The sufficient-ascent estimate and this gradient--step bound, together
with the analytic-gradient inequality on the compact accumulation set,
yield $\sum_n\|\Delta_n\|<\infty$. Hence the iterates converge to a
stationary point satisfying \eqref{eq:app_controlled_stationarity}.
The uniform lower step bound and strict upper margin in
\eqref{eq:app_penalty_step_bounds} are part of this discrete convergence
guarantee.

\subsection{How guidance perturbs manifold attraction}
\label{app:penalty_normal_robustness}

Inside the reach tube, write $z_x=\Pi_{\mathcal M}(x)$,
$r(x)=\|x-z_x\|$, and $u_x=(x-z_x)/r(x)$ for $r(x)>0$.
For the guided flow initialized at $x_0$, define the relevant tube region
by
\begin{equation}
\mathcal R_y
:=\Omega_y(x_0)\cap\{x:r(x)<r_0\}.
\label{eq:app_guided_tube_region}
\end{equation}
Retain the local geometry and contraction-gap conditions used in
Lemma~\ref{lemma:inward_score}, and require the posterior-localization
estimate on this guided region:
\begin{equation}
\sum_{j,i}\widetilde w_{ij}(x)\|\mu_j-z_x\|^2
\le
C_{\mathrm{loc}}m\max\{\sigma_{\min}^2,r(x)^2/c\},
\qquad x\in\mathcal R_y,\quad r(x)>0.
\label{eq:app_guided_posterior_localization}
\end{equation}
Here $\varepsilon_0,r_{\min},\kappa_0$ are defined as before using this
$C_{\mathrm{loc}}$, with $\kappa_0<1$ and
$\varepsilon_0<\min\{r_{\min},r_0\}$.
The domain in \eqref{eq:app_guided_posterior_localization} is essential:
guidance preserves a superlevel set of $F_y$, but need not preserve the
unguided superlevel set of $\log p_{MN}$. Localization on that unguided
set alone therefore does not justify the guided radial estimate.

The guided radial velocity is
\begin{equation}
\dot r
=
\langle u_x,\nabla\log p_{MN}(x)\rangle
-\lambda\langle u_x,\nabla E_y(x)\rangle.
\label{eq:app_guided_radial_identity}
\end{equation}
Only the outward component of $-\lambda\nabla E_y$ opposes the prior's
normal attraction. Define
\begin{equation}
\delta_y
:=
\sup_{\substack{x\in\mathcal R_y\\0<r(x)<r_0}}
\frac{\lambda[-\langle u_x,\nabla E_y(x)\rangle]_+}{\beta(x)},
\label{eq:penalty_outward_budget_definition}
\end{equation}
with $\delta_y=0$ if the indexed set is empty. This quantity is finite
under the assumptions above, since $\nabla E_y$ is bounded on the compact
set $\Omega_y(x_0)$ and $\beta(x)\ge\sigma_{\max}^{-2}$. Thus
\begin{equation}
\lambda[-\langle u_x,\nabla E_y(x)\rangle]_+
\le\beta(x)\delta_y
\qquad(x\in\mathcal R_y,\ r(x)>0).
\label{eq:penalty_outward_budget}
\end{equation}
The quantity $\delta_y$ measures harmful normal guidance relative to the
local contraction scale of the prior.

Applying the proof of Lemma~\ref{lemma:inward_score} on $\mathcal R_y$
and combining it with \eqref{eq:penalty_outward_budget} gives
\begin{equation}
\dot r\le
\begin{cases}
-\beta(x)(r-\varepsilon_0-\delta_y),
&0<r\le r_{\min},\\
-\beta(x)((1-\kappa_0)r-\delta_y),
&r_{\min}<r<r_0.
\end{cases}
\label{eq:penalty_radial_robustness}
\end{equation}
Set
\begin{equation}
\varepsilon_y
=
\max\left\{
\varepsilon_0+\delta_y,\,
\frac{\delta_y}{1-\kappa_0}
\right\},
\qquad\text{and assume }\varepsilon_y<r_0.
\label{eq:penalty_manifold_limit}
\end{equation}
If $r(x(t_0))<r_0$ at some time $t_0$, a first-exit argument keeps the
trajectory in the tube thereafter. The two radial inequalities imply
\begin{equation}
[r(x(t))-\varepsilon_y]_+
\le
[r(x(t_0))-\varepsilon_y]_+
\exp\!\left(-\frac{1-\kappa_0}{\sigma_{\max}^2}(t-t_0)\right),
\qquad t\ge t_0.
\label{eq:app_guided_normal_decay}
\end{equation}
Consequently, the stationary limit from the preceding subsection obeys
$\operatorname{dist}(x_{\infty,y},\mathcal M)\le\varepsilon_y$.
In particular, tangential or inward guidance has $\delta_y=0$ and leaves
the local normal bound unchanged, provided the localization estimate holds
on $\mathcal R_y$. A small outward component enlarges the bound
continuously. If $\varepsilon_y\ge r_0$, this sufficient contraction
guarantee no longer applies. This is an after-entry statement: global
entry from arbitrary initialization requires the corresponding off-tube
behavior of the guided field, which does not follow automatically from
the unguided assumptions.

The same local radial calculation covers norm-clipped guidance. If a
clipped penalty gradient $g_y$ is used in the field
$\dot x=\nabla\log p_{MN}(x)-\lambda g_y(x)$ and
$\|g_y(x)\|\le c_{\mathrm{clip}}$, then
\begin{equation}
\frac{\lambda[-\langle u_x,g_y(x)\rangle]_+}{\beta(x)}
\le\lambda c_{\mathrm{clip}}\sigma_{\max}^2.
\label{eq:clipped_guidance_normal_budget}
\end{equation}
This is a coarse worst-case budget; predominantly tangential directions
can have a much smaller effective $\delta_y$. The posterior-localization
estimate must hold on the region visited by this clipped flow. Since the
clipped field need not be a gradient of $F_y$, the normal estimate alone
does not imply the stationary-point convergence proved above.

\paragraph{Exact affine compatibility.}
If the data lie in an affine space $\mathcal A=a_0+V$ and
$P_{V^\perp}\nabla E_y(x)=0$, then the penalty changes only motion within
$V$. The normal component remains
\begin{equation}
P_{V^\perp}\dot x
=
-\beta(x)P_{V^\perp}(x-a_0),
\qquad
r_{\mathcal A}(x(t))
\le r_{\mathcal A}(x(0))e^{-t/\sigma_{\max}^2}.
\label{eq:penalty_affine_normal_flow}
\end{equation}
Hence a tangential task penalty can alter which point is selected inside
the data subspace without changing exact normal convergence.

\subsection{Guidance strength and retained diversity}
\label{app:controlled_coverage}

The normal calculation controls fidelity but not probability allocation along the manifold. A simple sufficient condition follows from the capture regions in Appendix~\ref{app:resolved_endpoint_capture}. Suppose an unguided region has a strict inward margin $\gamma>0$ on a boundary face with outward tangential normal $v$:
\[
\langle v,\nabla\log p_{MN}(x)\rangle\le-\gamma.
\]
If, on the same face,
\begin{equation}
\lambda|\langle v,\nabla E_y(x)\rangle|<\gamma,
\label{eq:penalty_tangential_margin}
\end{equation}
then the guided field still points inward on that face. If this condition
holds on every tangential face and the guided field also points inward
on the normal boundary, the compact capture region remains forward
invariant. Under the analytic assumptions above, trajectories entering
it converge to stationary points within that region. Multiple disjoint
task-compatible regions can therefore retain distinct outcomes.

The source--sink balance itself changes under guidance:
\begin{equation}
\begin{split}
\nabla\!\cdot\!\left(
A_Np_{MN}[\nabla\log p_{MN}-\lambda\nabla E_y]
\right)
={}&p_{\sigma_{\max}}^Q-p_{\sigma_{\min}}^Q+\mathcal R_N\\
&-\lambda A_N\nabla\!\cdot(p_{MN}\nabla E_y).
\end{split}
\label{eq:penalty_modified_flux}
\end{equation}
Thus penalty guidance modifies regional probability through the additional
flux term. Structurally, manifold fidelity is controlled by the normal
component of guidance, while retained diversity depends on whether its
tangential component preserves multiple task-compatible regions.

\subsection{Large-step controllable-generation comparison}
\label{app:control_implementation}

We evaluate robustness to guided displacements using time-varying penalties
in two samplers. Here we describe the ODE experiments on a synthetic four-disk
task and CIFAR-10 classifier guidance. Appendix~\ref{subsec:revision_guidance}
gives a matched reverse VE-SDE protocol on the four-disk task.

For the ODE experiments, we use the sampling framework of
\citet{zhou2026smoothing} with step budgets $K\in\{1000,100\}$. The conditioned model is evaluated with the externally
scheduled noise level, whereas the noise-unconditioned model receives only the
current state. Unless stated otherwise, the sampling setup is matched across
the two models.

For both tasks, the guidance strength is annealed over the sampling trajectory:
\begin{equation}
\lambda_k=1-\frac{k}{K-1},\qquad k=0,\ldots,K-1.
\label{eq:control_linear_guidance_schedule}
\end{equation}

\paragraph{Synthetic four-disk constraint.}
We use the same $100$-dimensional latent construction as in the toy studies:
the first two coordinates form four separated disks and the remaining $98$
coordinates are Gaussian, followed by the canonical embedding into
$\mathbb R^{3072}$. Both learned score models use the same standard U-Net
backbone implemented in PyTorch, differing in whether the noise level is
provided as an input.

The task requires the first two latent coordinates to be positive, selecting
the first-quadrant disk. Writing $y=W^\top x$, where $W$ denotes the canonical projection onto
the 100-dimensional latent coordinates, we use
\begin{equation}
E_{\mathrm{toy}}(x)
=
\operatorname{softplus}(-\alpha y_1)
+
\operatorname{softplus}(-\alpha y_2),
\label{eq:app_toy_control_penalty}
\end{equation}
and replace the prior score $s_k$ by
\begin{equation}
d_k=s_k-\lambda_k\nabla E_{\mathrm{toy}}(x_k),
\qquad
s_k=
\begin{cases}
s_\theta(x_k,\sigma_k),&\text{noise conditioned},\\
s_\theta(x_k),&\text{noise unconditioned}.
\end{cases}
\label{eq:app_toy_guided_score}
\end{equation}
The same penalty and linear schedule are used for both models.

\paragraph{CIFAR-10 classifier guidance.}
For CIFAR-10 we use the NCSN++ architecture of
\citet{song2020score} for both models, again differing only in whether the
scheduled noise level is supplied. A fixed classifier pretrained on CIFAR-10
provides guidance toward the \emph{horse} class, following the
classifier-guidance principle of \citet{dhariwal2021diffusion}. Define
\begin{equation}
E_{\mathrm{cls}}(x)
=
-\log p_\phi(y_{\mathrm{horse}}\mid\operatorname{prep}(x)),
\label{eq:app_cifar_control_penalty}
\end{equation}
where $\operatorname{prep}$ denotes the classifier preprocessing. At every sampling
step, PyTorch autograd computes the classifier direction
$g_k=-\nabla E_{\mathrm{cls}}(x_k)$ together with the current generative
score $s_k$. We normalize the guidance \emph{per sample} toward the current score
magnitude,
\begin{equation}
\rho_k=\frac{\|s_k\|_2}{\|g_k\|_2+\epsilon_{\mathrm{stab}}},
\qquad
d_k=s_k+\lambda_k\rho_k g_k,
\label{eq:app_cifar_norm_matched_guidance}
\end{equation}
with the norms taken over all image coordinates and
$\epsilon_{\mathrm{stab}}>0$ preventing division by zero. The classifier
direction is therefore rescaled toward the current score magnitude before
multiplication by the common factor $\lambda_k$. This removes guidance-scale differences
between the two score parameterizations and makes the 1000-versus-100-step
comparison focus on robustness to the guided displacement.

\begin{figure}[!htb]
    \centering
    \includegraphics[width=0.5\linewidth]{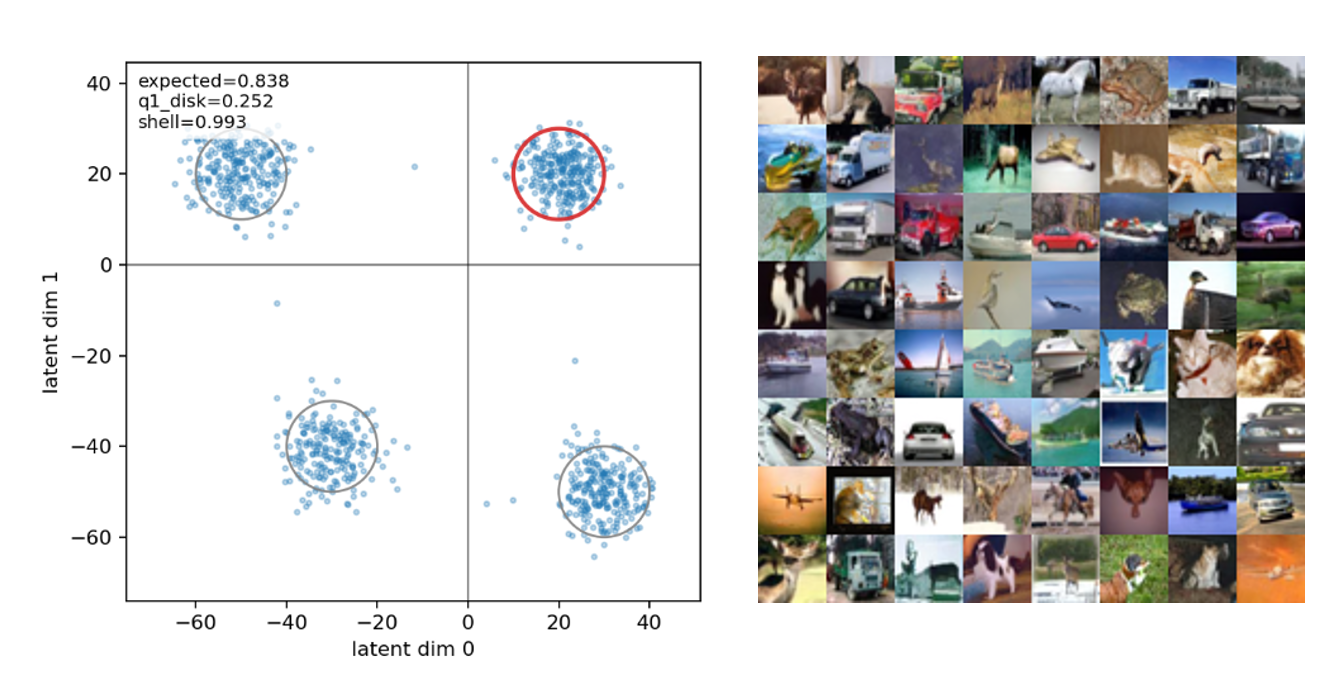}
    \caption{\textbf{Unguided 100-step conditioning baseline.} This ablation isolates the failure in
    Figure~\ref{fig:large_step_control} to the interaction between coarse steps and
    guidance rather than the reduced step count alone.}
    \label{fig:unguided_100step_control}
\end{figure}

\paragraph{Unguided 100-step control.}
To distinguish a coarse-solver failure from a guidance-induced failure, we also
run the noise-conditioned models with the same 100-step sampler but with
guidance removed as in the \cref{fig:unguided_100step_control}. These unguided runs still generate well-formed samples,
showing that the degradation in Figure~\ref{fig:large_step_control} is associated
with the interaction between large guided updates and noise conditioning rather
than the 100-step discretization by itself.

\paragraph{Interpretation.}
In a noise-conditioned sampler, the scheduled $\sigma_k$ determines the
marginal scored by the network. A large guided move can place $x_{k+1}$
outside the high-density region of the next scheduled marginal, so the
conditioned score is evaluated at an atypical state for
$\sigma_{k+1}$. In the unified field, the current state determines the
effective contribution of the noise scales. The guided and unguided
100-step results are consistent with this mechanism. The reverse VE-SDE
experiment quantifies the mismatch using a noise-scale proxy and a
paired score-error probe.

\subsubsection{Matched reverse VE-SDE comparison on four disks}
\label{subsec:revision_guidance}
\paragraph{Shared geometry and sampler.}
The four disks have radius 10 and centers $(20,20)$, $(-50,20)$, $(-30,-40)$, and $(30,-50)$. The first is the target. The remaining 98 latent coordinates are Gaussian, and we embed the 100-dimensional latent state canonically in $\mathbb R^{3072}$. We hold the trained noise-conditioned (C) and noise-unconditioned (U) scores fixed. The shared reverse VE sampler uses $K$ geometrically spaced noise levels with cutoffs $(0.01,50)$. At each of the $K-1$ stochastic steps, it adds $\Delta_i$ times the guided score and an independent $\mathcal N(0,\Delta_i I)$ increment, with $\Delta_i=\sigma_i^2-\sigma_{i+1}^2$. Both methods make one denoising call at the final level. The shared penalty on the planar coordinates is $\operatorname{softplus}(-8x_1)+\operatorname{softplus}(-8x_2)$, with its gradient norm capped at 10. During the main stage, the strength decays as $\lambda_i=\lambda_0(1-i/(K-1))$ and reaches zero at the final denoising call. U then takes 20 unguided updates $x\leftarrow x+1.35\,\sigma_*(x)f_U(x)$, where $f_U$ is its network output before score scaling and $\sigma_*$ is the continuous nearest-center scale defined below. The total score-call budgets are $K$ for C and $K+20$ for U.

\paragraph{Validation and test.}
For each $K\in\{50,100,250,500,1000\}$, both methods search the same candidates
\[
\lambda_0\in
\{0,0.001,0.0025,0.005,0.01,0.025,0.05,0.1,0.15,0.2,0.25,0.5,0.75,1\}.
\]
Preliminary validation favored $\lambda_0=0.025$ on a coarser grid, motivating the four smaller positive candidates. Each candidate is evaluated with 128 validation outputs. We maximize target success and break ties in favor of the smaller $\lambda_0$. C selects $0.01$ at $K=50$ and $0.025$ otherwise. U selects $0.025$ at all budgets. The test results in Table~\ref{tab:revision_guidance} aggregate three independent paired streams of 128 draws, with shared initial states and noise increments within each budget. Failures count as unsuccessful. Entropy uses the distribution of successful hits across four equal-area rings and eight sectors. A hit in a wrong disk has zero distance to the union and counts as a target miss.

Paired bootstrap intervals at 95\% confidence for the success differences are positive at all budgets, conditional on the trained networks and validation choices. In the matched-strength test of all five budgets and nine nonzero strengths in $[0.025,1]$, U has higher target success in all 45 pairs. At $K=50$, its success remains $98.2$--$100\%$ over $\lambda_0=0.025$--$0.25$.

\paragraph{Measuring state--noise mismatch.}
Following \citet[Section 4 and Appendix A.7]{zhou2026smoothing}, we compare the assigned $\sigma_i$ at each conditioned score query with the distance proxy $\sigma_*(x_i)=\min_j\|x_i-\mu_j\|/\sqrt{3072}$. We average $|\log(\sigma_*/\sigma_i)|$ over 24 preselected query times on 192 held-out trajectories per setting. To calibrate the influence of finite sample spacing, we also measure this gap on target-disk reference points with known Gaussian corruption at the same scales. The reference mean is $0.66$. At $K=50$, C's mean gaps are $0.84$, $1.39$, $3.87$, and $4.64$ for $\lambda_0=0,0.1,0.5,1$, respectively. At $K=1000$ they remain within $0.66$--$0.67$.

In a paired one-step probe, the guided and unguided proposals use the same incoming state and the same Gaussian increment. Because the penalty acts in the plane, the normal component outside the embedded data subspace and its exact score $-x_\perp/\sigma^2$ agree in the two proposals. Evaluating C on both states isolates the change in noise-scaled normal-score error due to the guided displacement.

\begin{table}[!htb]
\centering
\small
\caption{Percentage increase in C's mean noise-scaled normal-score RMSE at the guided proposal, relative to its paired unguided proposal.}
\label{tab:revision_guidance_probe}
\begin{tabular}{@{}rrrr@{}}
\toprule
Main steps & $\lambda_0=0.1$ & $\lambda_0=0.5$ & $\lambda_0=1$ \\
\midrule
50 & $1.32$ & $15.71$ & $32.59$ \\
1,000 & $0.01$ & $0.10$ & $0.52$ \\
\bottomrule
\end{tabular}
\end{table}

The increase in score error at stronger guidance and coarser steps is
consistent with guided moves scaling as $\Delta_i\lambda_i$. At the
validation-selected strengths, the probe error changes by less than $0.06\%$
in magnitude. The two samplers thus probe guidance robustness under
time-varying penalties, while the analysis above concerns fixed penalties.

\subsection{Downstream classification under data imbalance}
\label{app:downstream_augmentation}

\paragraph{Task and fixed real-data split.}
We construct a binary CIFAR-10 cat--dog task from the official training split.
The low-data training set contains 100 cats and 1000 dogs, and the same selected
images are reused in every comparison. The balanced-real reference keeps these
100 cats and adds 900 distinct real cats. The repetition baseline repeats the same 100 cats to obtain 1000 cat
instances, increasing minority-class weight without adding new source images. Evaluation uses all cat and dog images from the
official CIFAR-10 test split, and balanced accuracy is the mean of the two
per-class accuracies.

\paragraph{Generated-data conditions.}
For the main low-data comparison, the noise-conditioned diffusion model and
noise-unconditioned model are each trained using \emph{only the same 100 real
cats} available to the downstream task. Each generates 900 additional cats,
which are combined with the 100 real cats and 1000 real dogs. As a separate
control, we also evaluate 900 cats generated by models pretrained on the full
CIFAR-10 training distribution; these serve as broad-data augmentation controls. For the low-data comparison, both generators follow the training and sampling configuration of \citet{zhou2026smoothing}, with the training subset restricted to the same 100 cats.

\paragraph{Why this probes finite-data generalization.}
Sample-level memorization and generalization of noise-unconditioned scores
are studied directly in \citet{zhou2026smoothing}; here we ask whether the
difference is useful downstream. In the memorization limit, generated
augmentation contributes little new information beyond repeatedly presenting
the same finite minority set, so its effect approaches reweighting. A
generator that generalizes beyond those examples can instead provide useful
within-class variation, effectively broadening the minority-class support
available to the classifier. We therefore use held-out classification
accuracy as a functional measure of whether this additional variation is
useful downstream. 

\paragraph{Downstream classifier.}
Every condition trains the same CIFAR-style ResNet-18 binary classifier from
scratch. We replace the ImageNet stem by a $3\times3$, stride-one convolution
and remove max pooling. Training uses random $32\times32$ crops with padding
four, horizontal flips, CIFAR-10 normalization, SGD with learning rate $0.1$,
momentum $0.9$, weight decay $5\times10^{-4}$, Nesterov momentum, and a
100-epoch cosine schedule. The batch size is 128. To match the optimization
budget, every method draws 2000 training examples per epoch; for the
1100-image biased baseline this sampling is with replacement and therefore
maintains a 10:1 imbalance in expectation. Within each classifier
seed, all methods use the same initial ResNet-18 weights. Results report mean and standard deviation over three classifier seeds (0, 1, 2).
\begin{table}[!htb]
\centering
\small
\setlength{\tabcolsep}{4.0pt}
\caption{\textbf{Full downstream augmentation results.}
Test accuracy (\%, mean$\pm$std). The two 100-cat generator rows are the
data-scarce comparison emphasized in
Section~\ref{subsec:downstream_experiments}.}
\label{tab:downstream_full}
\begin{tabular}{@{}lccc@{}}
\toprule
Training condition & Cat & Dog & Bal. Acc. \\
\midrule
Biased real & 11.80$\pm$1.25 & 96.73$\pm$0.21 & 54.27$\pm$0.56 \\
Balanced real reference & 79.17$\pm$0.76 & 79.63$\pm$1.40 & 79.40$\pm$1.01 \\
Repeated cats (1000) & 13.63$\pm$0.76 & 97.13$\pm$0.46 & 55.38$\pm$0.41 \\
Pretrained diffusion aug. & 77.53$\pm$1.34 & 82.07$\pm$1.36 & 79.80$\pm$1.35 \\
Pretrained unconditioning aug. & 77.00$\pm$2.62 & 82.33$\pm$2.01 & 79.67$\pm$2.28 \\
100-cat diffusion aug. & 26.17$\pm$1.65 & 93.33$\pm$0.47 & 59.75$\pm$0.95 \\
100-cat unconditioning aug. & 38.80$\pm$3.45 & 91.87$\pm$0.95 & 65.33$\pm$2.03 \\
\bottomrule
\end{tabular}
\end{table}

The pretrained-generator controls reach 79.80\% and 79.67\% balanced accuracy,
close to the 79.40\% balanced-real reference, showing that both model classes
can provide effective augmentation when trained on broad CIFAR-10 data. The
data-scarce comparison is more informative: with both generators trained on
only the same 100 cats, noise unconditioning improves balanced accuracy by
5.58 percentage points over diffusion and by 9.95 points over repetition. The
corresponding cat-class gains are 12.63 and 25.17 points.

The gain over repetition cannot be explained by class-count balancing alone,
while the comparison with diffusion holds the scarce real training set fixed.
The results are therefore consistent with finite-data generative generalization
providing additional minority-class variation that is useful for classification
of held-out real data.

\paragraph{Ten-class CIFAR-10 extension.}
To test whether the downstream effect persists beyond binary classification,
we repeat the experiment on the full CIFAR-10 ten-class task. The cat class
remains data-scarce, with only 100 fixed real training images, while each of
the other nine classes uses 1000 fixed real images. The seven conditions
mirror the binary experiment: biased real data, a balanced-real reference,
repeated minority examples, and augmentation from pretrained or
100-cat-trained diffusion and noise-unconditioned generators. Each
augmentation condition adds 900 generated cats, producing 1000 cat entries
in total.

We use the same CIFAR-style ResNet-18 training protocol, now with ten output
classes. Each method uses 10,000 sampled training examples per epoch; within
each classifier seed, all seven conditions share the same network
initialization. Results are averaged over the same three classifier seeds.
Evaluation uses the complete CIFAR-10 test set. Since the test set is balanced
across classes, overall and macro accuracy coincide; we additionally report
cat accuracy and the mean accuracy over the other nine classes.

\begin{table}[!htb]
\centering
\small
\setlength{\tabcolsep}{4.0pt}
\caption{\textbf{Ten-class CIFAR-10 classification with a scarce cat class.}
Accuracy (\%, mean$\pm$std over three classifier seeds). ``Other-9'' is the
mean per-class accuracy over the nine non-cat classes.}
\label{tab:downstream_tenclass}
\begin{tabular}{@{}lccc@{}}
\toprule
Training condition & Overall & Cat & Other-9 \\
\midrule
Biased real
& 81.49$\pm$1.02 & 22.40$\pm$2.25 & 88.06$\pm$0.95 \\
Balanced real reference
& 86.00$\pm$1.01 & 71.30$\pm$2.31 & 87.63$\pm$0.90 \\
Repeated cats (1000)
& 79.90$\pm$1.43 & 14.73$\pm$1.04 & 87.14$\pm$1.53 \\
Pretrained diffusion aug.
& 86.17$\pm$0.93 & 69.40$\pm$2.26 & 88.03$\pm$0.83 \\
Pretrained unconditioning aug.
& 84.77$\pm$1.05 & 66.87$\pm$1.72 & 86.76$\pm$1.27 \\
100-cat diffusion aug.
& 81.59$\pm$0.29 & 29.50$\pm$0.53 & 87.37$\pm$0.31 \\
100-cat unconditioning aug.
& {83.50$\pm$0.73}
& {42.97$\pm$3.30}
& {88.00$\pm$0.48} \\
\bottomrule
\end{tabular}
\end{table}

The ten-class results preserve the low-data pattern observed in the binary
task. With both generators trained on the same 100 cats, noise-unconditioned
augmentation improves overall accuracy by 1.91 percentage points over
diffusion (83.50\% versus 81.59\%) and cat accuracy by 13.47 points
(42.97\% versus 29.50\%). The advantage appears in all three classifier
seeds. At the same time, mean accuracy over the other nine classes is
essentially preserved, and is slightly higher for noise unconditioning
(88.00\% versus 87.37\%). Relative to simply repeating the same 100 cats,
noise-unconditioned augmentation improves overall accuracy by 3.60 points
and cat accuracy by 28.23 points. Thus the downstream gain is not specific
to binary cat--dog discrimination and remains visible in the full ten-class
recognition problem.

\subsubsection{Independent training subsets and image diagnostics}
\label{subsec:revision_augmentation}
We run the binary augmentation protocol on three independent draws of 100 cats from the 1,000-cat real-data pool. For each subset and method, three independently trained generators each produce 900 cats. Each classifier uses those 900 generated cats, the corresponding 100 real cats, and the same 1,000 real dogs. We train three classifier seeds per generator. The diffusion and unconditioning entries in Table~\ref{tab:revision_subsets} average classifier seeds within each generator, then report the mean and sample standard deviation across generators. The two real-only conditions summarize three classifier seeds each.

\begin{table}[!htb]
\centering
\small
\caption{Balanced accuracy (mean $\pm$ sample SD, in percent) across three independently drawn 100-cat subsets.}
\label{tab:revision_subsets}
\begin{tabular}{@{}lrrr@{}}
\toprule
Training condition & Subset 1 & Subset 2 & Subset 3 \\
\midrule
Biased real & $54.48\pm0.28$ & $54.69\pm0.03$ & $53.28\pm0.66$ \\
Repeated cats & $55.29\pm0.36$ & $55.20\pm0.41$ & $55.13\pm0.62$ \\
Diffusion augmentation & $59.21\pm1.07$ & $59.98\pm0.71$ & $60.31\pm0.45$ \\
Unconditioning augmentation & $64.34\pm0.68$ & $65.05\pm0.14$ & $65.95\pm0.51$ \\
\bottomrule
\end{tabular}
\end{table}

For the generators trained on the fixed split in Appendix~\ref{app:downstream_augmentation}, we draw 1,000 cats per method and compute two image diagnostics. The nearest-neighbor ratio of \citet[Appendix B.3.2]{zhou2026smoothing} flags an image when its nearest-to-second-nearest training-image distance ratio in pixel space is below $2/3$. We use the fraction flagged as a proxy for concentration around individual training images. We measure PRDC coverage \citep{naeem2020reliable} in pretrained ResNet-18 features with identical preprocessing. The reference set contains the 100 training cats and 900 other real cats. Coverage is the fraction of real-reference neighborhoods containing at least one generated sample, with each neighborhood radius set by the distance to the fifth nearest other real reference image.

\begin{table}[!htb]
\centering
\small
\caption{Training-image proximity and reference coverage on the same 1,000 generated cats per method.}
\label{tab:revision_image_diagnostics}
\begin{tabular}{@{}lrr@{}}
\toprule
Method & NN-ratio proxy (\%) $\downarrow$ & PRDC coverage (\%) $\uparrow$ \\
\midrule
Diffusion & $47.1$ & $72.2$ \\
Unconditioning & $23.6$ & $89.1$ \\
\bottomrule
\end{tabular}
\end{table}

A separate comparison uses 100 generated cats per method and the same 100 training images. The mean nearest-training pixel-space $L_2$ distance is $6.62$ for diffusion and $10.33$ for noise unconditioning. The mean leave-one-out distance among the training images is $15.79$, providing a reference for the spacing of the real training set. Together with the ratio and coverage measurements, the larger nearest-training distance indicates reduced proximity to individual training images in this setting.

\clearpage

\section{Population Objectives: Convergence and Regional Coverage}
\label{app:population_extension}

We prove Remark~\ref{remark:main_population_extension} directly for a
\emph{smoothed population mixture}. No training-size limit is needed, and no
score of the possibly singular clean distribution $Q_{\mathcal M}$ is assumed.

\subsection{The population learning target and posterior identities}
\label{app:population_objective}
Let $Q_{\mathcal M}$ be a probability distribution with compact support in
$\mathcal M$, and let $\pi$ be a probability measure on a fixed interval
$[\sigma_{\min},\sigma_{\max}]\subset(0,\infty)$. Normalizing a finite nonzero
scale measure does not change the density in
\eqref{eq:proper_population_mixture}, which we reuse here:
\begin{equation}
 p^{\mathrm{pop}}(x)
 =\frac{\int\sigma^2p_\sigma^Q(x)\,\pi(d\sigma)}
        {\int\sigma^2\,\pi(d\sigma)},\qquad
 p_\sigma^Q(x)=\int\mathcal N(x;\mu,\sigma^2I)Q_{\mathcal M}(d\mu).
 \label{eq:population_general_objective}
\end{equation}
The distribution $Q_{\mathcal M}$ need not have a density. Taking
$\pi=N^{-1}\sum_i\delta_{\sigma_i}$ gives the finite-schedule population
objective in the main-text remark.

For independent $\mu\sim Q_{\mathcal M}$ and $\sigma\sim\pi$, draw
$X\mid\mu,\sigma\sim\mathcal N(\mu,\sigma^2I)$. The population state-only loss obeys
\begin{equation}
 \begin{aligned}
  \mathcal L_u^{\mathrm{pop}}(\theta)
  &=\frac12\mathbb E\left[
       \sigma^2\left\|s_\theta(X)-\frac{\mu-X}{\sigma^2}\right\|^2\right]\\
  &=\frac{\mathbb E_\pi\sigma^2}{2}
     \mathbb E_{X\sim p^{\mathrm{pop}}}
       \|s_\theta(X)-\nabla\log p^{\mathrm{pop}}(X)\|^2+C,
 \end{aligned}
 \label{eq:population_dsm_identity}
\end{equation}
where $C$ is independent of $\theta$.
This follows by the same conditional variance decomposition as
\eqref{eq:unconditioned_training_excess}. The weighted component-score second
moment is $\mathbb E[\sigma^2\| (\mu-X)/\sigma^2\|^2]=d$, so the
score-independent remainder is finite. Thus population noise unconditioning
also learns the gradient of a single fixed log density.

For the population objective, define the posterior weights and moments by
\begin{align}
 w_x(d\mu,d\sigma)
 &=\frac{\sigma^2\mathcal N(x;\mu,\sigma^2I)
                  Q_{\mathcal M}(d\mu)\pi(d\sigma)}
         {p^{\mathrm{pop}}(x)\int\sigma^2\pi(d\sigma)},
 \label{eq:population_posteriors}\\
 \beta(x)&=\int\sigma^{-2}w_x(d\mu,d\sigma),\qquad
 \bar\mu(x)=\int\mu\,\widetilde w_x(d\mu,d\sigma),\notag\\
 \widetilde w_x(d\mu,d\sigma)
 &=\frac{\sigma^{-2}}{\beta(x)}w_x(d\mu,d\sigma).\notag
\end{align}
Differentiation under the integral gives
\begin{equation}
 \begin{aligned}
  \nabla\log p^{\mathrm{pop}}(x)&=\beta(x)(\bar\mu(x)-x),\qquad
       \sigma_{\max}^{-2}\le\beta(x)\le\sigma_{\min}^{-2},\\
  \nabla^2\log p^{\mathrm{pop}}(x)&=C_x-\beta(x)I,\qquad
       C_x=\operatorname{Cov}_{w_x}\!\left(\frac{\mu-x}{\sigma^2}\right).
 \end{aligned}
 \label{eq:population_score_hessian}
\end{equation}
In particular, every critical point belongs to
$\operatorname{conv}(\operatorname{supp}Q_{\mathcal M})$.

\subsection{Manifold convergence for a population objective}
\label{app:population_convergence}
The same proof applies with posterior and landscape conditions imposed on
the population objective.

\textbf{Analyticity, bounded trajectories, and complete flow.}
By Tonelli's theorem, $p^{\mathrm{pop}}$ is a positive probability density.
Compact center support and positive bounded scales allow the Gaussian kernels
to be integrated locally uniformly in a complex neighborhood of each real
$x$. The resulting density is real analytic, as is its logarithm since the
density is strictly positive. With
$B=\sup_{\mu\in\operatorname{supp}Q_{\mathcal M}}\|\mu\|$, for $\|x\|\ge B$,
\begin{equation}
 p^{\mathrm{pop}}(x)
 \le(2\pi\sigma_{\min}^2)^{-d/2}
       \exp\!\left[-\frac{(\|x\|-B)^2}{2\sigma_{\max}^2}\right].
 \label{eq:population_tail_bound}
\end{equation}
Every positive-density superlevel is therefore compact. For nonempty compact
$K$, define $\Omega=\{x:p^{\mathrm{pop}}(x)\ge\min_Kp^{\mathrm{pop}}\}$.
The objective increases along $\dot x=\nabla\log p^{\mathrm{pop}}(x)$, and the
proof of Lemma~\ref{lemma:boundedness_and_convergence} gives global forward
existence, finite path length, and convergence from every point of $K$.
Applying this with $K=\{x_0\}$ gives convergence for every initialization.
Moreover,
$\|\nabla\log p^{\mathrm{pop}}(x)\|\le\sigma_{\min}^{-2}(B+\|x\|)$.
Both the field and its negative have at most linear growth, so the two-sided
complete-flow and Jacobian arguments used for source--sink marking also apply.

\textbf{Normal attraction and off-tube instability.}
Impose the population version of posterior localization on the corresponding
$\Omega$:
\begin{equation}
 \int\|\mu-z_x\|^2\,\widetilde w_x(d\mu,d\sigma)
 \le C_{\mathrm{loc}}m\max\{\sigma_{\min}^2,r(x)^2/c\},
 \qquad x\in\Omega,\quad0<r(x)<r_0.
 \label{eq:population_posterior_localization}
\end{equation}
The reach inequality holds for every support point $\mu$. Integrating it
against $\widetilde w_x$ gives the same posterior-alignment and radial bounds
\eqref{eq:posterior_alignment} and
\eqref{eq:piecewise_radial_distance_dynamics}. Jensen's inequality gives the
same projected-speed and retention bounds. The covariance identity
\eqref{eq:population_score_hessian} likewise gives the same positive
log-density Hessian eigenvalue under the population shell-rank condition.

\begin{proposition}[Population manifold convergence]
\label{prop:population_manifold_convergence}
Let $\mathcal M$ have the compact $C^2$ geometry of
Theorem~\ref{thm:manifold_convergence}, and suppose its local geometry,
posterior-localization, contraction-gap, and off-tube shell-rank assumptions
hold for \eqref{eq:population_general_objective} on its own $\Omega$.
If all off-tube critical points have nonsingular
$\nabla^2\log p^{\mathrm{pop}}$ and $\nu\ll\operatorname{Leb}_d$ with
$\nu(K)=1$, then every trajectory from $K$ converges and
\begin{equation}
 \operatorname{dist}(x_\infty,\mathcal M)
 \le\varepsilon_0=\frac{C_{\mathrm{loc}}m}{2\tau}\sigma_{\min}^2
 \quad\nu\text{-almost surely}.
 \label{eq:population_manifold_convergence}
\end{equation}
Every local maximizer in $\Omega$ and every global maximizer obey the same
bound without the hyperbolicity or initialization-density conditions.
For a compactly supported population in an affine space $\mathcal A=a_0+V$,
the exact normal identity and convergence in
\eqref{eq:affine_exact_normal_score} hold for every start, without the curved
manifold assumptions.
\end{proposition}
\begin{proof}
The analytic argument gives a critical limit. The population radial bound
excludes the intermediate contraction region. Off-tube hyperbolicity makes
the compact off-tube critical set finite; its ascent-unstable equilibria have
zero-volume stable sets by the same stable-manifold and inverse-flow argument
as Lemma~\ref{lemma:sufficient_saddle_avoidance}. This proves the almost-sure
bound. The Hessian test and Gaussian decay prove the optimizer statements as
in Proposition~\ref{prop:optimizers_in_manifold_tube}. In the affine case the
posterior barycenter remains in $\mathcal A$, giving
$\dot n=-\beta(x)n$ for $n=P_{V^\perp}(x-a_0)$. The analytic limit therefore
lies in $\mathcal A$ and in the convex hull of the population support.
\end{proof}

\subsection{Population coverage and the disappearance of sampling error}
\label{app:population_coverage}
For the finite geometric schedule
$\sigma_i=\sigma_{\min}e^{(i-1)h}$, $N\ge2$ and $h>0$, take
$\pi=N^{-1}\sum_i\delta_{\sigma_i}$. With the same
$A_N=h\sum_i\sigma_i^2$ as before,
\begin{equation}
 A_Np^{\mathrm{pop}}=h\sum_i\sigma_i^2p_{\sigma_i}^Q,\qquad
 \mathcal R_N=A_N\Delta p^{\mathrm{pop}}
                 -(p_{\sigma_{\max}}^Q-p_{\sigma_{\min}}^Q).
 \label{eq:population_flux_residual}
\end{equation}

For the population objective, the empirical--reference cutoff discrepancy
vanishes exactly, so $\epsilon_{\mathrm{Pois}}$ consists only of finite-grid
quadrature error. The same $L^1$ quadrature argument used in the proof of
Proposition~\ref{prop:source_sink_input_errors} applies with
$Q_{\mathcal M}$ in place of $\widehat Q_M$, giving
\begin{equation}
\epsilon_{\mathrm{Pois}}
\le
dh\log\frac{\sigma_{\max}}{\sigma_{\min}}
+\frac h2\sqrt{2d}.
\label{eq:population_quadrature_error}
\end{equation}
This estimate is sufficient to show that the error vanishes as the
log-scale grid is refined, but need not be quantitatively sharp in high
ambient dimension.

The residual is integrable with zero integral. The preceding complete-flow
and compact-superlevel properties allow the occupation-density proof of
Lemma~\ref{lemma:source_sink_mark} to be applied with $p^{\mathrm{pop}}$ in
place of $p_{MN}$. Its mark lies on the \emph{population ascent trajectory}
and satisfies the same total-variation bound, with
$\epsilon_{\mathrm{init}}=\operatorname{TV}(\nu,p_{\sigma_{\max}}^Q(x)\,dx)$.
The reference-label coupling and Gaussian projection argument are unchanged.

\begin{corollary}[Population regional coverage]
\label{cor:population_regional_coverage}
Under Proposition~\ref{prop:population_manifold_convergence}, suppose
$\mathcal M$ is connected and the radius ordering and Gaussian band conditions
of Theorem~\ref{thm:source_sink_coverage} hold for the population flow and the
finite geometric schedule. Then $r(X_{\mathrm{stop}})\le r_{\mathrm{stop}}$
almost surely, and for every Borel $U\subseteq\mathcal M$,
\begin{equation}
\mathbb P_\nu[
Z_{\mathrm{tr}}\in U,\,
Z_{\mathrm{stop}}\in U]
\ge
(1-\zeta)Q_{\mathcal M}(U^{-\delta_{\mathrm{cov}}})
-\epsilon_{\mathrm{Pois}}
-\epsilon_{\mathrm{init}},
\label{eq:population_regional_coverage}
\end{equation}
where
\[
\epsilon_{\mathrm{Pois}}
=
\frac12
\left\|
A_N\Delta p^{\mathrm{pop}}
-
(p_{\sigma_{\max}}^Q-p_{\sigma_{\min}}^Q)
\right\|_1.
\]
Here the stopping times, projections, and coverage resolution are defined as
before, using the population field and its posterior-localization constant.
\end{corollary}
\begin{proof}
The population radial and projected-speed inequalities give the same full
refinement-length bound. Apply the marked-point comparison above, then follow
its reference label through the proof of
Theorem~\ref{thm:source_sink_coverage}.
\end{proof}

\paragraph{Continuous log-scale mixing.}
For $\pi(d\sigma)=d\sigma/[\sigma\log(\sigma_{\max}/\sigma_{\min})]$ on
$0<\sigma_{\min}<\sigma_{\max}<\infty$, the same normalized objective becomes
\begin{equation}
 p^{\mathrm{pop}}(x)=
 \frac{\int_{\sigma_{\min}}^{\sigma_{\max}}\sigma p_\sigma^Q(x)\,d\sigma}
      {\int_{\sigma_{\min}}^{\sigma_{\max}}\sigma\,d\sigma}.
 \label{eq:population_continuous_objective}
\end{equation}
Using $\partial_\sigma p_\sigma^Q=\sigma\Delta p_\sigma^Q$ gives the exact
stationary balance
\begin{equation}
 \nabla\!\cdot\!\left[
  \left(\int_{\sigma_{\min}}^{\sigma_{\max}}\sigma\,d\sigma\right)
      p^{\mathrm{pop}}\nabla\log p^{\mathrm{pop}}\right]
 =p_{\sigma_{\max}}^Q-p_{\sigma_{\min}}^Q.
 \label{eq:population_continuous_flux}
\end{equation}
Thus $\epsilon_{\mathrm{Pois}}=0$. Under the corresponding population geometric
and initialization hypotheses, coverage holds with only the Gaussian event
loss and $\epsilon_{\mathrm{init}}$. Initializing exactly from
$p_{\sigma_{\max}}^Q$ also removes $\epsilon_{\mathrm{init}}$, provided the
geometric assumptions hold on the superlevels visited by that unbounded
initialization. For the compactly supported initialization of the main
theorem, or a Gaussian approximation to the broad smoothing, its mismatch
must still be retained. A general scale measure $\pi$ gives the convergence
result above, but this exact two-cutoff balance specifically uses log-scale
mixing with the stated $\sigma^2$ weight.

\paragraph{Distributional meaning of the coverage bound.}
The proof also yields an ambient Wasserstein bound for the stopped
\emph{population} output:
\begin{equation}
 W_1(P_{X_{\mathrm{stop}}},Q_{\mathcal M})
 \le r_{\mathrm{stop}}+\delta_{\mathrm{cov}}
      +\operatorname{diam}(\mathcal M)
          \min\{1,\zeta+\epsilon_{\mathrm{Pois}}+\epsilon_{\mathrm{init}}\}.
 \label{eq:population_stopped_wasserstein}
\end{equation}
Indeed, couple the actual mark to $Y^Q\sim p_{\sigma_{\min}}^Q$ with mismatch
probability at most $\epsilon_{\mathrm{Pois}}+\epsilon_{\mathrm{init}}$, and, given $Y^Q$,
attach $Z$ through the reference posterior kernel in
\eqref{eq:source_sink_reference_coupling}. Then $Z\sim Q_{\mathcal M}$.
On mark agreement and the Gaussian good event,
$d_{\mathcal M}(Z,Z_{\mathrm{stop}})\le\delta_{\mathrm{cov}}$; on their
complement the Euclidean distance between these two manifold points is at
most $\operatorname{diam}(\mathcal M)$. Adding
$\|X_{\mathrm{stop}}-Z_{\mathrm{stop}}\|\le r_{\mathrm{stop}}$ proves the
bound. In a family satisfying the hypotheses with the displayed resolution
and probability errors tending to zero on a fixed compact manifold, the
stopped output distributions therefore converge to $Q_{\mathcal M}$ in
$W_1$. This is a conditional population-approximation conclusion, not exact
distribution recovery at fixed positive cutoffs.

\paragraph{Scope and relation to empirical non-memorization.}
The population results require the corresponding posterior and landscape
conditions on \(p^{\mathrm{pop}}\). General regional coverage concerns the
stopped output, while endpoint diversity additionally requires the
tail-retention control of Appendix~\ref{app:endpoint_retention}. The proper
sphere mixtures provide concrete population examples, and the
translation-invariant flat model serves as a local analytic reference.
Penalty guidance extends under the guided-region conditions of
Appendix~\ref{app:control_details}.

Corollary~\ref{cor:sample_resolution_nonmemorization} addresses the
complementary empirical question of diverse endpoints separated from a
finite training set. Exact reverse diffusion initialized from the appropriate
noisy distribution instead approaches its center distribution in the
zero-noise limit \citep{song2020score}: \(Q_{\mathcal M}\) for a population
score and \(\widehat Q_M\) for an empirical score
(Appendix~\ref{subsec:diffusion_endpoint_comparison}).

\end{document}